\documentclass[11pt,twoside]{article}
\usepackage{fullpage}

\usepackage{epsf}
\usepackage{fancyhdr}
\usepackage{graphics}
\usepackage{graphicx}
\usepackage{psfrag}
\usepackage{microtype}
\usepackage{subfigure}

\usepackage{algorithmic}
\usepackage{color,xcolor}
\usepackage[linesnumbered,ruled]{algorithm2e}
\DontPrintSemicolon
\usepackage{color}
\usepackage{mathrsfs}
\usepackage{stackengine}

\usepackage{amsthm}
\usepackage{amsfonts}
\usepackage{amsmath}
\usepackage{amssymb,bbm}
\usepackage{comment}

\newcommand{\stepsize}{\eta}

\newcommand{\real}{\ensuremath{\mathbb{R}}}

\newcommand{\thetahat}{\ensuremath{\widehat{\theta}}}

\newcommand{\thetabar}{\ensuremath{\bar{\theta}}}

\newcommand{\Xspace}{\ensuremath{\mathbb{X}}}
\newcommand{\Yspace}{\ensuremath{\mathbb{Y}}}

\newcommand{\abss}[1]{\left| #1 \right |}

\newcommand{\law}{\ensuremath{\mathcal{L}}}

\newcommand{\sphere}{\ensuremath{\mathbb{S}}}

\newcommand{\mydefn}{\ensuremath{:=}}
\newcommand{\poincare}{\ensuremath{\rho_P}}

\newcommand{\defn}{:=}

\newcommand{\matsnorm}[2]{|\!|\!| #1 | \! | \!|_{{#2}}}
\newcommand{\vecnorm}[2]{\| #1\|_{#2}}

\newcommand{\opnorm}[1]{\ensuremath{\matsnorm{#1}{\tiny{\mbox{op}}}}}

\newcommand{\weightedopnorm}[1]{\ensuremath{\matsnorm{#1}{ \xi \rightarrow \xi }}}

\newcommand{\inprod}[2]{\ensuremath{\langle #1 , \, #2 \rangle}}

\newcommand{\Exs}{\ensuremath{{\mathbb{E}}}}
\newcommand{\Prob}{\ensuremath{{\mathbb{P}}}}

\DeclareMathOperator{\cov}{cov}
\DeclareMathOperator{\trace}{trace}

\newtheoremstyle{named}{}{}{\itshape}{}{\bfseries}{.}{.5em}{\thmnote{#3's }#1}
\theoremstyle{named}

\theoremstyle{plain}

\newtheorem{theorem}{Theorem}
\newtheorem{proposition}{Proposition}

\newtheorem{lemma}{Lemma}

\newtheorem{corollary}{Corollary}

\newlength{\widebarargwidth}
\newlength{\widebarargheight}
\newlength{\widebarargdepth}
\DeclareRobustCommand{\widebar}[1]{%
  \settowidth{\widebarargwidth}{\ensuremath{#1}}%
  \settoheight{\widebarargheight}{\ensuremath{#1}}%
  \settodepth{\widebarargdepth}{\ensuremath{#1}}%
  \addtolength{\widebarargwidth}{-0.3\widebarargheight}%
  \addtolength{\widebarargwidth}{-0.3\widebarargdepth}%
  \makebox[0pt][l]{\hspace{0.3\widebarargheight}%
    \hspace{0.3\widebarargdepth}%
    \addtolength{\widebarargheight}{0.3ex}%
    \rule[\widebarargheight]{0.95\widebarargwidth}{0.1ex}}%
  {#1}}

\makeatletter
\long\def\@makecaption#1#2{
        \vskip 0.8ex
        \setbox\@tempboxa\hbox{\small {\bf #1:} #2}
        \parindent 1.5em  
        \dimen0=\hsize
        \advance\dimen0 by -3em
        \ifdim \wd\@tempboxa >\dimen0
                \hbox to \hsize{
                        \parindent 0em
                        \hfil
                        \parbox{\dimen0}{\def\baselinestretch{0.96}\small
                                {\bf #1.} #2
                                }
                        \hfil}
        \else \hbox to \hsize{\hfil \box\@tempboxa \hfil}
        \fi
        }
\makeatother

\long\def\comment#1{}
\definecolor{battleshipgrey}{rgb}{0.52, 0.52, 0.51}
\definecolor{darkgray}{rgb}{0.66, 0.66, 0.66}
\definecolor{darkgreen}{rgb}{0.0, 0.2, 0.13}
\definecolor{darkspringgreen}{rgb}{0.09, 0.45, 0.27}
\definecolor{dukeblue}{rgb}{0.0, 0.0, 0.61}
\definecolor{olivedrab7}{rgb}{0.24, 0.2, 0.12}
\definecolor{darkblue}{rgb}{0.0, 0.0, 0.55}
\definecolor{darkscarlet}{rgb}{0.34, 0.01, 0.1}
\definecolor{candyapplered}{rgb}{1.0, 0.03, 0.0}
\definecolor{ao(english)}{rgb}{0.0, 0.5, 0.0}
\definecolor{applegreen}{rgb}{0.55, 0.71, 0.0}

\newcommand{\widgraph}[2]{\includegraphics[keepaspectratio,width=#1]{#2}}

\usepackage{url}
\usepackage[colorlinks,linkcolor=magenta,citecolor=blue, pagebackref=true,backref=true]{hyperref}
\renewcommand*{\backrefalt}[4]{%
    \ifcase #1 \footnotesize{(Not cited.)}%
    \or        \footnotesize{(Cited on page~#2.)}%
    \else      \footnotesize{(Cited on pages~#2.)}%
    \fi}

\usepackage{nicefrac}

\usepackage{chngpage}

 \usepackage{tabularx}%

\usepackage{enumitem}
\usepackage{booktabs}

\renewenvironment{comment}{}{}

\usepackage{bm,bbm}
\usepackage{mathtools}

\newcommand{\strongconvex}{\mu}
\newcommand{\smooth}{\beta}

\newcommand{\simiid}{\overset{\mathrm{i.i.d.}}{\sim}}

\newcommand{\filtration}{\mathcal{F}}

\newcommand{\statespace}{\mathcal{S}}
\newcommand{\goodendex}{\ensuremath{\clubsuit}}

\newcommand{\Bmat}{B}

\newcommand{\neighborhood}{\mathfrak{N}}

\newcommand{\plantedbit}{z}
\newcommand{\measureQ}{\mathbb{Q}}

\newcommand{\classApprox}{\class_\mathsf{approx}}
\newcommand{\classNoise}{\probClass_\mathsf{var}}
\newcommand{\classCov}{\probClass_\mathsf{cov}}
\newcommand{\classEst}{\class_{\mathsf{est}}}
\newcommand{\classFinal}{\class_{\mathsf{final}}}
\newcommand{\classMRP}{\class_{\mathsf{MRP}}}

\newcommand{\bcar}{\begin{carlist}}
\newcommand{\ecar}{\end{carlist}}

\newcommand{\SpecMatPsilam}{\SpecMat^{(\psi, \lambda)}}
\newcommand{\projectedbvecPsilam}{\projectedbvec^{(\psi, \lambda)}}
\newcommand{\LmatPsilam}{\Lmat^{(\psi, \lambda)}}
\newcommand{\bvecPsilam}{\bvec^{(\psi, \lambda)}}
\newcommand{\projectedbvecZero}{\projectedbvec_0}
\theoremstyle{definition}
\newtheorem{example}{Example}

\newcommand{\support}{\mathrm{supp}}

\newcommand{\discount}{\gamma}

\newcommand{\oracleErr}{\delta}
\newtheorem{assumption}{Assumption}

\makeatletter
\def\namedlabel#1#2{\begingroup
   \def\@currentlabel{#2}%
   \label{#1}\endgroup
}
\makeatother

\newenvironment{carlist}
 {\begin{list}{$\bullet$}
 {\setlength{\topsep}{0in} \setlength{\partopsep}{0in}
  \setlength{\parsep}{0in} \setlength{\itemsep}{\parskip}
  \setlength{\leftmargin}{0.07in} \setlength{\rightmargin}{0.08in}
  \setlength{\listparindent}{0in} \setlength{\labelwidth}{0.08in}
  \setlength{\labelsep}{0.1in} \setlength{\itemindent}{0in}}}
 {\end{list}}

\newtheorem*{assumption1w}{Assumption 1(W)}
\newtheorem*{assumption1s}{Assumption 1(S)}

\newcommand{\numobs}{\ensuremath{n}}
\newcommand{\contraction}{\gamma}

\newcommand{\usedim}{\ensuremath{d}}
\newcommand{\SigStar}{\ensuremath{{\Sigma^*}}}

\newcommand{\reward}{r}

\newcommand{\AClass}{\mathfrak{L}}

\newcommand{\totalvariation}{d_{\mathrm{TV}}}

\newcommand{\Vhatclass}{\widehat{\mathcal{V}}}

\newcommand{\sigmaA}{\sigma_\Lmat}
\newcommand{\sigmab}{\sigma_b}

\newcommand{\varA}{\sigma_\Lmat}
\newcommand{\varb}{\sigma_b}

\newcommand{\lspace}{\LinSpace}
\newcommand{\LinSpace}{\mathbb{S}}
\newcommand{\projection}{\Pi}

\newcommand{\projecttolin}{\Pi_{\lspace}}
\newcommand{\projecttoorth}{\Pi_{\lspace^\perp}} 

\newcommand{\plantedvec}{y}

\newcommand{\projectto}[1]{\Pi_{#1}}

\newcommand{\martingale}{\Psi}

\newcommand{\rade}{\ensuremath{\varepsilon}}

\newcommand{\transition}{P}
\newcommand{\stationary}{\xi}

\newcommand{\class}{\mathbb{C}}
\newcommand{\probClass}{\mathbf{G}}

\newcommand{\hadamard}[1]{H_{#1}}

\newcommand{\adjoint}[1]{#1^*}

\newcommand{\Event}{\mathscr{E}}

\long\def\comment#1{}

\newcommand\undermat[2]{%
  \makebox[0pt][l]{$\smash{\underbrace{\phantom{%
    \begin{matrix}#2\end{matrix}}}_{\text{$#1$}}}$}#2}

\newcommand{\funcClass}{\mathcal{F}}

\newcommand{\substationary}{}
\newcommand{\statnorm}[1]{\vecnorm{#1}{\substationary}}
\newcommand{\statinprod}[2]{\inprod{#1}{#2}_{\substationary}}

\newcommand{\Amat}{\ensuremath{A}}

\newcommand{\valuefunc}{v}
\newcommand{\valuestar}{\valuefunc^*}
\newcommand{\valuebar}{\bar{\valuefunc}}
\newcommand{\valuehat}{\widehat{\valuefunc}}

\newcommand{\bigdim}{\ensuremath{D}}

\newcommand{\Lmat}{\ensuremath{L}}
\newcommand{\bvec}{\ensuremath{b}}

\newcommand{\IdMat}{\ensuremath{I}}

\newcommand{\conmax}{\ensuremath{\contraction_{\tiny{\operatorname{max}}}}}

\newcommand{\AppErr}{\ensuremath{\mathcal{A}}}
\newcommand{\EstErr}{\ensuremath{\mathcal{E}}}
\newcommand{\ResErr}{\ensuremath{\mathcal{H}}}
\newcommand{\numburn}{\ensuremath{{\numobs_{0}}}}
\newcommand{\offpar}{\ensuremath{\omega}}

\newcommand{\SpecMat}{\ensuremath{M}}
\newcommand{\SpecMatZero}{\ensuremath{\SpecMat_0}}

\newcommand{\BigEstErr}{\EstErr_\numobs(\SpecMat, \Sigma^*)}
\newcommand{\BigResErr}{\ResErr_\numobs(\sigmaA, \sigmab,
  \vbar)}

\newcommand{\vbarZero}{\bar{x}_0}
\newcommand{\vbarpsilam}{\bar{x}_{\psi, \lambda}}

\newcommand{\lammax}{\ensuremath{\lambda_{\mbox{\tiny{max}}}}}
\newcommand{\lammin}{\ensuremath{\lambda_{\mbox{\tiny{min}}}}}

\newcommand{\prefact}{\ensuremath{\alpha}}

\newcommand{\hilopnorm}[1]{\ensuremath{\matsnorm{#1}{\Xspace}}}

\newcommand{\vvec}{v}
\newcommand{\vstar}{\ensuremath{\vvec^*}}
\newcommand{\vbar}{\ensuremath{\widebar{\vvec}}}
\newcommand{\vhat}{\ensuremath{\widehat{\vvec}}}
\newcommand{\vtil}{\ensuremath{\widetilde{\vvec}}}
\newcommand{\vvecperp}{\ensuremath{\vvec^\perp}}

\newcommand{\PhiOp}{\ensuremath{\Phi_\usedim}}

\newcommand{\ProjLin}{\ensuremath{\Pi_\LinSpace}}

\newcommand{\ProbInst}{\mathbb{P}}

\newcommand{\myspan}{\ensuremath{\operatorname{span}}}

\newcommand{\Normal}{\ensuremath{N}}

\newcommand{\projectedbvec}{h}

\newcommand{\sobolevone}{\dot{\mathbb{H}}^1}
\newcommand{\Ltwospace}{\mathbb{L}^2}
\newcommand{\domain}{\Omega}
\newcommand{\afunc}{a}
\newcommand{\bfunc}{f}
\newcommand{\domaindim}{m}

\newcommand{\specpar}{\ensuremath{\varsigma}}

\newcommand{\npar}{\ensuremath{N}}

\newcommand{\kappamrp}{\nu}

\newcommand{\mytau}{\ensuremath{\tau}}

\makeatletter
\long\def\@makecaption#1#2{
        \vskip 0.8ex
        \setbox\@tempboxa\hbox{\small {\bf #1:} #2}
        \parindent 1.5em  
        \dimen0=\hsize
        \advance\dimen0 by -3em
        \ifdim \wd\@tempboxa >\dimen0
                \hbox to \hsize{
                        \parindent 0em
                        \hfil 
                        \parbox{\dimen0}{\def\baselinestretch{0.96}\small
                                {\bf #1.} {#2}
                                } 
                        \hfil}
        \else \hbox to \hsize{\hfil \box\@tempboxa \hfil}
        \fi
        }
\makeatother

\begin{document}

\begin{center}
{\bf{\LARGE{Optimal oracle inequalities for solving projected
      fixed-point equations}}}

\vspace*{.2in} {\large{
 \begin{tabular}{ccc}
  Wenlong Mou$^{ \diamond}$ & Ashwin Pananjady$^{\star}$ &
  Martin J. Wainwright$^{\diamond, \dagger}$
 \end{tabular}

}

\vspace*{.2in}

 \begin{tabular}{c}
 Department of Electrical Engineering and Computer
 Sciences$^\diamond$\\ Simons Institute for the Theory of
 Computing$^\star$\\ Department of Statistics$^\dagger$ \\ University
 of California Berkeley
 \end{tabular}

}

\vspace*{.2in}
\begin{abstract}
Linear fixed point equations in Hilbert spaces arise in a variety of
settings, including reinforcement learning, and computational methods
for solving differential and integral equations.  We study methods
that use a collection of random observations to compute approximate
solutions by searching over a known low-dimensional subspace of the
Hilbert space. First, we prove an instance-dependent upper bound on
the mean-squared error for a linear stochastic approximation scheme
that exploits Polyak--Ruppert averaging. This bound consists of two
terms: an approximation error term with an instance-dependent
approximation factor, and a statistical error term that captures the
instance-specific complexity of the noise when projected onto the
low-dimensional subspace. Using information-theoretic methods, we also
establish lower bounds showing that both of these terms cannot be
improved, again in an instance-dependent sense. A concrete consequence
of our characterization is that the optimal approximation factor in
this problem can be much larger than a universal constant.  We show
how our results precisely characterize the error of a class of
temporal difference learning methods for the policy evaluation problem
with linear function approximation, establishing their optimality.
\end{abstract}
\end{center}

\section{Introduction}
\label{sec:intro}

Linear fixed point equations over a Hilbert space, with the Euclidean
space being an important special case, arise in various contexts.
Depending on the application, such fixed point equations take
different names, including estimating equations, Bellman equations,
Poisson equations and inverse
systems~\cite{bertsekas2011temporal,krasnosel2012approximate,wooldridge2016introductory}.
More specifically, given a Hilbert space $\Xspace$, we consider a
fixed point equation of the form
\begin{align}
\label{EqnFixedPoint}  
  \vvec & = \Lmat \vvec + \bvec,
\end{align}
where $\bvec$ is some member of the Hilbert space, and $\Lmat$ is a
linear operator mapping $\Xspace$ to itself.  

When the Hilbert space is infinite-dimensional---or has a finite but
very large dimension $\bigdim$---it is common to seek approximate
solutions to equation~\eqref{EqnFixedPoint}.  A standard approach is
to choose a subspace $\LinSpace$ of the Hilbert space, of dimension
$\usedim \ll \bigdim$, and to search for solutions within this
subspace.  In particular, letting $\ProjLin$ denote the orthogonal
projection onto this subspace, various methods seek (approximate)
solutions to the \emph{projected fixed point equation}
\begin{align}
\label{EqnProjFixedPoint}  
\vvec & = \ProjLin \big( \Lmat \vvec + \bvec \big).
\end{align}
In order to set the stage, let us consider some generic examples that
illustrate the projected fixed point
equation~\eqref{EqnProjFixedPoint}. We eschew a fully rigorous
exposition at this stage, deferring technical details and specific
examples to Section~\ref{SecExamples}.

\begin{example}[Galerkin methods for differential equations]

Let $\Xspace$ be a Hilbert space of suitably differentiable functions,
and let $\Amat$ be a linear differential operator of order $k$, say of
the form $\Amat(\vvec) = \omega_0 \vvec + \sum_{j=1}^k \omega_j
\vvec^{(j)}$, where $\vvec^{(j)}$ denotes the $j^{th}$-order
derivative of the function $\vvec \in \Xspace$. Given a function $\bvec \in \Xspace$, suppose that we are
interested in solving the differential equation $\Amat(\vvec) = \bvec$.
This represents a particular case of our fixed point equation with
$\Lmat = \IdMat - \Amat$.
  
Let $\LinSpace$ be a finite-dimensional subspace of $\Xspace$, say
spanned by a set of basis functions $\{\phi_j\}_{j=1}^\usedim$.  A
Galerkin method constructs an approximate solution to the differential
equation $\Amat(\vvec) = \bvec$ by solving the projected fixed point
equation~\eqref{EqnProjFixedPoint} over a subspace of this type.
Concretely, any function $\vvec \in \LinSpace$ has a representation of
the form $\vvec = \sum_{j=1}^\usedim \vartheta_j \phi_j$ for some
weight vector $\vartheta \in \real^\usedim$.  Applying the operator
$\Amat$ to any such function yields the residual
$\Amat(\vvec) = \sum_{j=1}^\usedim \vartheta_j \Amat(\phi_j)$,
and the Galerkin method chooses the weight vector $\vartheta \in
\real^\usedim$ such that $\vvec$ satisfies the equation $\vvec =
\ProjLin( (I - \Amat) \vvec + \bfunc)$.  A specific version of the Galerkin
method for a second-order differential equation called the
\emph{elliptic boundary value problem} is presented in detail in
Section~\ref{subsubsec:elliptic-example}.  \hfill \goodendex
\end{example}

\smallskip

\begin{example}[Instrumental variable methods for nonparametric regression] 
Let $\Xspace$ denote a suitably constrained space of square-integrable
functions mapping $\real^p \to \real$, and suppose that we have a
regression model of the form\footnote{For a more detailed discussion
of existence and uniqueness of the various objects in this model, see
Darolles et al.~\cite{darolles2011nonparametric}.}  $Y = f^*(X) +
\epsilon$. Here $X$ is a random vector of covariates taking values in
$\real^p$, the pair $(Y, \epsilon)$ denote scalar random variables,
and $f^* \in \Xspace$ denotes an unknown function of interest.  In the
classical setup of nonparametric regression, it is assumed that
$\Exs[\epsilon \mid X] = 0$, an assumption that can be violated.
Instead, suppose that we have a vector of \emph{instrumental
variables} $Z \in \real^p$ such that $\Exs[\epsilon \mid Z] = 0$.  Now
let $T: \Xspace \to \Xspace$ denote a linear operator given by $T(f) =
\Exs [ f(X) | Z ]$, and denote by $r = \Exs[Y | Z]$ a point in
$\Xspace$.  Instrumental variable approaches to estimating $f^*$ are
based on the equality
\begin{align} \label{fp-IV}
\Exs[Y - f^*(X) \mid Z] \; = \; r - T (f^*) = 0,
\end{align}
which is a linear fixed point relation of the
form~\eqref{EqnFixedPoint} with $L = I - T$ and $b = r$.

Now let $\{\phi_j\}_{j \geq 1}$ be an orthonormal basis of $\Xspace$,
and let $\lspace$ denote the subspace spanned by the first $d$ such
eigenfunctions. Then each function $f \in \lspace$ can be represented
as $f = \sum_{j = 1}^d \vartheta_j \phi_j$, and approximate solutions
to the fixed point equation~\eqref{fp-IV} may be obtained via solving
a projected variant~\eqref{EqnProjFixedPoint}, i.e., the equation $f =
\projecttolin((I - T)f + r)$.

A specific example of an instrumental variables method is the class of
temporal difference methods for policy evaluation, introduced and
discussed in detail in Section~\ref{ex:td-discounted}.  \hfill
\goodendex
\end{example}

\vspace*{0.25in}

In particular instantiations of both of the examples above, it is
typical for the ambient dimension $D$ to be very large (if not
infinite) and for us to only have sample access to the pair $(\Lmat,
\bvec)$. This paper\footnote{Especially for applications in
reinforcement learning, another natural setting is that of Markov
noise, which we handle in a companion paper.} treats the setting in
which $n$ observations $\left\{ (\Lmat_i, b_i) \right\}_{i = 1}^n$ are
drawn $\mathrm{i.i.d.}$ from some distribution with mean $(\Lmat,
\bvec)$.  Letting $\vstar$ denote the solution to the fixed point
equation~\eqref{EqnFixedPoint}, our goal is to use these observations
in order to produce an estimate $\vhat_n$ of $\vstar$ that satisfies
an \emph{oracle inequality} of the form
\begin{align}
    \Exs \statnorm{\vhat_n - \vstar}^2 \leq \alpha
    \cdot \inf_{\vvec \in \lspace} \statnorm{\vvec - \vstar}^2
    + \varepsilon_n. \label{eq:oracle-inequality-objective-genral}
\end{align}
Here we use $\| \cdot \|$ to denote the Hilbert norm associated with
$\Xspace$.  The three terms appearing on the RHS of
inequality~\eqref{eq:oracle-inequality-objective-genral} all have
concrete interpretations. The term
\begin{align}
\label{EqnDefnAppErr}  
\AppErr (\LinSpace, \vstar) \mydefn \inf_{\vvec \in \LinSpace}
\statnorm{\vvec - \vstar}^2
\end{align}
defines the \emph{approximation error}; this is the error incurred by
an oracle procedure that knows the fixed point $\vstar$ in advance and
aims to output the best approximation to $\vstar$ within the
subspace~$\lspace$.  The term $\alpha$ is the \emph{approximation
factor}, which indicates how poorly the estimator $\vhat_n$ performs
at carrying out the aforementioned approximation; note that $\alpha
\geq 1$ by definition, and it is most desirable for $\alpha$ to be as
small as possible. The final term $\varepsilon_n$ is a proxy for the
\emph{statistical error} incurred due to our stochastic observation
model; indeed, one expects that as the sample size $n$ goes to
infinity, this error should tend to zero for any reasonable estimator,
indicating consistent estimation when $\vstar \in \lspace$. More
generally, we would like our estimator to also have as small a
statistical error as possible in terms of the other parameters that
define the problem instance.

In an ideal world, we would like both desiderata to hold
simultaneously: the approximation factor should be as close to one as
possible while the statistical error stays as small as possible.  As
we discuss shortly, such a ``best-of-both-worlds'' guarantee can
indeed be obtained in many canonical problems, and ``sharp'' oracle
inequalities---meaning ones in which the approximation factor is equal
to $1$---are known~\cite{rakhlin2017empirical,dalalyan2012sharp}.  On
the other hand, such oracle equalities with unit factors are not known
for the fixed point equation~\eqref{EqnFixedPoint}. Tsitsiklis and Van
Roy~\cite{tsitsiklis1997analysis} show that if the operator $\Lmat$ is
$\contraction_{\max}$-contractive in the norm $\| \cdot \|$, then the
(deterministic) solution $\vbar$ to the projected fixed point
equation~\eqref{EqnProjFixedPoint} satisfies the bound
\begin{align}
\label{eq:contraction-approx-factor-worst-case}  
\statnorm{\vbar - \vstar}^2 \leq \frac{1}{1 -
  \conmax^2}
\inf_{\vvec \in \LinSpace}
\statnorm{\vvec - \vstar}^2.
\end{align}
The bound~\eqref{eq:contraction-approx-factor-worst-case} has a
potentially large approximation factor that can be quite far from one
(as would be the case for a ``sharp'' oracle inequality).  One
motivating question for our work is whether or not this bound can be
improved, and if so, to what extent.\footnote{Note that one can
achieve an approximation factor arbitrarily close to one provided that
$\numobs \gg \bigdim$.  One way to do so is as follows: form the
plug-in estimate that solves the original fixed point
relation~\eqref{EqnFixedPoint} on the sample averages
$\tfrac{1}{\numobs} \sum_{i = 1}^n \Lmat_i$ and $\tfrac{1}{\numobs}
\sum_{i = 1}^n \bvec_i$, and then project this solution onto the
subspace $\lspace$. In this paper, our principal interest---driven by
the practical examples of Galerkin approximation and temporal
difference learning---is in the regime $\usedim \ll \numobs \ll
\bigdim$.}

Our work is also driven by the complementary question of whether a
sharp bound can be obtained on the statistical error of an estimator
that, unlike $\vbar$, has access only to the samples $\left\{(\Lmat_i,
\bvec_i)\right\}_{i = 1}^n$.  In particular, we would like the
statistical error $\varepsilon_\numobs$ to depend on some notion of
complexity within the subspace $\lspace$, and \emph{not} on the
ambient space.  Recent work by Bhandari et
al.~\cite{bhandari2018finite} provides worst-case bounds on the
statistical error of a stochastic approximation scheme, showing that
the \emph{parametric rate} $\epsilon_n \lesssim d / n$ is
attainable. In this paper, we study how to derive a more fine-grained
bound on the statistical error that reflects the practical performance
of the algorithm and depends optimally on the geometry of our problem
instance.


\subsection{Contributions and organization}

The main contribution of this paper is to resolve both of the
aforementioned questions, in particular by deriving upper bounds and
information-theoretic lower bounds on both the approximation factor
and statistical error that are \emph{instance-dependent}.

On one hand, these bounds demonstrate that in most cases, the optimal
oracle inequality no longer has approximation factor $1$ in our
general setting, but on the other hand, that the optimal approximation
factor can be much better in many cases than what is suggested by the
worst-case bound~\eqref{eq:contraction-approx-factor-worst-case}. We
also derive a significantly sharper bound on the statistical error of
a stochastic approximation scheme that is instance-optimal in a
precise sense.  In more detail, we present the following results:
\bcar
\item Theorem~\ref{thm:linear-oracle-ineq} establishes an
  instance-dependent risk upper bound of the
  form~\eqref{eq:oracle-inequality-objective-genral} for the
  Polyak--Ruppert averaged stochastic approximation estimator, whose
  approximation factor $\alpha$ depends in a precise way on the
  projection of the operator $\Lmat$ onto the subspace $\lspace$, and
  the statistical error $\epsilon_n$ matches the Cram\'{e}r--Rao lower
  bound for the instance within the subspace.
\item In Theorem~\ref{thm:linear-lower-bound}, we prove an
  information-theoretic lower bound on the approximation factor.  It
  is a local analysis, in that the bound depends critically on the
  projection of the population-level operator.  This lower bound
  certifies that the approximation factor attained by our estimator is
  optimal. To the best of our knowledge, this is also the first
  instance of an optimal oracle inequality with a non-constant and
  problem-dependent approximation factor.
\item In Theorem~\ref{thm:linear-stat-error-lower-bound}, we establish
  via a Bayesian Cram\'{e}r-Rao lower bound that the leading
  statistical error term for our estimator is also optimal in an
  instance-dependent sense.
\item In Section~\ref{SecConsequences}, we derive specific
  consequences of our results for several examples, including for the
  problem of Galerkin approximation in second-order elliptic equations
  and temporal difference methods for policy evaluation with linear
  function approximation. A particular consequence of our results
  shows that in a minimax sense, the approximation
  factor~\eqref{eq:contraction-approx-factor-worst-case} is optimal
  for policy evaluation with linear function approximation
  (cf. Proposition~\ref{prop:mrp-approx-factor-lb}).

  \ecar

  \vspace*{0.1in}

The remainder of this paper is organized as follows.
Section~\ref{sec:related-work} contains a detailed discussion of
related work. We introduce formal background and specific examples in
Section~\ref{SecBackground}. Our main results under the general model
of projected fixed point equations are introduced and discussed in
Section~\ref{sec:main-results}. We then specialize these results to
our examples in Section~\ref{SecConsequences}, deriving several
concrete corollaries for Galerkin methods and temporal difference
methods. Our proofs are postponed to Section~\ref{SecProofs}, and
technical results are deferred to the appendix.


\subsection{Related work} \label{sec:related-work}

Our paper touches on various lines of related work, including oracle
inequalities for statistical estimation, stochastic approximation and
its application to reinforcement learning, and projected linear
equation methods.  We provide a brief discussion of these connections
here.

\paragraph{Oracle inequalities:} There is a large literature on
misspecified statistical models and oracle inequalities (e.g., see
the monographs~\cite{massart2007concentration,koltchinskii2011oracle}
for overviews).  Oracle inequalities in the context of penalized
empirical risk minimization (ERM) are quite well-understood
(e.g.,~\cite{bartlett2005local,koltchinskii2006local,massart2006risk}). Typically,
the resulting approximation factor is exactly $1$ or arbitrarily close
to $1$, and the statistical error term depends on the localized
Rademacher complexity or metric entropy of this function class.
Aggregation methods have been developed in order to obtain
\emph{sharp} oracle inequalities with approximation factor exactly $1$
(e.g.~\cite{tsybakov2004optimal,bunea2007aggregation,dalalyan2012sharp,rakhlin2017empirical}).
Sharp oracle inequalities are now available in a variety of settings
including for sparse linear models~\cite{bunea2007sparsity}, density
estimation~\cite{dalalyan2018optimal}, graphon
estimation~\cite{klopp2017oracle}, and shape-constrained
estimation~\cite{bellec2018sharp}. As previously noted, our setting
differs qualitatively from the ERM setting, in that as shown in this
paper, sharp oracle inequalities are no longer possible.  There is
another related line of work on oracle inequalities of density
estimation. Yatracos~\cite{yatracos1985rates} showed an oracle
inequality with the non-standard approximation factor $3$, and with a
statistical error term depending on the metric entropy.  This non-unit
approximation factor was later shown to be optimal for the class of
one-dimensional piecewise constant
densities~\cite{chan2014near,bousquet2019optimal,zhu2020deconstructing}. The
approximation factor lower bound in these papers and our work both
make use of the birthday paradox to establish information-theoretic
lower bounds.

\paragraph{Stochastic approximation:} Stochastic approximation algorithms for linear
and nonlinear fixed-point equations have played a central role in
large-scale machine learning and
statistics~\cite{robbins1951stochastic,lai2003stochastic,nemirovski2009robust}. See
the books~\cite{benveniste2012adaptive,borkar2009stochastic} for a
comprehensive survey of the classical methods of analysis. The seminal
works by Polyak, Ruppert, and
Juditsky~\cite{polyak1990new,polyak1992acceleration,ruppert1988efficient}
propose taking the average of the stochastic approximation iterates,
which stabilizes the algorithm and achieves a Gaussian limiting
distribution. This asymptotic result is also known to achieve the
local asymptotic minimax lower
bound~\cite{duchi2016asymptotic}. Non-asymptotic guarantees matching
this asymptotic behavior have also been established for stochastic
approximation algorithms and their variance-reduced
variants~\cite{moulines2011non,khamaru2020temporal,mou2020linear,li2020root}.

Stochastic approximation is also a fundamental building block for
reinforcement learning algorithms, wherein the method is used to
produce an iterative, online solution to the Bellman equation from
data; see the
books~\cite{szepesvari2010algorithms,bertsekas2019reinforcement} for a
survey. Such approaches include temporal difference (TD)
methods~\cite{sutton1988learning} for the policy evaluation problem
and the $Q$-learning algorithm~\cite{watkins1992q} for policy
optimization. Variants of these algorithms also abound, including
LSTD~\cite{boyan2002technical}, SARSA~\cite{rummery1994line},
actor-critic algorithms~\cite{konda2000actor}, and gradient TD
methods~\cite{sutton2009fast}. The analysis of these methods has
received significant attention in the literature, ranging from
asymptotic guarantees
(e.g.,~\cite{bradtke1996linear,tsitsiklis1997analysis,tsitsiklis1999optimal})
to more fine-grained finite-sample bounds
(e.g.,~\cite{bhandari2018finite,srikant2019finite,lakshminarayanan2018linear,pananjady2020instance,wainwright2019stochastic,wainwright2019variance}). Our
work contributes to this literature by establishing finite-sample
upper bounds for temporal difference methods with Polyak--Ruppert
averaging, as applied to the policy evaluation problem with linear
function approximation.

\paragraph{Projected methods for linear equations:}
Galerkin~\cite{galerkin1915series} first proposed the method of
approximating the solution to a linear PDE by solving the projected
equation in a finite-dimensional subspace. This method later became a
cornerstone of finite-element methods in numerical methods for PDEs;
see the books~\cite{fletcher1984computational,brenner2007mathematical}
for a comprehensive survey. A fundamental tool used in the analysis of
Galerkin methods is C\'{e}a's lemma~\cite{cea1964approximation}, which
corresponds to a special case of the approximation factor upper bounds
that we establish. As mentioned before, projected linear equations
were also considered independently by Tsitsiklis and Van
Roy~\cite{tsitsiklis1997analysis} in the context of reinforcement
learning; they established the worst-case upper
bound~\eqref{eq:contraction-approx-factor-worst-case} on the
approximation factor under contractivity assumptions. These
contraction-based bounds were further extended to the analysis of
$Q$-learning in optimal stopping
problems~\cite{tsitsiklis1999optimal}. The connection between the
Galerkin method and TD methods was discovered by Yu and
Bertsekas~\cite{yu2010error,bertsekas2011temporal}, and the former
paper shows an instance-dependent upper bound on the approximation
factor. This analysis was later applied to Monte--Carlo methods for
solving linear inverse
problems~\cite{polydorides2009approximate,polydorides2012quasi}.

We note that the Bellman equation can be written in infinitely many
equivalent ways---by using powers of the transition kernel and via the
formalism of resolvents---leading to a continuous family of projected
equations indexed by a scalar parameter $\lambda$ (see, e.g., Section
5.5 of Bertsekas~\cite{bertsekas2019reinforcement}). Some of these
forms can be specifically leveraged in other observation models; for
instance, by observing the trajectory of the Markov chain instead of
i.i.d. samples, it becomes possible to obtain unbiased observations
for integer powers of the transition kernel. This makes it possible to
efficiently estimate the solution to the projected linear equation for
various values of $\lambda$, and underlies the family of TD$(\lambda)$
methods~\cite{sutton1988learning,boyan2002technical}. Indeed,
Tsitsiklis and Van Roy~\cite{tsitsiklis1997analysis} also showed that
the worst-case approximation factor in
equation~\eqref{eq:contraction-approx-factor-worst-case} can be
improved by using larger values of $\lambda$.  Based on this
observation, a line of work has studied the trade-off between
approximation error and estimation measure in model selection for
reinforcement learning
problems~\cite{bertsekas2016proximal,scherrer2010should,munos2008finite,van2006performance}.
However, unlike this body of work, our focus in the current paper is
on studying the i.i.d. observation model; we postpone a detailed
investigation of the Markov setting to a companion paper.


\subsection{Notation} 

Here we summarize some notation used throughout the paper. For a
positive integer $m$, we define the set $[m] \defn \{1,2, \cdots,
m\}$.  For any pair $(\Xspace, \Yspace)$ of real Hilbert spaces and a
linear operator $A: \Xspace \rightarrow \Yspace$, we denote by
$\adjoint{A}: \Yspace \rightarrow \Xspace$ the adjoint operator of
$A$, which by definition, satisfies $\inprod{A x}{y} =
\inprod{x}{\adjoint{A} y}$ for all $(x,y) \in \Xspace \times \Yspace$.
For a bounded linear operator $A$ from $\Xspace$ to $\Yspace$, we
define its operator norm as: $\matsnorm{A}{\Xspace \rightarrow \Yspace
} \mydefn \sup_{x \in \Xspace \setminus \{0 \} }
\tfrac{\vecnorm{Ax}{\Yspace}}{\vecnorm{x}{\Xspace}}$.  We use the
shorthand notation $\hilopnorm{A}$ to denote its operator norm when
$A$ is a bounded linear operator mapping $\Xspace$ to itself.  When
$\Xspace = \real^{d_1}$ and $\Yspace = \real^{d_2}$ are
finite-dimensional Euclidean spaces equipped with the standard inner
product, we denote by $\opnorm{A}$ the operator norm in this case. We
also use $\vecnorm{\cdot}{2}$ to denote the standard Euclidean norm,
in order to distinguish it from the Hilbert norm $\statnorm{\cdot}$.

For a random object $X$, we use $\law(X)$ to denote its probability
law. Given a vector $\mu \in \real^d$ and a positive semi-definite
matrix $\Sigma \in \real^{d \times d}$, we use $\mathcal{N} (\mu,
\Sigma)$ to denote the Gaussian distribution with mean $\mu$ and
covariance $\Sigma$. We use $\mathcal{U} (\Omega)$ to
denote the uniform distribution over a set $\Omega$.  Given a Polish
space $\mathcal{S}$ and a positive measure $\mu$ associated to its
Borel $\sigma$-algebra, for $p \in [1, + \infty)$, we define
  $\mathbb{L}^p (\mathcal{S}, \mu) \mydefn \big \{ f: \mathcal{S}
  \rightarrow \real,~ \vecnorm{f}{\mathbb{L}^p} \mydefn \left(
  \int_\mathcal{S} |f|^p d \mu \right)^{1/p} < + \infty \big \}$.
  When $\mathcal{S}$ is a subset of $\real^d$ and $\mu$ is the
  Lebesgue measure, we use the shorthand notation $\mathbb{L}^p
  (\mathcal{S})$.  For a point $x \in \real^d$, we use $\delta_x$ to
  denote the Dirac $\delta$-function at point $x$.

We use $\{e_j\}_{j=1}^\usedim$ to denote the standard basis vectors in
the Euclidean space $\real^\usedim$, i.e., $e_i$ is a vector with a
$1$ in the $i$-th coordinate and zeros elsewhere.  For two matrices $A
\in \real^{d_1 \times d_2}$ and $B \in \real^{d_3 \times d_4}$, we
denote by $A \otimes B$ their Kronecker product, a $d_1 d_3 \times d_2
d_4$ real matrix. For symmetric matrices $A, B \in \real^{d \times
  d}$, we use $A \preceq B$ to denote the fact $B - A$ is a positive
semi-definite matrix, and denote by $A \prec B$ when $B - A$ is
positive definite.  For a positive integer $d$ and indices $i, j \in
[d]$, we denote by $E_{ij}$ a $d \times d$ matrix with a $1$ in the
$(i, j)$ position and zeros elsewhere. More generally, given a set
$\statespace$ and $s_1, s_2, \in \statespace$, we define $E_{s_1,
  s_2}$ to be the linear operator such that $E_{s_1, s_2} f (x)
\mydefn f (s_2) \bm{1}_{x = s_1}$ \mbox{for all $f: \statespace
  \rightarrow \real$.}


\section{Background}
\label{SecBackground}

We begin by formulating the projected fixed point problem more
precisely in Section~\ref{sec:formulation}.  Section~\ref{SecExamples}
provides illustrations of this general set-up with some concrete
examples.


\subsection{Problem formulation}
\label{sec:formulation}

Consider a separable Hilbert space $\Xspace$ with (possibly infinite)
dimension $\bigdim$, equipped with the inner product
$\statinprod{\cdot}{\cdot}$. Let $\AClass$ denote the set of all
bounded linear operators mapping $\Xspace$ to itself. Given one such
operator $\Lmat \in \AClass$ and some $\bvec \in \Xspace$, we consider
the fixed point relation $\vvec = \Lmat \vvec + \bvec$, as previously
defined in equation~\eqref{EqnFixedPoint}.  We assume that the
operator $I - \Lmat$ has a bounded inverse, which guarantees the
existence and uniqueness of the fixed point satisfying
equation~\eqref{EqnFixedPoint}.  We let $\vstar$ denote this unique
solution.

As previously noted, in general, solving a fixed point equation in the
Hilbert space can be computationally challenging.  Consequently, a
natural approach is to seek approximations to the fixed point $\vstar$
based on searching over a finite-dimensional subspace of the full
Hilbert space.  More precisely, given some $\usedim$-dimensional
subspace $\LinSpace$ of $\Xspace$, we seek to solve the projected
fixed point equation~\eqref{EqnProjFixedPoint}.

\paragraph{Existence and uniqueness of projected fixed point:}

For concreteness in analysis, we are interested in problems for which
the projected fixed equation has a unique solution.  Here we provide a
sufficient condition for such existence and uniqueness.  In doing so
and for future reference, it is helpful to define some mappings
between $\Xspace$ and the subspace $\LinSpace$.  Let us fix some
orthogonal basis $\{\phi_j\}_{j \geq 1}$ of the full space $\Xspace$
such that $\LinSpace = \myspan \{ \phi_1, \ldots, \phi_\usedim \}$.
In terms of this basis, we can define the projection operator
$\PhiOp: \Xspace \rightarrow \real^\usedim$ via $\PhiOp(x) \defn \big(
\statinprod{x}{\phi_j} \big)_{j= 1}^\usedim$.  The adjoint operator of
$\PhiOp$ is a mapping from $\real^\usedim$ to $\Xspace$, given by
\begin{align}
\label{EqnDefnAdjoint}
\PhiOp(v) & \defn \sum_{j = 1}^\usedim v_j \phi_j.
\end{align}
Using these operators, we can define the \emph{projected operator}
associated with $\Lmat$---namely
\begin{align}
  \label{EqnDefnSpecMat}
\SpecMat \mydefn \PhiOp \Lmat \adjoint{\PhiOp}.
\end{align}
Note that $\SpecMat$ is simply a $\usedim$-dimensional matrix, one
which describes the action of $\Lmat$ on $\LinSpace$ according to the
basis that we have chosen. As we will see in the main theorems, our
results do not depend on the specific choice of the orthonormal basis,
but it is convenient to use a given one, as we have done here.

Consider the quantity
\begin{align}
\kappa(\SpecMat) & \defn \tfrac{1}{2} \lammax \Big( \SpecMat +
\SpecMat^\top \Big),
  \end{align}
corresponding to the maximal eigenvalue of the symmetrized version of
$\SpecMat$.  One sufficient condition for there be a unique solution
to the fixed point equation~\eqref{EqnProjFixedPoint} is the bound
$\kappa (\SpecMat) < 1$.  When this bound holds, the matrix
$(I_\usedim - \SpecMat)$ is invertible, and hence for any $\bvec
\in \Xspace$, there is a unique solution $\vbar$ to the equation
$\vvec = \ProjLin (\Lmat \vvec + \bvec)$.

\paragraph{Stochastic observation model:}

As noted in the introduction, this paper focuses on an observation
model in which we observe i.i.d. random pairs $(\Lmat_i, \bvec_i)$ for
$i=1,\ldots, \numobs$ that are unbiased estimates of the pair $(\Lmat,
\bvec)$ so that
\begin{align}
\label{EqnUnbiased}
  \Exs[\Lmat_i] = \Lmat, \quad \mbox{and} \quad \Exs[\bvec_i] = \bvec.
\end{align}
In addition to this unbiasedness, we also assume that our observations satisfy a certain second-moment bound. 
A weaker and a stronger version of this assumption are both considered.
\begin{assumption1w}[Second-moment bound in projected space]
\namedlabel{assume-second-moment}{1(W)}
There exist scalars \mbox{$\varA, \varb > 0$} such that for any unit-norm
vector $u \in \LinSpace$ and any basis vector in $\{ \phi_j
\}_{j=1}^\usedim$ we have the bounds
\begin{subequations}
  \begin{align}
\label{eq:assume-second-moment-L-noise}     
\Exs \statinprod{\phi_j}{(\Lmat_i - \Lmat) u}^2 & \leq \sigmaA^2
\statnorm{u}^2, \quad \mbox{and} \\
\label{eq:assume-second-moment-b-noise}
\Exs \statinprod{\phi_j}{\bvec_i - \bvec}^2 & \leq \sigmab^2.
\end{align}
\end{subequations}
\end{assumption1w}

\begin{assumption1s}[Second-moment bound in ambient space]
\namedlabel{assume-second-moment-strong}{1(S)}
There exist scalars $\varA, \varb > 0$ such that for any unit-norm
vector $u \in \Xspace$ and any basis vector in $\{ \phi_j
\}_{j=1}^\bigdim$ we have the bounds
\begin{subequations}
  \begin{align}
\label{eq:assume-second-moment-L-noise-strong}     
\Exs \statinprod{\phi_j}{(\Lmat_i - \Lmat) u}^2 & \leq \sigmaA^2
\statnorm{u}^2, \quad \mbox{and} \\
\label{eq:assume-second-moment-b-noise-strong}
\Exs \statinprod{\phi_j}{\bvec_i - \bvec}^2 & \leq \sigmab^2.
\end{align}
\end{subequations}
\end{assumption1s}
\noindent In words, Assumption~\ref{assume-second-moment} guarantees
that the random variable obtained by projecting the ``noise'' onto any
of the basis vectors $\phi_1, \ldots, \phi_\usedim$ in the subspace
$\LinSpace$ has bounded second
moment. Assumption~\ref{assume-second-moment-strong} further requires
the projected noise onto any basis vector of the entire space
$\Xspace$ to have bounded second moment. In
Section~\ref{SecConsequences}, we show that there are various
settings---including Galerkin methods and temporal difference
methods---for which at least one of these assumptions is satisfied.

\subsection{Examples}
\label{SecExamples}

We now present some concrete examples to illustrate our general
formulation.  In particular, we discuss the problems of linear
regression, temporal difference learning methods from reinforcement
learning\footnote{As noted by Bradtke and
Barto~\cite{bradtke1996linear}, this method can be understood as an
instrumental variable method~\cite{wooldridge2016introductory}, and
our results also apply to this more general setting.}, and Galerkin
methods for solving partial differential equations.


\subsubsection{Linear regression on a low-dimensional subspace}
\label{example:linear-regression}

Our first example is the linear regression model when true parameter
is known to lie approximately in a low-dimensional subspace. This
example, while rather simple, provides a useful pedagogical starting
point for the others to follow.

For this example, the underlying Hilbert space $\Xspace$ from our
general formulation is simply the Euclidean space $\real^\bigdim$,
equipped with the standard inner product $\langle \cdot, \cdot
\rangle$.  We consider zero-mean covariates $X \in \real^\bigdim$ and
a response $Y \in \real$, and our goal is to estimate the best-fitting
linear model $x \mapsto \inprod{v}{x}$.  In particular, the
mean-square optimal fit is given by $\vstar \defn \arg \min_{v \in
  \real^\bigdim} \big(Y - \inprod{v}{X})^2$.  From standard results on
linear regression, this vector must satisfy the normal equations
$\Exs[ X X^\top] \vstar = \Exs[Y X]$.  We assume that the
second-moment matrix $\Exs[X X^\top]$ is non-singular, so that
$\vstar$ is unique.

Let us rewrite the normal equations in a form consistent with our
problem formulation.  An equivalent definition of $\vstar$ is in terms
of the fixed point relation
\begin{align}
\label{eq:fp-linear}
\vstar = \left( I - \frac{1}{\smooth} \Exs [X X^\top] \right) \vstar +
\frac{1}{\smooth} \Exs [Y X],
\end{align}
where $\smooth \mydefn \lambda_{\max} (\Exs [XX^\top])$ is the maximum
eigenvalue.  This fixed point condition is a special case of our
general equation~\eqref{EqnFixedPoint} with the operator $\Lmat = I -
\frac{1}{\smooth} \Exs [XX^\top]$ and vector $\bvec =
\frac{1}{\smooth} \Exs [Y X]$. Note that we have
\begin{align*}
    \opnorm{\Lmat} = \opnorm{I - \frac{1}{\smooth} \Exs [XX^\top]}
    \leq 1 - \frac{\strongconvex}{\smooth} < 1,
\end{align*}
where $\strongconvex = \lammin(\Exs[X X^\top]) > 0$ is the minimum
eigenvalue of the covariance matrix.

In the well-specified setting of linear regression, we observe i.i.d.
pairs $(X_i, Y_i) \in \real^\bigdim \times \real$ that are linked by
the standard linear model
\begin{align}
\label{eq:linear-regression-model}  
Y_i = \inprod{\vstar}{X_i} + \varepsilon_i, \quad \mbox{for $i = 1,2,
  \cdots, \numobs$,}
\end{align}
where $\varepsilon_i$ denotes zero-mean noise with finite second
moment.  Each such observation can be used to form the matrix-vector
pair
\begin{align*}
\Lmat_i = I - \smooth^{-1} X_i X_i^\top, \quad \mbox{and} \quad \bvec_i =
\smooth^{-1} X_i Y_i,
\end{align*}
which is in the form of our assumed observation model.

Thus far, we have simply reformulated linear regression as a fixed
point problem.  In order to bring in the projected aspect of the
problem, let us suppose that the ambient dimension $\bigdim$ is much
larger than the sample size $\numobs$, but that we have the prior
knowledge that $\vstar$ lies (approximately) within a known subspace
$\lspace$ of $\real^\bigdim$, say of dimension $\usedim \ll \bigdim$.
Our goal is then to approximate the solution to the associated
projected fixed-point equation.

Using $\{\phi_j\}_{j=1}^\usedim$ to denote an orthonormal basis of
$\lspace$, the population-level projected linear
equation~\eqref{EqnProjFixedPoint} in this case takes the form
\begin{align}
\label{eq:linear-regression-projected-popultion-problem}  
\Exs \left[ (\projecttolin X) (\projecttolin X)^\top \right] \vbar =
\Exs \left[ Y \cdot \projecttolin X \right],
\end{align}
Thus, the population-level projected
problem~\eqref{eq:linear-regression-projected-popultion-problem}
corresponds to performing linear regression using the projected
version of the covariates, thereby obtaining a vector of weights
$\vbar \in \lspace$ in this low-dimensional space.


\subsubsection{Galerkin methods for second-order elliptic equations}
\label{subsubsec:elliptic-example}

We now turn to the Galerkin method for solving differential equations,
a technique briefly introduced in Section~\ref{sec:intro}.  The
general problem is to compute an approximate solution to a partial
differential equation based on a limited number of noisy observations
for the coefficients.  Stochastic inverse problems of this type arise
in various scientific and engineering
applications~\cite{nickl2017bayesian,arridge2019solving}.

For concreteness, we consider a second-order elliptic equation with
Dirichlet boundary conditions.\footnote{It should be noted that
Galerkin methods apply to a broader class of problems, including
linear PDEs of parabolic and hyperbolic
type~\cite{larsson2008partial}, as well as kernel integral
equations~\cite{polydorides2009approximate,polydorides2012quasi}.}
Given a bounded, connected and open set $\domain \subseteq
\real^\domaindim$ with unit Lebesgue measure, let $\partial \domain$
denote its boundary.  Consider the Hilbert space of functions
\begin{align*}
  \Xspace \mydefn \left\{ \vvec : \domain \rightarrow \real, ~
  \int_\domain \vecnorm{\nabla \vvec (x)}{2}^2 dx < \infty, ~
  \vvec|_{\partial \domain} = 0 \right\}
\end{align*}
equipped with the inner product $\inprod{u}{v}_{\sobolevone} \mydefn
\int_\domain \nabla u (x)^\top \nabla v (x) dx$.

Given a symmetric matrix-valued function $\afunc$ and a
square-integrable function $\bfunc \in \Ltwospace$, the
\emph{boundary-value problem} is to find a function $\vvec: \domain
\rightarrow \real$ such that
\begin{align}
    \begin{cases}
    \nabla \cdot (\afunc (x) \nabla \vvec (x)) + \bfunc = 0 & \mbox{in
      $\domain$},\\ \vvec (x) = 0 & \mbox{on $\partial \domain$}.
    \end{cases} \label{eq:elliptic-boundary-value-problem}
\end{align}
We impose a form of uniform ellipticity by requiring that
$\strongconvex I_{\domaindim} \preceq \afunc (x) \preceq \smooth
I_{\domaindim}$, for some positive scalars $\strongconvex \leq
\smooth$, valid uniformly over $x$.

The problem can be equivalently stated in terms of the elliptic
operator $\Amat \mydefn - \nabla \cdot (\afunc \nabla)$; as shown in
Appendix~\ref{Appendix:subsubsec-proof-of-elliptic-lemma}, the pair
$(\Amat, \bfunc)$ \emph{induces} a bounded, self-adjoint linear
operator $\widetilde{A}$ on $\Xspace$ and a function $g \in \Xspace$
such that the solution to the boundary value problem can be written as
\begin{align}
\label{eq:elliptic-linear-fixed-pt}  
\vstar = \left( I - \frac{1}{\smooth} \widetilde{\Amat} \right) \vstar
+ \smooth^{-1} g.
\end{align}
By construction, this is now an instance of our general fixed point
equation~\eqref{EqnFixedPoint} with $\Lmat \mydefn I -
\frac{1}{\smooth} \widetilde{\Amat}$ and $\bvec \mydefn \smooth^{-1}
g$.  Furthermore, our assumptions imply that $\hilopnorm{\Lmat} \leq 1
- \frac{\strongconvex}{\smooth}$.

We consider a stochastic observation model that is standard in the
literature (see, e.g., the paper~\cite{giordano2020consistency}).
Independently for each $i \in [n]$, let $W_i$ denote an $\domaindim
\times \domaindim$ symmetric random matrix with entries on the
diagonal and upper-diagonal given by $\mathrm{i.i.d.}$ standard
Gaussian random variables. Let $w_i' \sim \mathcal{N} (0, 1)$ denote a
standard Gaussian random variable.  Suppose now that we observe the
pair $x_i, y_i \sim \mathcal{U}(\domain)$; the observed values for the
$i$-th sample are then given by
\begin{align} \label{eq:obs-Galerkin}
(\afunc_i, \bfunc_i) \mydefn \big( \afunc (x_i) + W_i, \bfunc (y_i) +
  w_i' \big) \quad \text{ with } \quad x_i, y_i \sim \mathcal{U}
  (\domain).
\end{align}
The unbiased observations $(\Lmat_i, \bvec_i)$ can then be constructed
by replacing $(\afunc, \bfunc)$ with $\big( \afunc_i \delta_{x_i},
\bfunc_i \delta_{y_i} \big)$ in the constructions above.

For such problems, the finite-dimensional projection not only serves
as a fast and cheap way to compute solutions from
simulation~\cite{lung2020sketched}, but also makes the solution stable
and robust to noise~\cite{kaltenbacher2011adaptive}.  Given a
finite-dimensional linear subspace $\lspace \subseteq \Xspace$ spanned
by orthogonal basis functions $(\phi_i)_{i = 1}^\usedim$, we consider
the projected version of equation~\eqref{eq:elliptic-linear-fixed-pt},
with solution denoted by $\vbar$:
\begin{align}
    \vbar = \projecttolin (\Lmat \vbar +
    \bvec). \label{eq:elliptic-projected-linear-fixed-pt}
\end{align}
Straightforward calculation in conjunction with
Lemma~\ref{lemma:elliptic-bounded-linear-operator} shows that
equation~\eqref{eq:elliptic-projected-linear-fixed-pt} is equivalent
to the conditions $\vbar \in \LinSpace$, and
\begin{align}
  \label{eq:galerkin-orthogonal-condition}
  \inprod{\widetilde{\Amat} \vbar}{\phi_j}_{\sobolevone} =
  \inprod{g}{\phi_j}_{\sobolevone} \quad \mbox{for all $j \in
    [\usedim]$,}
\end{align}
with the latter equality better known as the \emph{Galerkin
orthogonality condition} in the
literature~\cite{brenner2007mathematical}.


\subsubsection{Temporal difference methods for policy evaluation}
\label{ex:td-discounted}

Our final example involves the policy evaluation problem in
reinforcement learning. This is a special case of an instrumental
variable method, as briefly introduced in Section~\ref{sec:intro}.  We
require some additional terminology to describe the problem of policy
evaluation.  Consider a Markov chain on a state space $\statespace$
and a transition kernel $\transition: \statespace \times \statespace
\rightarrow \real$.  It becomes a discounted Markov reward process
when we introduce a reward function $\reward: \statespace \rightarrow
\real$, and discount factor $\discount \in (0, 1)$.  The goal of the
policy evaluation problem to estimate the value function, which is the
expected, long-term, discounted reward accrued by running the process.
The value function exists under mild assumptions such as boundedness
of the reward, and is given by the solution to the Bellman equation
$\valuestar = \discount \transition \valuestar + \reward$, which is a
fixed point equation of the form~\eqref{EqnFixedPoint} with $\Lmat =
\discount \transition$ and $\bvec = \reward$.

Throughout our discussion, we assume that the transition kernel
$\transition$ is ergodic and aperiodic, so that its stationary
distribution $\stationary$ is unique. We define $\Xspace$ to be the
Hilbert space $\Ltwospace (\statespace, \stationary)$, and for any pair of vectors $\vvec, \vvec' \in \Xspace$, we define the inner product as follows
\begin{align*}
     \statinprod{\vvec}{\vvec'}
    \mydefn \int_\statespace \vvec (s) \vvec' (s) d\stationary (s).
\end{align*}
In the special case of a finite state space, the Hilbert space
$\Xspace$ is a finite-dimensional Euclidean space with dimension
$\bigdim = |\statespace|$ and equipped with a weighted $\ell_2$-norm.

We consider the i.i.d. observation model in this paper\footnote{As
mentioned before, we undertake a more in-depth study of the case with
Markov observations, which is particularly relevant to the MRP
example, in a companion paper.}.  For each $i = 1,2, \cdots, n$,
suppose that we observe an independent tuple $(s_i, s_i^+, R_i
(s_i))$, such that
\begin{align}
\label{eq:mrp-iid-observation-model}  
  s_i \sim \stationary,~ s_i^+ \sim \transition (s_i, \cdot),
  ~\mbox{and}~ \Exs [R_i (s_i) | s_i] = \reward(s_i).
\end{align}
The $i$-th observation $(\Lmat_i, \bvec_i)$ is then obtained by
plugging in these observations to compute unbiased estimates of $P$
and $r$, respectively.

A common practice in reinforcement learning is to employ
\emph{function approximation}, which in its simplest form involves
solving a projected linear equation on a subspace. In particular,
consider a set $\{\psi_1, \psi_2, \cdots, \psi_d \}$ of basis
functions in $\Xspace$, and suppose that they are linearly independent
on the support of $\stationary$.  We are interested in projections
onto the subspace $\lspace = \myspan(\psi_1, \ldots, \psi_d)$, and in
solving the population-level projected fixed point
equation~\eqref{EqnProjFixedPoint}, which takes the form
\begin{align}
    \valuebar = \projecttolin (\discount \transition \valuebar +
    \reward). \label{eq:projected-bellman-discount}
\end{align}

The basis functions $\psi_i$ are not necessarily orthogonal, and it is
common for the projection operation to be carried out in a somewhat
non-standard fashion. In order to describe this, it is convenient to
write equation~\eqref{eq:projected-bellman-discount} in the projected
space. For each $s \in \mathcal{S}$, let $\psi(s) = [\psi_1(s) \;
  \psi_2(s) \; \ldots \; \psi_d(s)]$ denote a vector in $\real^d$, and
note that we may write $\valuebar(s) = \psi(s)^{\top} \bar{\vartheta}$
for a vector of coefficients $\bar{\vartheta} \in \real^d$.  Now
observe that equation~\eqref{eq:projected-bellman-discount} can be
equivalently written in terms of the coefficient vector
$\bar{\vartheta}$ as
\begin{align}
\label{eq:lstd-in-low-dimensional-space}  
    \Exs_{s \sim \stationary} [\psi (s) \psi (s)^\top] \bar{\vartheta}
    = \discount \Exs_{s \sim \stationary} \left[ \Exs_{s^+ \sim
        \transition (s, \cdot)} [\psi (s) \psi (s^+)^\top] \right]
    \bar{\vartheta} + \Exs_{s \sim \stationary} [r (s) \psi (s)].
\end{align}
Equation~\eqref{eq:lstd-in-low-dimensional-space} is the population
relation underlying the canonical \emph{least squares temporal
difference} (LSTD) learning
method~\cite{bradtke1996linear,boyan2002technical}.

\section{Main results for general projected linear equations} \label{sec:main-results}

Having set-up the problem and illustrated it with some examples, we
now turn to the statements of our main results.  We begin in
Section~\ref{SecUpper} by stating an upper bound on the mean-squared
error of a stochastic approximation scheme that uses Polyak--Ruppert
averaging.  We then discuss the form of this upper bound for various
classes of operator $\Lmat$, with a specific focus on producing
transparent bounds on the approximation factor.
Section~\ref{SecLower} is devoted to information-theoretic lower
bounds that establish the sharpness of our upper bound.


\subsection{Upper bounds}
\label{SecUpper}

In this section, we describe a standard stochastic approximation
scheme for the problem based on combining ordinary stochastic updates
with Polyak--Ruppert
averaging~\cite{polyak1990new,polyak1992acceleration,ruppert1988efficient}.
In particular, given an oracle that provides observations $(\Lmat_i,
\bvec_i)$, consider the stochastic recursion parameterized by a
positive stepsize $\stepsize$:
\begin{subequations}
  \label{eq:SA}
  \begin{align}
 \label{eq:lsa-iterates}    
    \vvec_{t + 1} = (1 - \stepsize) \vvec_t + \stepsize \projecttolin
    \big( \Lmat_{t + 1} \vvec_t + b_{t + 1} \big), \quad \mbox{for $t
      = 1, 2, \ldots$.}
  \end{align}
This is a standard stochastic approximation scheme for attempting to
solve the projected fixed point relation.  In order to improve it, we
use the standard device of applying Polyak--Ruppert averaging so as to
obtain our final estimate.  For a given sample size $\numobs \geq 2$,
our final estimate $\vhat_\numobs$ is given by taking the average of
these iterates from time $\numburn$ to $\numobs$---that is
\begin{align}
\label{eq:PR-average}  
\vhat_\numobs & \mydefn \frac{1}{\numobs - \numburn} \sum_{t =
  \numburn + 1}^\numobs \vvec_t.
\end{align}
\end{subequations}
Here the ``burn-in'' time $\numburn$ is an integer parameter to be
specified.

The stochastic approximation procedure~\eqref{eq:SA} is defined in the
entire space $\Xspace$; note that it can be equivalently written as
iterates in the projected space $\real^d$, via the recursion
\begin{align} \label{eq:projected-SA}
    \vartheta_{t + 1} = (1 - \stepsize) \vartheta_t + \stepsize
    (\PhiOp \Lmat_{t + 1} \adjoint{\PhiOp} \vartheta_t + \PhiOp b_{t +
      1}).
\end{align}
The original iterates can be recovered by applying the adjoint
operator---that is, $\vvec_t = \PhiOp^* \vartheta_t$ for $t = 1,2,
\ldots$.


\subsubsection{A finite-sample upper bound}

Having introduced the algorithm itself, we are now ready to provide a
guarantee on its error.  Two matrices play a key role in the statement
of our upper bound.  The first is the $\usedim$-dimensional matrix
$\SpecMat \mydefn \PhiOp \Lmat \adjoint{\PhiOp}$ that we introduced in
Section~\ref{sec:formulation}.  We show that the mean-squared error is
upper bounded by the approximation error $\inf_{v \in \lspace}
\statnorm{v - \vstar}^2$ along with a pre-factor of the form
\begin{align}
\label{EqnPrefactor}  
\prefact(\SpecMat, s) & = 1 + \lammax \Big( (I - \SpecMat)^{-1} (s^2
\, I_\usedim - \SpecMat \SpecMat^T ) (I - \SpecMat)^{-T} \Big),
\end{align}
for $s = \opnorm{\Lmat}$.  Our bounds also involve the quantity
$\kappa(\SpecMat) = \tfrac{1}{2} \lammax \big( \SpecMat + \SpecMat^T
\big)$, which we abbreviate by $\kappa$ when the underlying matrix
$\SpecMat$ is clear from the context.

The second matrix is a covariance matrix, capturing the noise
structure of our observations, given by
    \begin{align*}
        \SigStar \mydefn \cov \left( \PhiOp (\bvec_1 - \bvec) \right)
        + \cov \left( \PhiOp (\Lmat_1 - \Lmat) \vbar \right).
    \end{align*}
This matrix, along with the constants $(\sigmaA, \sigmab)$ from
Assumption~\ref{assume-second-moment}, arise in the definition of two
additional error terms, namely
\begin{subequations}
\begin{align}    
\BigEstErr & \defn \frac{\trace \left( (I - \SpecMat)^{-1} \SigStar (I
  - \SpecMat)^{- \top} \right)}{\numobs}, \mbox{and} \label{eq:def-big-est-err} \\
\BigResErr & \defn \frac{\sigmaA}{(1 - \kappa)^3} \left(
\frac{\usedim}{\numobs} \right)^{3/2} \left( \statnorm{\vbar}^2
\sigmaA^2 + \sigmab^2 \right). \label{eq:def-big-res-err}
\end{align}
\end{subequations}
As suggested by our notation, the error $\BigResErr$ is a higher-order
term, decaying as $\numobs^{-3/2}$ in the sample size, whereas the
quantity $\BigEstErr$ is the dominant source of statistical error.
With this notation, we have the following:
\begin{theorem}
  \label{thm:linear-oracle-ineq}
  Suppose that we are given $\numobs$ i.i.d. observations $\{
  (\Lmat_i, \bvec_i) \}_{i=1}^\numobs$ that satisfy the noise
  conditions in Assumption~\ref{assume-second-moment}.  Then there are
  universal constants $(c_0, c)$ such that for any sample size
  $\numobs \geq \frac{c_0 \sigmaA^2 d}{(1 - \kappa)^2} \log^2 \left(
  \frac{\statnorm{\vvec_0 - \vbar}^2 d}{1 - \kappa} \right)$, then
  running the algorithm~\eqref{eq:SA} with
  \begin{align*}
   \mbox{stepsize $\stepsize = \frac{1}{c_0 \sigmaA \sqrt{d \numobs}}$},
   \quad \mbox{ and burn-in period $\numburn = \numobs/2$}
  \end{align*}
  yields an estimate $\vhat_\numobs$ such that
\begin{equation}
\label{eq:thm1-rate}
\Exs \statnorm{\vhat_\numobs - \vstar}^2 \leq (1 + \offpar) \cdot
\prefact(\SpecMat, \hilopnorm{\Lmat}) \inf_{v \in \lspace} \statnorm{v
  - \vstar}^2 + c \left( 1 + \tfrac{1}{\offpar} \right) \cdot \big \{
\BigEstErr + \BigResErr \big \},
\end{equation}
valid for any $\offpar > 0$.
\end{theorem}
\noindent
We prove this theorem in Section~\ref{SecProofThmLinearOracleIneq}. \\

A few comments are in order. First, the quantity $\prefact(\SpecMat,
\hilopnorm{\Lmat}) \inf_{v \in \lspace} \statnorm{v - \vstar}^2$ is an
upper bound on the approximation error $\statnorm{\vbar - \vstar}^2$
incurred by the (deterministic) projected fixed point $\vbar$.  The
pre-factor $\prefact(\SpecMat, \hilopnorm{\Lmat}) \geq 1$ measures the
instance-specific deficiency of $\vbar$ relative to an optimal
approximating vector from the subspace, and we provide a more in-depth
discussion of this factor in Section~\ref{sec:approx} to follow.  Note
that Theorem~\ref{thm:linear-oracle-ineq} actually provides a family
of bounds, indexed by the free parameter $\offpar > 0$.  By choosing
$\offpar$ arbitrarily close to zero, we can make the pre-factor in
front of $\inf_{v \in \lspace} \statnorm{v - \vstar}^2$ arbitrarily
close to $\prefact (\SpecMat, \hilopnorm{\Lmat})$---albeit at the
expense of inflating the remaining error terms.  In
Theorem~\ref{thm:linear-lower-bound} to follow, we prove that the
quantity $\prefact(\SpecMat, \hilopnorm{\Lmat})$ is, in fact, the
smallest approximation factor that can be obtained in any such bound.

The latter two terms in the bound~\eqref{eq:thm1-rate} correspond to
estimation error that arises from estimating $\vbar$ based on a set of
$\numobs$ stochastic observations. While there are two terms here in
principle, we show in Corollary~\ref{CorExtreme} to follow that the
estimation error is dominated by the term $\BigEstErr$ under some
natural assumptions.  Note that the leading term $\BigEstErr$ scales
with the local complexity for estimating $\vbar$, and we show in
Theorem~\ref{thm:linear-stat-error-lower-bound} that this term is also
information-theoretically optimal.


In the next subsection, we undertake a more in-depth exploration of
the approximation factor in this problem, discussing prior work in the
context of the term $\prefact(\SpecMat, \hilopnorm{\Lmat})$ appearing
in Theorem~\ref{thm:linear-oracle-ineq}.

\subsubsection{Detailed discussion of the approximation error}
\label{sec:approx}

As mentioned in the introduction, upper bounds on the approximation
factor have received significant attention in the literature, and it
is interesting to compare our bounds.

\paragraph{Past results:}
In the case where $\conmax \mydefn \hilopnorm{\Lmat} < 1$, the
approximation-factor
bound~\eqref{eq:contraction-approx-factor-worst-case} was established
by Tsitsiklis and Van Roy~\cite{tsitsiklis1997analysis}, via the
following argument.  Letting $\vtil \defn \projecttolin (\Lmat \vstar
+ \bvec)$, we have
\begin{align}
      \statnorm{\vbar - \vstar}^2 \overset{(i)}{=} \statnorm{\vbar -
        \vtil}^2 + \statnorm{\vtil - \vstar}^2 &=
      \statnorm{\projecttolin (\Lmat \vbar + \bvec) - \projecttolin
        (\Lmat \vstar + \bvec)}^2 + \statnorm{\vtil -
        \vstar}^2 \notag \\ 
        &\overset{(ii)}{\leq} \statnorm{\Lmat \vbar - \Lmat
        \vstar}^2 + \statnorm{\vtil - \vstar}^2 \notag \\
        &\overset{(iii)}{\leq}
      \conmax^2 \statnorm{\vbar - \vstar}^2 + \statnorm{\vtil -
        \vstar}^2. \label{eq:contraction-approx-factor-back-of-envelop}
\end{align}
Step (i) uses Pythagorean theorem; step (ii) follows from the
non-expansiveness of the projection operator; and step (iii) makes use
of the contraction property of the operator $\Lmat$. Note that by
definition, we have $\prefact(M, \hilopnorm{\Lmat}) \leq (1 -
\hilopnorm{\Lmat})^{-2}$, and so the approximation factor in
Theorem~\ref{thm:linear-oracle-ineq} recovers the
bound~\eqref{eq:contraction-approx-factor-worst-case} in the worst
case. In general, however, the factor $\prefact(M, \hilopnorm{\Lmat})$
can be significantly smaller.

Yu and Bertsekas~\cite{yu2010error} derived two fine-grained
approximation factor upper; in terms of our notation, their bounds
take the form
\begin{align*}
  \prefact_{\mathsf{YB}}^{(1)} &\mydefn 1 + \hilopnorm{\Lmat}^2 \cdot
  \lammax \left( (I - \SpecMat)^{-1} (I -
  \SpecMat)^{-\top}\right),\\
  \prefact_{\mathsf{YB}}^{(2)} & \mydefn 1 + \hilopnorm{(I -
    \projecttolin \Lmat)^{-1} \projecttolin \Lmat
    \projectto{\LinSpace^\perp}}^2.
\end{align*}
It is clear from the definition that $\prefact (\SpecMat,
\hilopnorm{\Lmat}) \leq \prefact_{\mathsf{YB}}^{(1)}$, but $\prefact
(\SpecMat, \hilopnorm{\Lmat})$ can often provide an improved
bound. This improvement is indeed significant, as will be shown
shortly in Lemma~\ref{lem:approx-factor-upper-bounds}.  On the other
hand, the term $\prefact_{\mathsf{YB}}^{(2)}$ is never larger than
$\prefact (\SpecMat, \hilopnorm{\Lmat})$, and is indeed the smallest
possible bound that depends only on $\Lmat$ and \emph{not}
$\bvec$. However, as pointed out by Yu and Bertsekas, the value of
$\prefact_{\mathsf{YB}}^{(2)}$ is not easily accessible in practice,
since it depends on the precise behavior of the operator $\Lmat$ over
the orthogonal complement $\LinSpace^\perp$.  Thus, estimating the
quantity $\prefact_{\mathsf{YB}}^{(2)}$ requires $O (\bigdim)$
samples. In contrast, the term $\prefact (\SpecMat,
\hilopnorm{\Lmat})$ depends only on the projected operator $\SpecMat$
and the operator norm $\hilopnorm{\Lmat}$. The former can be easily
estimated using $\usedim$ samples and at smaller computational cost,
while the latter is usually known a priori.  The discussion in
Section~\ref{SecConsequences} to follow fleshes out these
distinctions.

\paragraph{A simulation study:}

In order to compare different upper bounds on the approximation
factor, we conducted a simple simulation study on the problem of value function
estimation, as previously introduced in
Section~\ref{ex:td-discounted}.  For this problem, the approximation
factor $\prefact (\SpecMat, \discount)$ is computed more explicitly in
Corollary~\ref{corr:oracle-ineq-lstd-iid-discount}. The Markov
transition kernel is given by the simple random walk on a graph. We
consider Gaussian random feature vectors and associate them with two
different random graph models, Erd\"{os}-R\'{e}nyi graphs and random
geometric graphs, respectively. The details for these models are
described and discussed in Appendix~\ref{Appendix:simulation-details}.

In Figure~\ref{figure:simulation}, we show the simulation results for
the values of the approximation factor. Given a sample from above
graphs and feature vectors, we plot the value of $\prefact(\SpecMat,
\discount)$, $\prefact^{(1)}_{\mathsf{YB}}$ and
$\prefact^{(2)}_{\mathsf{YB}}$ against the discount rate $1 -
\discount$, which ranges from $10^{-5}$ to $10^{-0.5}$. Note that the
two plots use different scales: Panel (a) is a linear-log plot, whereas
panel (b) is a log-log plot.
\begin{figure}[htb]
  \begin{center}
    \begin{tabular}{ccc}
      \widgraph{0.45\textwidth}{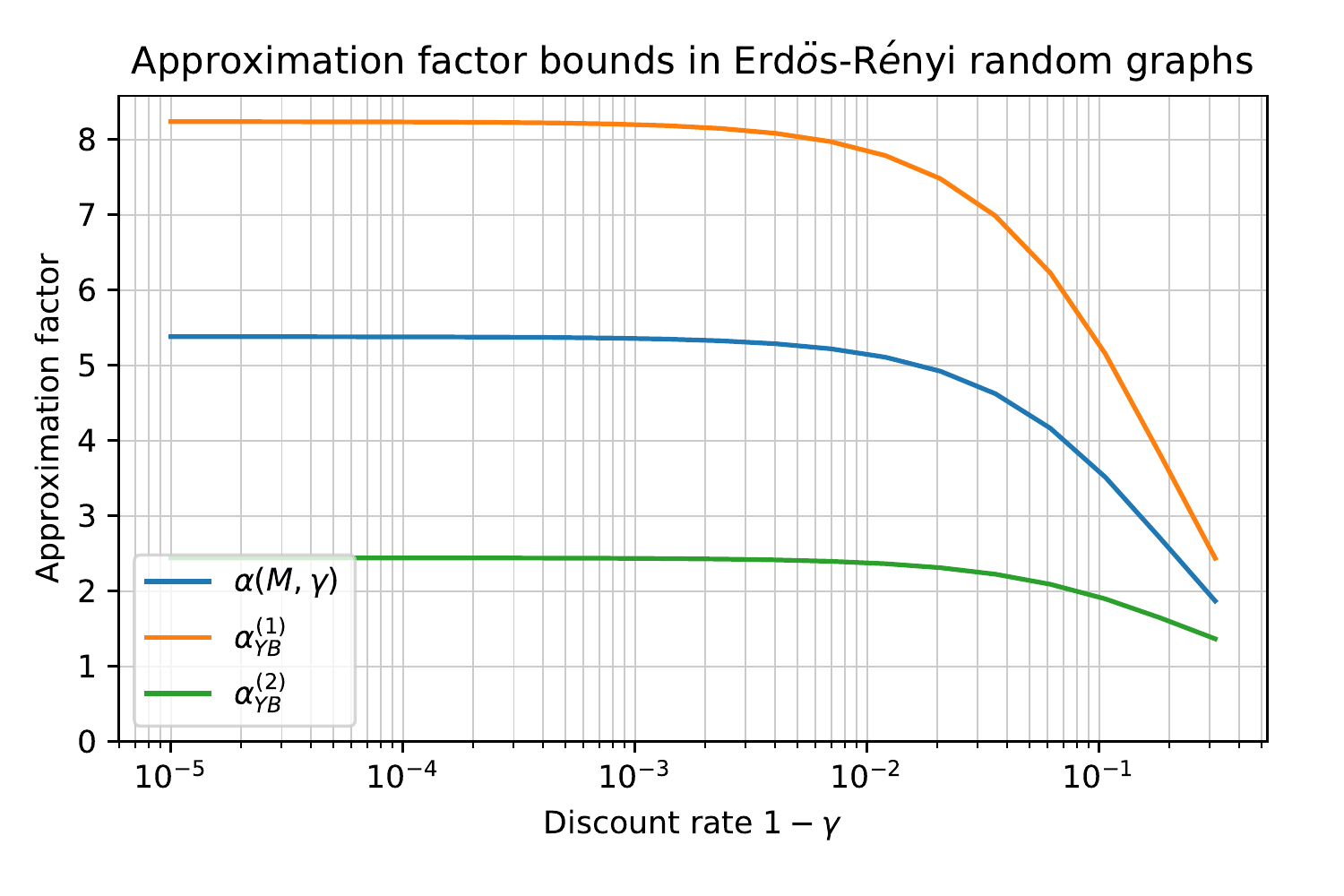} &&
      \widgraph{0.45\textwidth}{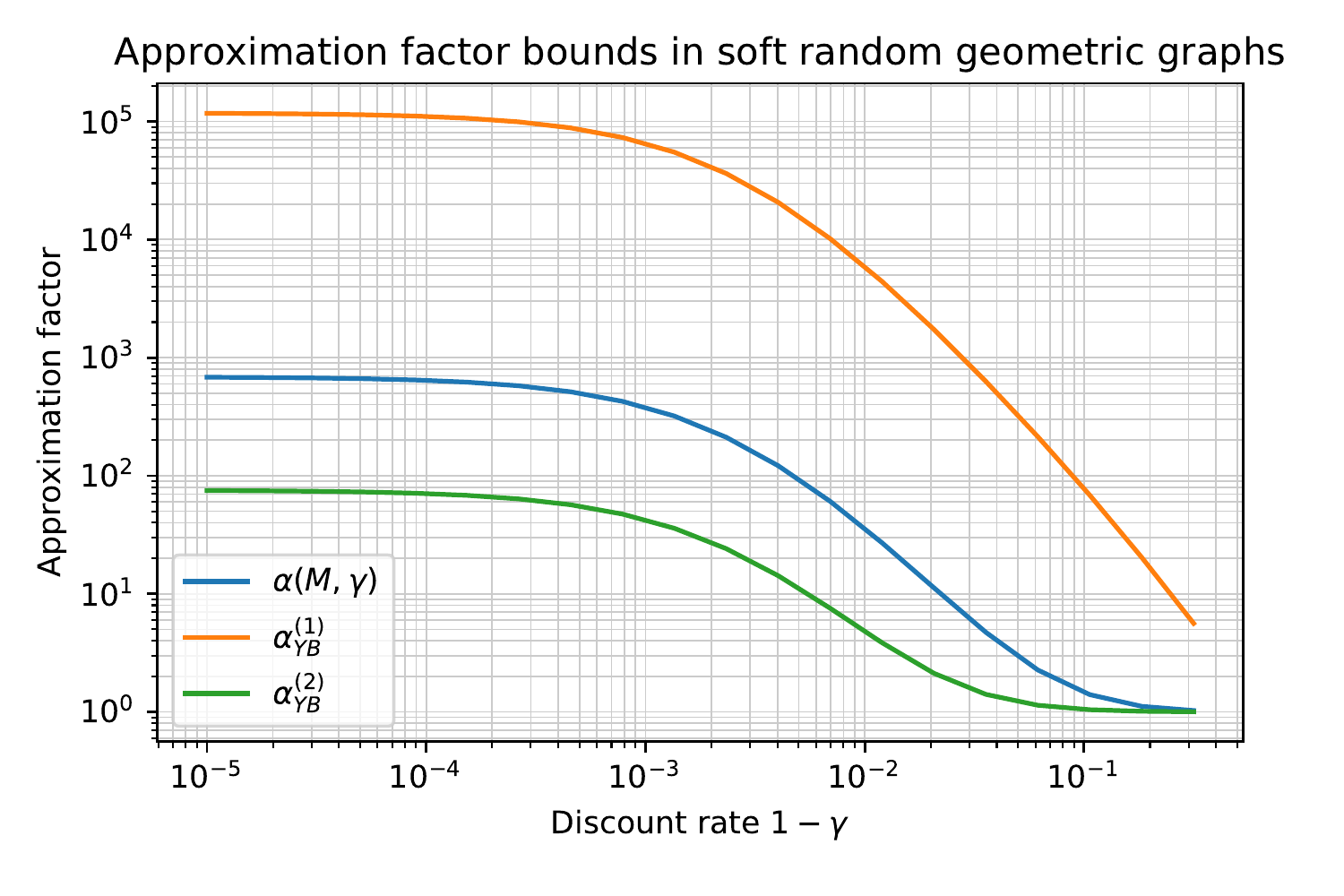} \\
      (a) && (b)
    \end{tabular}
    \caption{Plots of various approximation factor as a function of
      the discount factor $\discount$ in the policy evaluation
      problem. (See the text for a discussion.) 
      (a) Results for an Erd\"{o}s-R\'{e}nyi random graph
      model with $\npar = 3000$, projected dimension $\usedim = 1000$,
      and $a = 3$. The resulting number of vertices in the graph
      $\widetilde{G}$ is $2813$. The value of $1 - \discount$ is
      plotted in log-scale, and the value of approximation factor is
      plotted on the standard scale. (b) Results for a random
      geometric graph model with $\npar = 3000$, projected dimension
      $\usedim = 2$, and $r = 0.1$. The resulting number of vertices
      in the graph $\widetilde{G}$ is $2338$. Both the discount rate
      $1 - \discount$ and the approximation factor are plotted on the
      log-scale.}
\label{figure:simulation}
  \end{center}
\end{figure}
Figure~\ref{figure:simulation} shows that the approximation factor
$\prefact(\SpecMat, \discount)$ derived in
Theorem~\ref{thm:linear-oracle-ineq} is always between
$\prefact^{(1)}_{\mathsf{YB}}$ and~$\prefact^{(2)}_{\mathsf{YB}}$. 
As mentioned before, the latter quantity depends
on the particular behavior of the linear operator $\Lmat$ in the subspace
$\LinSpace^\perp$, which can be difficult to estimate. The improvement
over $\alpha^{(1)}_{\mathsf{YB}}$, on the other hand, can be
significant. 

In the Erd\"{o}s-R\'{e}nyi model, all the three quantities are
bounded by relatively small constant, regardless of the value of
$\discount$. The bound $\prefact(\SpecMat, \discount)$ is roughly at
the midpoint between the bounds $\prefact^{(1)}_{\mathsf{YB}}$ and
$\prefact^{(2)}_{\mathsf{YB}}$. The differences are much starker in 
the random geometric graph case:
The bound improves over $\prefact^{(1)}_{\mathsf{YB}}$ by several
orders of magnitude, while being off from
$\prefact^{(2)}_{\mathsf{YB}}$ by a factor of $10$ for large
$\discount$. As we discuss shortly in
Lemma~\ref{lem:approx-factor-upper-bounds}, this is because the
approximation factor $\prefact (\SpecMat, \discount)$ scales as $O
\big( \frac{1}{1 - \kappa (\SpecMat)} \big)$ while
$\prefact^{(1)}_{\mathsf{YB}}$ scales as $O \big( \frac{1}{(1 - \kappa
  (\SpecMat))^2} \big)$, making a big difference for the case with
large correlation.


\paragraph{Some useful bounds on $\prefact (\SpecMat, \hilopnorm{\Lmat})$:}

We conclude our discussion of the approximation factor with some
bounds that can be derived under different assumptions on the operator
$\Lmat$ and its projected version $\SpecMat$. The following lemma is
useful in understanding the behavior of the approximation factor as a
function of the contractivity properties of the operator $\Lmat$; this
is particularly useful in analyzing convergence rates in numerical
PDEs.
\begin{lemma}
\label{lem:approx-factor-upper-bounds}
    Consider a projected matrix $\SpecMat \in \real^{\usedim \times
      \usedim}$ such that $(I - \SpecMat)$ is invertible and
    $\kappa(\SpecMat) < 1$.
    \begin{enumerate}
    \item[(a)] For any $s > 0$, we have the bound
\begin{subequations}      
\begin{align}
        \prefact(\SpecMat, s) \leq 1 + \opnorm{(I - \SpecMat)^{-1}}^2
        \cdot s^2 \leq 1 + \frac{s^2}{(1 - \kappa(\SpecMat))^2}.
\end{align}
    \item[(b)] For $s \in [0,1]$, we have
    \begin{align}
      \alpha (\SpecMat, s) & \leq 1+ 2 \opnorm{(I - \SpecMat)^{-1}}
      \leq 1 + \frac{2}{1 - \kappa(\SpecMat)}.
    \end{align}
\end{subequations}
    \end{enumerate}
\end{lemma}
\noindent See Appendix~\ref{app:proof-of-approx-factor-upper-bounds}
for the proof of this lemma. 

A second special case, also useful, is when the matrix $\SpecMat$ is
symmetric, a setting that appears in least-squares regression, value
function estimation in reversible Markov chains, and self-adjoint
elliptic operators.  The optimal approximation factor $\prefact
(\SpecMat, \conmax)$ can be explicitly computed in such cases.

\begin{lemma}
\label{lem:approx-factor-symmetric}
Suppose that $\SpecMat$ is symmetric with eigenvalues
$\{\lambda_j(\SpecMat) \}_{j=1}^\usedim$ such that $\lammax(\SpecMat)
< 1$.  Then for any $s > 0$, we have
\begin{align}
  \prefact(\SpecMat, s) & = 1 + \max_{j = 1, \ldots, \usedim}
  \frac{s^2 - \lambda_j^2}{(1 - \lambda_j)^2 }.
\end{align}
\end{lemma}
\noindent See Appendix~\ref{app:proof-of-approx-factor-sym} for the
proof of this lemma. \\

Lemma~\ref{lem:approx-factor-upper-bounds} reveals that there is a
qualitative shift between the non-expansive case $\hilopnorm{\Lmat}
\leq 1$ and the complementary expansive case. In the latter case, the
optimal approximation factor always scales as $O \big( \frac{1}{(1 -
  \kappa (\SpecMat))^2} \big)$, but below the threshold
$\hilopnorm{\Lmat} = 1$, the approximation factor drastically improves
to become $O \big( \frac{1}{1 - \kappa (\SpecMat)} \big)$.  It is
worth noting that both bounds can be achieved up to universal constant
factors.  In the context of differential equations, the bound of the
form $(a)$ in Lemma~\ref{lem:approx-factor-upper-bounds} is known as
C\'{e}a's lemma~\cite{cea1964approximation}, which plays a central
role in the convergence rate analysis of the Galerkin methods for
numerical differential equations. However, the instance-dependent
approximation factor $\prefact(\SpecMat, \hilopnorm{\Lmat} )$ can
often be much smaller: the global coercive parameter needed in
C\'{e}a's estimate is replaced by the bounds on the behavior of the
operator $\Lmat$ in the finite-dimensional subspace. The part $(b)$ in
Lemma~\ref{lem:approx-factor-upper-bounds} generalizes C\'{e}a's
energy estimate from the symmetric positive-definite case to the
general non-expansive setting.
See Corollary~\ref{corr:elliptic} for a more detailed
discussion on the consequences of our results to elliptic PDEs.


Lemmas~\ref{lem:approx-factor-upper-bounds} and~\ref{lem:approx-factor-symmetric} yield the
following corollary of the general bound~\eqref{eq:thm1-rate} under different conditions on
the operator $\Lmat$.
\begin{corollary}
\label{CorExtreme}  
Under the conditions of Theorem~\ref{thm:linear-oracle-ineq} and given
a sample size $\numobs \geq \frac{c_0 \sigmaA^2 \usedim}{(1 -
  \kappa)^2} \log^2 \left( \frac{\statnorm{\vvec_0 - \vbar}^2
  \usedim}{1 - \kappa} \right)$:
  \begin{enumerate}
  \item[(a)] There is a universal positive constant $c$ such that
    \begin{subequations}
    \begin{align}
      \Exs \statnorm{\vhat_\numobs - \vstar}^2 & \leq c \, \left
      \{ \frac{\hilopnorm{\Lmat}^2}{\big(1 - \kappa(\SpecMat)
        \big)^2} \cdot \inf_{\vvec \in \lspace} \statnorm{\vvec
        - \vstar}^2 + \frac{(\sigmab^2 + \sigmaA
        \statnorm{\vbar}^2) }{\big( 1 - \kappa(\SpecMat) \big)^2
      } \; \frac{\usedim}{\numobs} \right \}
    \end{align}
    for any operator $\Lmat$, and its associated projected operator
    $\SpecMat = \PhiOp \Lmat \adjoint{\PhiOp}$.
\item[(b)] Moreover, when $\Lmat$ is non-expansive ($\hilopnorm{\Lmat}
  \leq 1$), we have
    \begin{align}
      \Exs \statnorm{\vhat_\numobs - \vstar}^2 & \leq c \, \left
      \{ \frac{1}{1 - \kappa(\SpecMat)} \cdot \inf_{\vvec
        \in \lspace} \statnorm{\vvec - \vstar}^2 +
      \frac{(\sigmab^2 + \sigmaA \statnorm{\vbar}^2) }{\big( 1 -
              \kappa(\SpecMat) \big)^2 } \; \frac{\usedim}{\numobs}
      \right \}.
    \end{align}
    \end{subequations}
  \end{enumerate}
\end{corollary}
\noindent See Section~\ref{SecProofCorExtreme} for the proof of this
claim.

As alluded to before, the simplified form of
Corollary~\ref{CorExtreme} no longer has an explicit higher order
term, and the statistical error now scales at the parametric rate $d/
n$. It is worth noting that the lower bound on $\numobs$ required in
the assumption of the corollary is a mild requirement: in the absence
of such a condition, the statistical error term $ \frac{(\sigmab^2 +
  \sigmaA\statnorm{\vbar}^2) }{(1 - \kappa)^2} \;
\frac{\usedim}{\numobs}$ in both bounds would blow up, rendering the
guarantee vacuous.


\subsection{Lower bounds}
\label{SecLower}

In this section, we establish information-theoretic lower bounds on
the approximation factor, as well as the statistical error. Our
eventual result (in Corollary~\ref{cor:lower-bound-final}) shows that
the first two terms appearing in Theorem~\ref{thm:linear-oracle-ineq}
are both optimal in a certain instance-dependent sense. However, a 
precise definition of the local neighborhood of instances over which
the lower bound holds requires some definitions.  In order to motivate
these definitions more transparently and naturally arrive at both
terms of the bound, the following section presents individual bounds
on the approximation and estimation errors, and then combines them to
obtain Corollary~\ref{cor:lower-bound-final}.


\subsubsection{Lower bounds on the approximation error}

As alluded to above, the first step involved in a lower bound is a
precise definition of the collection of problem instances over which
it holds; let us specify a natural such collection for lower bounds on
the approximation error. Each problem instance is specified by the
joint distribution of the observations $(\Lmat_i, \bvec_i)$, which
implicitly specifies a pair of means $(\Lmat, \bvec) = (\Exs
[\Lmat_i], \Exs [\bvec_i])$. For notational convenience, we define
this class by first defining a collection comprising instances
specified solely by the mean pair $(\Lmat, \bvec)$, and then providing
restrictions on the distribution of $(\Lmat_i, \bvec_i)$.  Let us
define the first such component. For a given matrix $\SpecMat_0 \in
\real^{\usedim \times \usedim}$ and vector $\projectedbvec_0 \in
\real^{\usedim}$, write
\begin{align*}
    \classApprox (\SpecMat_0, \projectedbvec_0, \bigdim, \oracleErr,
    \conmax) \mydefn \left\{ (\Lmat, \bvec) ~\Big| \begin{array}{c}
      \hilopnorm{\Lmat} \leq \conmax, \quad \AppErr (\LinSpace,
      \vstar) \leq \oracleErr^2, \quad \mathrm{dim} (\Xspace) = D,
      \\ \PhiOp \Lmat \adjoint{\PhiOp} = \SpecMat_0, \;\; \text{ and }
      \;\; \PhiOp b = \projectedbvec_0.
      \end{array}
    \right\}.
\end{align*}
In words, this is a collection of all instances of the pair $(\Lmat,
\bvec) \in \AClass \times \real^D$ whose projections onto the subspace
of interest are fixed to be the pair $(\SpecMat_0, \projectedbvec_0)$,
and whose approximation error is less than $\delta^2$. In addition,
the operator $\Lmat$ satisfies a certain bound on its operator norm.

Having specified a class of $(\Lmat, \bvec)$ pairs, we now turn to the
joint distribution over the pair of observations $(\Lmat_i, \bvec_i)$,
which we denote for convenience by $\ProbInst_{\Lmat, \bvec}$. Now
define the collection of instances
\begin{align*}
  \classNoise (\sigmaA, \sigmab) \mydefn \left\{
  \ProbInst_{\Lmat, \bvec} ~\Big| \mbox{ $(\Lmat_i, b_i)$
    satisfies Assumption~\ref{assume-second-moment-strong} with constants } (\sigmaA, \sigmab)    \right\}.
\end{align*}
This is simply the class of all distributions such that our
observations satisfy Assumption~\ref{assume-second-moment-strong} with
pre-specified constants. As a point of clarification, it is useful to
recall that our upper bound in Theorem~\ref{thm:linear-oracle-ineq}
only needed Assumption~\ref{assume-second-moment} to hold, and we
could have chosen to match this by defining the $\classNoise$ under
Assumption~\ref{assume-second-moment}.  We comment further on this issue
following the theorem statement.

We are now ready to state Theorem~\ref{thm:linear-lower-bound}, which
is a lower bound on the worst-case approximation factor over all
problem instances such that $(\Lmat, \bvec) \in \classApprox
(\SpecMat_0, \projectedbvec_0, \bigdim, \oracleErr, \conmax)$ and
$\ProbInst_{\Lmat, \bvec} \in \classNoise (\sigmaA, \sigmab)$.  Note
that such a collection of problem instances is indeed \emph{local}
around the pair $(\SpecMat_0, \projectedbvec_0)$.  Two settings are
considered in the statement of the theorem: \emph{proper} estimators
when $\vhat_n$ is restricted to take values in the subspace $\lspace$;
and \emph{improper} estimators, where $\vhat_n$ can take values in the
entire space $\Xspace$. We use $\Vhatclass_\LinSpace$ and
$\Vhatclass_\Xspace$ to denote the class of proper and improper
estimators, respectively.  Finally, we use the shorthand $\classApprox
\equiv \classApprox(\SpecMatZero, \projectedbvecZero, \bigdim,
\oracleErr, \conmax)$ for convenience.

\begin{theorem}
\label{thm:linear-lower-bound}
Suppose $\SpecMat_0 \in \real^{d \times d}$ is a matrix such that $I -
\SpecMat_0$ is invertible, and that the scalars $(\sigmaA, \sigmab)$
are such that $\sigmaA \geq \conmax$ and $\sigmab \geq \oracleErr$.
If the ambient dimension satisfies $D \geq d + \frac{12 }{\offpar}
n^2$ for some scalar $\offpar \in (0, 1)$, then we have the lower
bounds
\begin{subequations}
\begin{align}
  \inf_{\vhat_n \in \Vhatclass_\lspace} ~\sup_{\substack{ (\Lmat,
      \bvec) \in \classApprox \\ \ProbInst_{\Lmat, \bvec} \in
      \classNoise (\sigmaA, \sigmab) }} \Exs \statnorm{\vhat_n -
    \vstar}^2 &\geq (1 - \offpar) \cdot \prefact (M_0, \conmax) \cdot
  \oracleErr^2 \quad \text{ and }\\ \inf_{\vhat_n \in
    \Vhatclass_\Xspace} ~\sup_{\substack{ (\Lmat, \bvec) \in
      \classApprox \\ \ProbInst_{\Lmat, \bvec} \in \classNoise
      (\sigmaA, \sigmab) }} \Exs \statnorm{\vhat_n - \vstar}^2 &\geq
  (1 - \offpar) \cdot \big( \alpha (M_0, \conmax) - 1 \big) \cdot
  \oracleErr^2.
\end{align}
\end{subequations}
\end{theorem}
\noindent See Section~\ref{subsec:proof-approx-factor-lower-bound} for the proof of this claim.

A few remarks are in order. First,
Theorem~\ref{thm:linear-lower-bound} shows that the approximation
factor upper bound in Theorem~\ref{thm:linear-oracle-ineq} is
information-theoretically optimal in the instance-dependent sense: in
the case of proper estimators, the upper and lower bound can be made
arbitrarily close by choosing the constant $\offpar$ arbitrarily small
in both theorems. Both bounds depend on the projected matrix
$\SpecMatZero$, characterizing the fundamental impact of the geometry
in the projected space on the complexity of the estimation
problem. The lower bound for improper estimators is slightly smaller,
but for most practical applications we have $\prefact (\SpecMatZero,
\conmax) \gg 1$ and so this result should be viewed as almost
equivalent.
 
 Second, note that we may also extract a worst-case lower bound on the
 approximation factor from
 Theorem~\ref{thm:linear-lower-bound}. Indeed, for a scalar $\conmax
 \in (0, 1)$, consider the family of instances in the aforementioned
 problem classes satisfying $\hilopnorm{\Lmat} \leq \conmax$. Setting
 $\SpecMatZero = \conmax^2 I_\usedim$ and applying
 Theorem~\ref{thm:linear-lower-bound}, we see that (in a worst-case
 sense over this class), the risk of any estimator is lower bounded by
 $\frac{1}{1 - \conmax^2} \AppErr (\LinSpace, \vstar)$. This
 establishes the optimality of the classical worst-case upper
 bound~\eqref{eq:contraction-approx-factor-worst-case}.

Third, notice that theorem requires the noise variances $(\sigmaA,
\sigmab)$ to be large enough, and this is a natural requirement in
spite of the fact that we seek lower bounds on the approximation
error.  Indeed, in the extreme case of noiseless observations, we have
access to the population pair $(\Lmat, \bvec)$ with a single sample,
and can compute both $\vstar$ and its projection onto the subspace
$\lspace$ without error.  From a more quantitative standpoint, it is
worth noting that our requirements $\sigmaA \geq \conmax$ and $\sigmab
\geq \oracleErr$ are both mild, since the scalars $\conmax$ and
$\oracleErr$ are typically order $1$ quantities. Indeed, if both of
these bounds held with equality, then Corollary~\ref{CorExtreme}
yields that the statistical error would be of the order $O (\usedim /
\numobs)$, and so strictly smaller than the approximation error we
hope to capture\footnote{As a side remark, we note that our noise
conditions can be further weakened, if desired, via a mini-batching
trick. To be precise, given any problem instance $\ProbInst_{\Lmat,
  \bvec} \in \classNoise ( \sigmaA, \sigmab )$ and any integer $m >
0$, one could treat the sample mean of $m$ independent samples as a
single sample, resulting in a problem instance in the class
$\classNoise \big( \frac{\sigmaA}{\sqrt{m}}, \frac{\sigmab}{\sqrt{m}}
\big)$. The same lower bound still applies to the class $\classNoise
\big( \frac{\sigmaA}{\sqrt{m}}, \frac{\sigmab}{\sqrt{m}} \big)$, at a
cost of stronger dimension requirement $\bigdim \geq \usedim +
\frac{12}{\offpar} n^2 m^2$.}.

Observe that Theorem~\ref{thm:linear-lower-bound} requires the ambient
dimension $\bigdim$ to be larger than $\numobs^2$. As mentioned in the
introduction, we should not expect any non-trivial approximation
factor when $n \geq D$, but this leaves open the regime $\numobs \ll
\bigdim \ll \numobs^2$. Is a smaller approximation factor achievable
when $D$ is not extremely large?  We revisit this question in
Section~\ref{subsubsec:intermediate-regime}, showing that while there
are some quantitative differences in the lower bound, the qualitative
nature of the message remains unchanged.

Regarding our noise assumptions, it should be noted that the class of
instances satisfying Assumption~\ref{assume-second-moment} is strictly
larger than the corresponding class satisfying
Assumption~\ref{assume-second-moment-strong}, and so our lower bound
to follow extends immediately to the former case. Second, it is
important to note that imposing only
Assumption~\ref{assume-second-moment} would in principle allow the
noise in the orthogonal complement $\LinSpace^{\perp}$ to grow in an
unbounded fashion, and one should expect that it is indeed optimal to
return an estimate of the projected fixed point $\vbar$. To establish
a more meaningful (but also more challenging) lower bound, we operate
instead under the stronger
Assumption~\ref{assume-second-moment-strong}, enforcing second moment
bounds on the noise not only for basis vectors in $\LinSpace$, but
also its orthogonal
complement. Assumption~\ref{assume-second-moment-strong} allows for
other natural estimators: For instance, the plug-in estimator of
$\vstar$ via the original fixed point equation~\eqref{EqnFixedPoint}
would now incur finite error. Nevertheless, as shown by our lower
bound, the stochastic approximation estimator analyzed in
Theorem~\ref{thm:linear-oracle-ineq} is optimal \emph{even if} the
noise in $\LinSpace^\perp$ behaves as well as that in~$\LinSpace$.


\subsubsection{Lower bounds on the estimation error}

We now turn to establishing a minimax lower bound on the estimation
error that matches the statistical error term in
Theorem~\ref{thm:linear-oracle-ineq}.  This lower bound takes a
slightly different form from Theorem~\ref{thm:linear-lower-bound}:
rather than studying the total error $\statnorm{\vhat_n - \vstar}$
directly, we establish a lower bound on the error $\statnorm{\vhat_n -
  \vbar}$ instead.

Indeed, the latter term is more meaningful to study in order to
characterize the estimation error---which depends on the sample size
$n$---since for large sample sizes, the total error $\statnorm{\vhat_n
  - \vstar}$ will be dominated by a constant approximation error.  As
we demonstrate shortly, the term $\statnorm{\vhat_n - \vbar}$ depends
on noise covariance and the geometry of the matrix $M_0$ in the
\emph{projected space}, while having the desired dependence on the
sample size $n$.  It is worth noting also that this automatically
yields a lower bound on the error $\statnorm{\vhat_n - \vstar}$ when
we have $\vbar = \vstar$.

We are now ready to prove a local minimax lower bound for estimating
$\vbar \in \lspace$, which is given by the solution to the projected
linear equation $\vbar = \projecttolin (\Lmat \vbar + \bvec)$. While
our objective is to prove a local lower bound around each pair
$(\Lmat_0, \bvec_0) \in \AClass \times \Xspace$, the fact that we are
estimating $\vbar$ implies that it suffices to define our set of local
instances in the $d$-dimensional space of projections. In particular,
our means $(\Lmat, \bvec)$ are specified by those pairs for which
$\PhiOp \Lmat \adjoint{\PhiOp}$ is close to $\SpecMatZero \defn \PhiOp
\Lmat_0 \adjoint{\PhiOp}$, and $\PhiOp b$ is close to
$\projectedbvec_0 \defn \PhiOp b_0$. Specifically, let $\vbar_0$
denote the solution to the projected linear equation $\vbar_0 =
\projecttolin (\Lmat_0 \vbar_0 + \bvec_0)$, and define the
neighborhood
\begin{align}
\label{def:local-neighborhood}  
\neighborhood (\SpecMat_0, \projectedbvec_0) \mydefn \left\{
(\SpecMat', \projectedbvec'): \matsnorm{\SpecMat' - \SpecMat_0}{F}
\leq \sigmaA \sqrt{\frac{\usedim}{\numobs}}, ~\mbox{and}~
\vecnorm{\projectedbvec' - \projectedbvec_0}{2} \leq \sigmab
\sqrt{\frac{\usedim}{\numobs}} \right\},
\end{align}
which, in turn, defines a local class of problem instances $(\Lmat,
\bvec)$ given by
\begin{align*}
  \classEst \mydefn \Big\{ (\Lmat, \bvec) \mid \big( \PhiOp \Lmat
  \adjoint{\PhiOp}, \PhiOp \bvec \big) \in \neighborhood
  (\SpecMat_0, \projectedbvec_0) \Big\}.
\end{align*}

We have thus specified our local neighborhood in terms of the mean
pair $(\Lmat, \bvec)$, and as before, it remains to define a local
class of distributions on these instances. Toward this end, define the
class
\begin{multline}
  \label{defn:classcov}
  \classCov (\Sigma_\Lmat, \Sigma_\bvec, \sigmaA, \sigmab)  \\
  \mydefn \classNoise(\sigmaA, \sigmab) \cap \left\{
\ProbInst_{\Lmat, \bvec} ~\Big| \cov \left( \PhiOp (\bvec_1 - \bvec)
\right) \preceq \Sigma_\bvec \;\; \text{ and } \;\; \cov \left( \PhiOp
(\Lmat_1 - \Lmat) \vbar_0 \right) \preceq \Sigma_\Lmat \right\},
\end{multline}
corresponding to distributions on the observation pair $(\Lmat_i,
\bvec_i)$ that satisfy Assumption~\ref{assume-second-moment-strong}
and whose ``effective noise'' covariances are dominated by the PSD
matrices $\Sigma_\Lmat$ and $\Sigma_\bvec$.

Note that Assumption~\ref{assume-second-moment-strong} implies the
diagonal elements of above two covariance matrices are bounded by
$\sigmab^2$ and $\sigmaA^2 \statnorm{\vbar_0}^2$, respectively. In
order to avoid conflicts between assumptions, we assume throughout
that for all indices $j \in [d]$, the diagonal entries of the
covariance matrices satisfy the conditions
\begin{align}
\label{eq:compatibility-condition-cov}  
 (\Sigma_\bvec)_{j, j} \leq \sigmab^2 \qquad \text{ and } \qquad
(\Sigma_\Lmat)_{j, j} \leq \sigmaA^2 \statnorm{\vbar_0}^2.
\end{align}
We then have the following theorem for the estimation error
$\statnorm{\vhat_n - \vbar}$, where we use the shorthand $\classCov
\equiv \classCov (\Sigma_\Lmat, \Sigma_\bvec, \sigmaA, \sigmab)$ for
brevity.
\begin{theorem}
\label{thm:linear-stat-error-lower-bound}
  Under the setup above, suppose the matrix $I - \SpecMat_0$ is
  invertible, and suppose that $\numobs \geq 16 \sigmaA^2 \opnorm{(I -
    \SpecMat_0)^{-1}}^2 d$.  Then there is a universal constant $c >
  0$ such that
  \begin{align*}
    \inf_{\vhat_n \in \Vhatclass_\Xspace} \; \sup_{ \substack{
        (\Lmat, \bvec) \in \classEst \\ \ProbInst_{\Lmat,
              \bvec}\in \classCov }} \; \Exs \statnorm{\vhat_n -
      \vbar}^2 \geq c \cdot \EstErr_n (\SpecMat_0, \Sigma_\Lmat +
    \Sigma_\bvec).
  \end{align*}
\end{theorem}

\noindent See Section~\ref{subsec:proof-est-error-lower-bound} for the
proof of this claim.

The estimation error lower bound in
Theorem~\ref{thm:linear-stat-error-lower-bound} matches the
statistical error term $\BigEstErr$ in
Theorem~\ref{thm:linear-oracle-ineq}, up to a universal constant.
Indeed, in the asymptotic limit $\numobs \rightarrow \infty$, the
regularity of the problem can be leveraged in conjunction with
classical Le Cam theory (see, e.g.,~\cite{van2000asymptotic}) to show
that the asymptotic optimal limiting distribution is a Gaussian law
with covariance $(I - \SpecMat)^{-1} \SigStar (I - \SpecMat)^{-
  \top}$. (See the paper~\cite{khamaru2020temporal} for a detailed
analysis of this type in the special case of policy evaluation in
tabular MDPs.) This optimality result holds in a ``local'' sense: it
is minimax optimal in a small neighborhood of radius $O (1/\sqrt{n})$
around a given problem instance $(\SpecMatZero,
\projectedbvecZero)$. Theorem~\ref{thm:linear-stat-error-lower-bound},
on the other hand, is non-asymptotic, showing that a similar result
holds provided $\numobs$ is lower bounded by an explicit,
problem-dependent quantity of the order $\sigmaA^2 \usedim \opnorm{(I
  - \SpecMatZero)^{-1}}^2$. This accommodates a broader range of
sample sizes than the upper bound in
Theorem~\ref{thm:linear-oracle-ineq}.


\subsubsection{Combining the bounds}

Having presented separate lower bounds on the approximation and
estimation errors in conjunction with definitions of local classes of
instances over which they hold, we are now ready to present a
corollary which combines the two lower bounds in
Theorems~\ref{thm:linear-lower-bound}
and~\ref{thm:linear-stat-error-lower-bound}.

We begin by defining the local classes of instances over which our
combined bound holds.  Given a matrix-vector pair $(\SpecMat_0,
\projectedbvec_0)$, covariance matrices $(\Sigma_\Lmat,
\Sigma_\bvec)$, ambient dimension $\bigdim > 0$, and scalars
$\oracleErr, \conmax, \sigmaA, \sigmab > 0$, we begin by specifying a
collection of mean pairs $(\Lmat, \bvec)$ via
\begin{align}
      \classFinal (\SpecMat_0, \projectedbvec_0, \bigdim,
      \oracleErr, \conmax) \mydefn \bigcup_{(\SpecMat',
          \projectedbvec') \in \neighborhood_\numobs (\SpecMat_0,
          \projectedbvec_0)} \classApprox (\SpecMat', \projectedbvec',
      \bigdim, \oracleErr, \conmax). \label{eq:final-func-class-def}
\end{align}
Clearly, this represents a natural combination of the classes
$\classApprox$ and $\classEst$ introduced above.  We use the shorthand
$\classFinal$ for this class for brevity. Our collection of
distributions $\ProbInst_{\Lmat, \bvec}$ is still given by the class
$\classCov$ from equation~\eqref{defn:classcov}.

With these definitions in hand, we are now ready to state our combined
lower bound.
\begin{corollary}
\label{cor:lower-bound-final}
Under the setup above, suppose that the pair $(\sigmaA, \sigmab)$
satisfies the conditions in Theorem~\ref{thm:linear-lower-bound} and
equation~\eqref{eq:compatibility-condition-cov}, and that the matrix
$\SpecMat_0$ satisfies $\opnorm{\SpecMat_0} \leq \conmax - \sigmaA
\sqrt{\usedim / \numobs}$. Moreover, suppose that the sample size and
ambient dimension satisfy $\numobs \geq 16 \sigmaA^2 \opnorm{(I -
  \SpecMat_0)^{-1}}^2 \usedim$ and $\bigdim \geq \usedim + 36
\numobs^2$, respectively. Then the following minimax lower bound holds
for a universal positive constant $c$:
\begin{align*}
\inf_{\vhat_n \in \Vhatclass_\Xspace } ~\sup_{\substack{(\Lmat, \bvec)
    \in \classFinal\\ \ProbInst_{\Lmat, \bvec} \in \classCov}} \Exs
\statnorm{\vhat_n - \vstar}^2 \geq c \cdot \bigg \{ \big(\prefact
(\SpecMatZero, \conmax) - 1\big) \cdot \delta^2 + \EstErr_\numobs
(\SpecMatZero, \Sigma_\Lmat + \Sigma_\bvec) \bigg\}.
\end{align*}
\end{corollary}
\noindent We prove this corollary in
Section~\ref{subsec:proof-of-corollary-combined}.  It is relatively
straightforward consequence of combining
Theorems~\ref{thm:linear-lower-bound}
and~\ref{thm:linear-stat-error-lower-bound}.  \\

\vspace*{0.02in}

The combined lower bound matches the expression $\prefact
(\SpecMatZero, \conmax) \AppErr (\LinSpace, \vstar) + \EstErr_\numobs
(\SpecMatZero, \Sigma_\Lmat + \Sigma_\bvec)$, given by the first two
terms of Theorem~\ref{thm:linear-oracle-ineq}, up to universal constant
factors. Recall from our discussion of
Theorem~\ref{thm:linear-oracle-ineq} that the high-order term
$\BigResErr$ represents the ``optimization error'' of the stochastic
approximation algorithm, which depends on the coercive condition
$\kappa (\SpecMatZero)$ instead of the natural geometry $I -
\SpecMatZero$ of the problem. While we do not expect this term to
appear in an information-theoretic lower bound, the leading estimation
error term $ \EstErr_\numobs (\SpecMatZero, \Sigma_\Lmat +
\Sigma_\bvec)$ will dominate the high-order term when the sample size
$\numobs$ is large enough. For such a range of $\numobs$, the bound in
Theorem~\ref{thm:linear-oracle-ineq} is information-theoretically
optimal in the local class specified above. More broadly, consider the
class of all instances satisfying
Assumption~\ref{assume-second-moment-strong}, with $\kappa (\SpecMat)
\leq \kappa$ and $\hilopnorm{\Lmat} \leq 1$. Then the bound in
Theorem~\ref{thm:linear-oracle-ineq} is optimal, in a worst-case
sense, over this class as long as the sample size exceeds the
threshold $\frac{c \sigmaA^2}{(1 - \kappa)^2} \usedim$.


\subsubsection{The intermediate regime}
\label{subsubsec:intermediate-regime}

It remains to tie up some loose ends.  Note that the lower bound in
Theorem~\ref{thm:linear-lower-bound} requires a condition $D \gg
n^2$. On the other hand, it is easy to see that the approximation
factor can be made arbitrarily close to $1$ when $n \gg D$. (For
example, one could run the estimator based on stochastic approximation
and averaging---which was analyzed in
Theorem~\ref{thm:linear-oracle-ineq}---with the entire Euclidean space
$\Xspace$, and project the resulting estimate onto the subspace
$\lspace$.) In the middle regime $n \ll D \ll n^2$, however, it is not
clear which estimator is optimal.

In the following theorem, we present a lower bound for the
approximation factor in this intermediate regime, which establishes
the optimality of Theorem~\ref{thm:linear-oracle-ineq} up to a
constant factor.

\begin{theorem}
\label{thm:lower-bound-approx-factor-improved}
Suppose $\SpecMatZero \in \real^{d \times d}$ is a matrix such that $I
- \SpecMatZero$ is invertible, and that the scalars $(\sigmaA,
\sigmab)$ satisfy $\sigmaA \geq 1 + \conmax$ and $\sigmab \geq
\oracleErr$.  If the ambient dimension satisfies $\bigdim \geq \usedim
+ 3 q \numobs^{1 + 1/q}$ for some integer $q \in \left[2, \log n
  \wedge \frac{1}{\sqrt{2 (1 - \conmax \wedge 1)}} \right]$, then we
have the lower bound
\begin{align*}
  \inf_{\vhat_n \in \Vhatclass_\Xspace} ~\sup_{\substack{ (\Lmat,
      \bvec) \in \classApprox \\ \ProbInst_{\Lmat, \bvec} \in
      \classNoise (\sigmaA, \sigmab) }} \Exs \statnorm{\vhat_n -
    \vstar}^2 \geq \frac{\prefact (\SpecMat, \conmax) - 1}{4
    q^2}\oracleErr^2.
\end{align*}
\end{theorem}
\noindent See Appendix~\ref{AppIntermediate} for the proof of this
theorem.

Theorem~\ref{thm:lower-bound-approx-factor-improved} resolves the gap
in the intermediate regime, up to a constant factor that depends on
$q$. In particular, the stochastic approximation
estimator~\eqref{eq:SA} for projected equations still yields a
near-optimal approximation factor.  Compared to
Theorem~\ref{thm:linear-lower-bound},
Theorem~\ref{thm:lower-bound-approx-factor-improved} weakens the
requirement on the ambient dimension $\bigdim$ and covers the entire
regime $D \gg n$. Furthermore, using the same arguments as in
Corollary~\ref{cor:lower-bound-final}, this theorem can also be
combined with Theorem~\ref{thm:linear-stat-error-lower-bound} to
obtain the following lower bound in the regime $\bigdim \geq \usedim +
3 q \numobs^{1 + 1/q}$, for any integer $q > 0$:
\begin{align*}
  \inf_{\vhat_n \in \Vhatclass_\Xspace } ~\sup_{\substack{(\Lmat,
      \bvec) \in \classFinal\\ \ProbInst_{\Lmat, \bvec} \in
      \classCov}} \Exs \statnorm{\vhat_n - \vstar}^2 \geq c \cdot
  \bigg \{ \frac{\prefact (\SpecMatZero, \conmax) - 1}{q^2} \cdot
  \delta^2 + \EstErr_\numobs (\SpecMatZero, \Sigma_\Lmat +
  \Sigma_\bvec) \bigg\}.
\end{align*}

Let us summarize our approximation factor lower bounds in the various
regimes.  Consider a sequence of problem instances $\big(
\ProbInst_{\Lmat, \bvec}^{(\numobs)} \big)_{\numobs = 1}^{\infty}$
with increasing ambient dimension $\bigdim_\numobs$. Let the projected
dimension $\usedim$, noise variances $(\sigmaA, \sigmab)$, oracle
error $\oracleErr$, projected matrix $\PhiOp \Lmat^{(\numobs)}
\adjoint{\PhiOp} = \SpecMat$, and the operator norm bound
$\hilopnorm{\Lmat} \leq \conmax$ be all fixed. The following table
then presents a combination of our results from
Theorems~\ref{thm:linear-oracle-ineq},~\ref{thm:linear-lower-bound},
and~\ref{thm:lower-bound-approx-factor-improved}; our results suggest
that the optimal approximation factor exhibits a ``slow'' phase
transition phenomenon.
\begin{table}[htb]
    \centering
    \begin{tabular}{|c|c|c|c|}
    \hline $q = \lim_{\numobs \rightarrow \infty} \frac{\log
      \bigdim_\numobs}{\log \numobs}$ & $[2, \infty)$ & $(1, 2)$ &
      $(0, 1)$ \\ \hline Lower bound & $\prefact (\SpecMatZero,
      \conmax)$ & $c_q \cdot \prefact (\SpecMatZero, \conmax)$ &
      $1$\\ \hline Upper bound & $\prefact (\SpecMatZero, \conmax)$ &
      $\prefact (\SpecMatZero, \conmax)$ & $1$\\ \hline
    \end{tabular}
    \caption{Bounds on the approximation factor $\frac{\Exs
        \statnorm{\vhat_\numobs - \vstar}^2}{\AppErr (\LinSpace,
        \vstar)}$ for proper estimators in different ranges of ambient
      dimension. Here, $c_q \in (0, 1)$ represents a constant
      depending only on the aspect ratio $q$.}
    \label{table:phase-transition-approx-factor}
\end{table}
It is an interesting open question whether the phase transition is
sharp, and to identify the asymptotically optimal approximation factor
in the regime $\lim_{n \rightarrow \infty} \frac{\log
  \bigdim_\numobs}{\log \numobs} = 1$ since our lower bounds do not
apply in this linear regime.


\section{Consequences for specific models}
\label{SecConsequences}

We now discuss the consequences of our main theorems for the three
examples introduced in Section~\ref{SecExamples}.  For brevity, we
state only upper bounds for the first two examples; our third example
for temporal difference learning methods includes both upper and lower
bounds.


\subsection{Linear regression}

Recall the setting of linear regression from
Section~\ref{example:linear-regression}, including our
i.i.d. observation model~\eqref{eq:linear-regression-model}.  We
assume bounds on the second moment of $\varepsilon$ and fourth moment
of $X$---namely, the existence of some $\specpar > 0$ such that
\begin{align}
\label{eq:linear-regression-assumptions}  
\Exs \inprod{u}{X}^4 \leq \specpar^4, \quad \mbox{and} \quad \Exs
(\varepsilon^2) \leq \specpar^2 \qquad \mbox{for all $u \in \sphere^{D
    - 1}$.}
\end{align}
These conditions ensure that Assumption~\ref{assume-second-moment} is
satisfied with $(\sigma_\Lmat, \sigma_{\bvec}) = (\smooth^{-1} \specpar^2,
\smooth^{-1} \specpar^2)$.\footnote{Note that the stochastic approximation iterates are invariant under translation, and consequently we can assume without loss of generality that $\vbar = 0$.}  Recall that the (unprojected) covariance matrix
satisfies the PSD relations $\mu I \preceq \Exs [XX^\top]\preceq \beta
I$, and define the $\usedim$-dimensional covariance matrix $\Sigma
\mydefn \Exs \left[ (\PhiOp X) (\PhiOp X)^\top \right]$ for
convenience.

In this case, our stochastic approximation
iterates~\eqref{eq:lsa-iterates} take the form
\begin{align}
\label{eq:linear-regression-lsa}  
\vvec_{t + 1} = \vvec_t - \stepsize \left( \projecttolin X_{t + 1}
X_{t + 1}^\top \projecttolin \vvec_t + Y_{t + 1} \projecttolin X_{t +
  1} \right), \quad \mbox{for all $t = 0,1,2, \ldots,$}
\end{align}
and we take the averaged iterates $\vhat_\numobs \mydefn
\frac{2}{\numobs} \sum_{t = \numobs/2}^{\numobs - 1} \vvec_t$.  For
this procedure, we have the following guarantee:
\begin{corollary}
  \label{corr:oracle-ineq-linear-regression}
Suppose that we have $n$ i.i.d. observations $\{ (X_i, Y_i)\}_{i =
  1}^\numobs$ from the model~\eqref{eq:linear-regression-model}
satisfying the moment
conditions~\eqref{eq:linear-regression-assumptions}. Then there are
universal positive constants $(c, c_0)$ such that given a sample size
\mbox{$\numobs \geq \frac{c_0 \specpar^4 d}{\lambda^2_{\min} (\Sigma)}
  \log^2 \left( \frac{\smooth}{\strongconvex} \vecnorm{\vvec_0 -
    \vbar}{2}^2 d \right)$}, if the stochastic approximation
scheme~\eqref{eq:linear-regression-lsa} is run with step size
$\stepsize = \frac{1}{c_0 \specpar^2 \sqrt{d \numobs}}$, then the
averaged iterate satisfies the bound
\begin{multline*}
 \Exs \vecnorm{\vhat_\numobs - \vstar}{2}^2 \leq (1 + \offpar) \cdot
 \prefact \left(I_d - \frac{\Sigma}{\smooth}, 1 -
 \frac{\strongconvex}{\smooth} \right) \AppErr (\LinSpace, \vstar) \\
 + c \cdot \frac{\trace (\Sigma^{-1}) \cdot
   \Exs(\varepsilon^2)}{\offpar \numobs} + \frac{c}{\offpar} \left(
 \frac{\specpar^2}{\lambda_{\min} (\Sigma)} \cdot
 \sqrt{\frac{d}{\numobs}} \right)^3
\end{multline*}
    for each $\offpar > 0$.
\end{corollary}
\noindent This result is a direct consequence of
Theorem~\ref{thm:linear-oracle-ineq} in application to this model. \\

Note that the statistical error term $ \frac{\trace (\Sigma^{-1})
  \cdot \Exs(\varepsilon^2)}{\numobs}$ in this case corresponds to the
classical statistical rates for linear regression in this
low-dimensional subspace. The approximation factor, by
Lemma~\ref{lem:approx-factor-symmetric} admits the closed-form
expression
\begin{align*}
    \prefact \left(I_d - \frac{\Sigma}{\smooth}, 1 -
    \frac{\strongconvex}{\smooth} \right) = \max_{i \in [d]}
    \frac{\strongconvex^2 + 2 \smooth (\lambda_i -
      \strongconvex)}{\lambda_i^2},
\end{align*}
where $\{ \lambda_j \}_{j=1}^d$ denote the eigenvalues of the matrix
$\Sigma$.  Since $\lambda_j \in [\strongconvex, \smooth]$ for each $j
\in [\usedim]$, the approximation factor is at most of the order $O
\left(\frac{\smooth}{\lambda_{\min} (\Sigma)}\right)$.

Compared to known sharp oracle inequalities for linear regression
(e.g.,~\cite{rigollet2015high}), the approximation factor in our bound
is not $1$ but rather a problem-dependent quantity. This is because we
study the \emph{estimation error} under the standard Euclidean metric
$\vecnorm{\cdot}{2}$, as opposed to the \emph{prediction error} under
the data-dependent metric $\vecnorm{\cdot}{L^2 (P_X)}$. When the
covariance matrix $\Exs [XX^\top]$ is identity, the approximation
factor $\prefact \big(I_d - \frac{\Sigma}{\smooth}, 1 -
\frac{\strongconvex}{\smooth} \big)$ is equal to $1$, recovering
classical results. Another error metric of interest, motivated by
applications such as transfer learning~\cite{li2020transfer}, is the
prediction error when the covariates $X$ follow a different
distribution $Q$. For such a problem, the result above can be modified
straightforwardly by choosing the Hilbert space $\Xspace$ to be
$\real^\bigdim$, equipped with the inner product $\statinprod{u}{v}
\mydefn u^\top \big( \Exs_{Q} [XX^\top] \big)^{-1} v$.


\subsection{Galerkin methods}

We now return to the example of Galerkin methods, as previously
introduced in Section~\ref{subsubsec:elliptic-example}, with the
i.i.d. observation model~\eqref{eq:obs-Galerkin}. We assume the basis
functions $\phi_1, \ldots, \phi_d$ to have uniformly bounded function
value and gradient, and define the scalars
\begin{align}
\label{eq:noise-Galerkin}
\sigmaA \mydefn \left(1 + \frac{2}{\smooth} \right) \max_{j \in
  [\usedim]} \sup_{x \in \domain} \vecnorm{\nabla \phi_j (x)}{2},
\quad \mbox{and} \quad \sigmab \mydefn
\frac{\vecnorm{\bfunc}{\Ltwospace} + 1}{\smooth} \max_{j \in
  [\usedim]} \sup_{x \in \domain} |\phi_j (x)|.
\end{align}
These boundedness conditions are naturally satisfied by many
interesting basis functions such as the Fourier basis\footnote{In the
typical application of finite-element methods, basis functions based
on local interpolation are widely
used~\cite{brenner2007mathematical}. These basis functions can have
large $\sup$-norm, but via application of the Walsh--Hadamard
transform, a new basis can be obtained satisfying
condition~\eqref{eq:noise-Galerkin} with dimension-independent
constants. Since the stochastic approximation algorithm is invariant
under orthogonal transformation, this modification is only for the
convenience of analysis and does not change the algorithm itself.},
and ensure---we verify this concretely in the proof of
Corollary~\ref{corr:elliptic} to follow---that our observation model
satisfies Assumption~\ref{assume-second-moment} with parameters
$(\sigmaA, \sigmab)$.

Taking the finite-dimensional representation $\vvec = \vartheta^\top
\phi$, the stochastic approximation estimator for solving
equation~\eqref{eq:elliptic-projected-linear-fixed-pt} is given by
\begin{align*}
\vartheta_{t + 1} &= \vartheta_t - \smooth^{-1} \stepsize \left(
\nabla \phi (x_{t + 1})^\top \afunc_{t + 1} \nabla \phi (x_{t + 1})
\vartheta_t - \bfunc_{t + 1} \phi (y_{t + 1}) \right), \quad
\mbox{for}~ t = 0,1, \cdots\\ \widehat{\vartheta}_\numobs &\mydefn
\frac{2}{\numobs} \sum_{t = \numobs/2}^{\numobs - 1} \vartheta_t ,
\quad \mbox{and} \quad \vhat_\numobs \mydefn
\widehat{\vartheta}_\numobs^\top \phi.
\end{align*}
In order to state our statistical guarantees for $\vhat_\numobs$, we define the following matrices:
\begin{align*}
\SpecMat &\mydefn I_\usedim - \smooth^{-1} \int_\domain \nabla \phi
(x) ^\top \afunc (x) \nabla \phi (x) dx,\\ \Sigma_\Lmat &\mydefn
\frac{1}{\smooth^2} \int_\domain \big(\nabla \phi\big)^\top \afunc
\nabla \vbar (\nabla \vbar)^\top \afunc \nabla \phi ~dx -
\frac{1}{\smooth^2} \left(\int_\domain \big(\nabla \phi \big)^\top
\afunc \nabla \vbar ~dx \right) \left(\int_\domain \big(\nabla \phi
\big)^\top \afunc \nabla \vbar ~dx \right)^\top\\ & \quad \quad +
\frac{1}{\smooth^2} \int_\domain (\nabla \phi)^\top \left[ (\nabla
  \vbar) (\nabla \vbar)^\top + \mathrm{diag} \big( \vecnorm{\nabla
    \vbar}{2}^2 - (\partial_j \vbar)^2 \big)_{j = 1}^{\domaindim}
  \right] (\nabla \phi) dx, \\ \Sigma_\bvec &\mydefn
\frac{1}{\smooth^2} \int_\domain \big( \bfunc (x)^2 + 1\big) \phi (x)
\phi (x)^\top~ dx - \frac{1}{\smooth^2} \left( \int_\domain f (x) \phi
(x) dx \right)\left( \int_\domain f (x) \phi (x) dx \right)^\top.
\end{align*}
With these definitions in hand, we are ready to state the consequence
of our main theorems to the estimation problem of elliptic equations.
\begin{corollary}
\label{corr:elliptic}
Under the setup above, there are universal positive constants $(c, c_0)$ such
that if $\numobs \geq \frac{c_0 \sigmaA^2 d}{(1 - \kappa
  (\SpecMat))^2} \log^2 \left( \frac{\statnorm{\vvec_0 - \vbar}^2
  \smooth d}{\strongconvex} \right)$ and the stochastic approximation
scheme is run with step size $\stepsize = \frac{1}{c_0 \sigmaA \sqrt{d
    \numobs}}$, then the averaged iterates satisfy
\begin{align*}
  &\Exs \vecnorm{\vhat_\numobs - \vstar}{\Xspace}^2 \\ &\quad \leq (1
  + \offpar) \prefact \left(\SpecMat, 1 -
  \frac{\strongconvex}{\smooth}\right) \inf_{\vvec \in \LinSpace}
  \vecnorm{\vvec - \vstar}{\Xspace}^2 + c \left(1 + \frac{1}{\offpar}
  \right) \cdot \left( \EstErr_\numobs (\SpecMat, \Sigma_\Lmat +
  \Sigma_\bvec) + \BigResErr \right)
\end{align*}
for any $\offpar > 0$.
\end{corollary}
\noindent See Appendix~\ref{Appendix:subsubsec-proof-of-corr-elliptic}
for the proof of this corollary.

Note that the approximation factor $\prefact \left(\SpecMat, 1 -
\frac{\strongconvex}{\smooth}\right)$ is uniformly bounded on the
order of $O (\smooth / \strongconvex)$, which recovers C\'{e}a's
energy estimates in the symmetric and uniform elliptic
case~\cite{cea1964approximation}. On the other hand, for a suitable
choice of basis vectors, the bound in Corollary~\ref{corr:elliptic}
can often be much smaller: the parameter $\strongconvex$ corresponding
to a global coercive condition can be replaced by the smallest
eigenvalue of the \emph{projected} operator $\SpecMat$. Furthermore,
our analysis can also directly extend to the more general asymmetric
and semi-elliptic case, for which case the global coercive condition
may not hold true.

It is also worth noting that the bound in
Corollary~\ref{corr:elliptic} is given in terms of Sobolev norm $\|
\cdot \|_{\Xspace} = \vecnorm{\cdot}{\sobolevone}$, as opposed to
standard $\Ltwospace$-norm used in the nonparametric estimation
literature. By the Poincar\'{e} inequality, and $\sobolevone$-norm
bound implies an $\Ltwospace$-norm bound, and ensures stronger error
guarantees on the gradient of the estimated function.


\subsection{Temporal difference learning}
\label{subsec:td-discounted}

We now turn to the final example previously introduced in
Section~\ref{ex:td-discounted}, namely that of the TD algorithm in
reinforcement learning.  Recall the i.i.d. observation
model~\eqref{eq:mrp-iid-observation-model}.  Also recall the
equivalent form of the projected fixed point
equation~\eqref{eq:lstd-in-low-dimensional-space}, and note that the
population-level operator $\Lmat$ satisfies the norm bound
\begin{align*}
\hilopnorm{\Lmat} = \discount \cdot \sup_{\statnorm{\valuefunc} \leq
  1} \statnorm{\transition \valuefunc} \leq \discount \mydefn \conmax,
\end{align*}
since $\stationary$ is the stationary distribution of the transition
kernel $\transition$.


\subsubsection{Upper bounds on stochastic approximation with averaging}

As mentioned before, this example is somewhat non-standard in that the
basis functions $\psi_i$ are not necessarily orthonormal; indeed the
classical \emph{temporal difference} (TD) learning update in $\real^d$
involves the stochastic approximation algorithm
\begin{subequations} \label{eq:TD0cor}
\begin{align}
    \vartheta_{t + 1} = \vartheta_t - \stepsize \left( \psi (s_{t +
      1}) \psi (s_{t + 1})^\top \vartheta_t - \discount \psi (s_{t +
      1}) \psi (s_{t + 1}^+)^\top \vartheta_t - R_{t + 1} (s_{t + 1})
    \psi (s_{t + 1}) \right). \label{eq:lsa-lstd-iid-discount}
\end{align}
The Polyak--Ruppert averaged estimator is then given by the relations
\begin{align}
     \widehat{\vartheta}_\numobs = \frac{2}{\numobs} \sum_{t = \numobs/2}^{\numobs - 1}
     \vartheta_t, \quad \mbox{and} \quad \widehat{\valuefunc}_n \mydefn \widehat{\vartheta}_\numobs^\top \psi. \label{eq:lsa-lstd-iid-prj}
\end{align}
\end{subequations}
Note that the updates~\eqref{eq:projected-bellman-discount} are, strictly speaking, different from the canonical iterates~\eqref{eq:projected-SA}, but this should not be viewed as a fundamental difference since we are ultimately interested in the value function iterates $\widehat{\valuefunc}_n$; these are obtained from the iterates $\widehat{\vartheta}_n$ by passing back to the original Hilbert space.

Nevertheless, this cosmetic difference necessitates some natural basis transformations before stating our results.
Define the matrix\footnote{Since the functions $\psi_i$ are linearly independent, we have $\Bmat \succ 0$.} $\Bmat \in \real^{d \times d}$ by
$\Bmat_{ij} \mydefn \statinprod{\psi_i}{\psi_j}$ for $i, j \in [d]$; this defines an orthonormal basis given by
\begin{align*}
    \left[ \begin{matrix} \phi_1& \phi_2 & \cdots&
        \phi_d \end{matrix}\right] \mydefn \left[ \begin{matrix}
        \psi_1& \psi_2 & \cdots& \psi_d \end{matrix}\right] \Bmat^{- 1/2}.
\end{align*}
Let
\begin{align*}
    \smooth \mydefn \lambda_{\max} (\Bmat)\quad \mbox{and} \quad \strongconvex \mydefn \lambda_{\min} \left(\Bmat \right),
\end{align*}
so that $\beta / \mu$ is the condition number of the covariance matrix of the features.

Having set up this transformation, we are now ready to state the
implication of our main theorem to the case of LSTD problems. We
assume the following fourth-moment condition:
\begin{align}
\label{eq:lstd-moment-assumption}
  \forall u \in \sphere^{\usedim - 1}, ~\Exs_{\stationary} \left(
  u^\top \Bmat^{-1/2} \psi (s) \right)^4 \leq \specpar^4, \quad
  \mbox{and} \quad \Exs_\stationary \left[ R^4(s) \right] \leq \specpar^4.
\end{align}
As verified in the proof of
Corollary~\ref{corr:oracle-ineq-lstd-iid-discount} to follow,
equation~\eqref{eq:lstd-moment-assumption} suffices to guarantee that
Assumption~\ref{assume-second-moment} is satisfied with parameters $(\sigmaA, \sigmab) = (2 \specpar^2, \specpar^2 / \sqrt{\smooth})$.  We also require
the following matrices to be defined:
    \begin{align*}
        M \mydefn \discount \Bmat^{-1/2} \Exs_{\stationary} [\psi (s) \psi
          (s^+)^\top] \Bmat^{-1/2}, &\quad \Sigma_\Lmat \mydefn
        \cov_{\stationary} \left[ \Bmat^{-1/2} \psi (s) \left( \psi (s) -
          \discount \psi (s^+) \right)^\top
          \bar{\vartheta}\right],\\ \Sigma_\bvec \mydefn &
        \cov_{\stationary} \left[ R (s) \Bmat^{- 1/2} \psi
          (s) \right].
    \end{align*}

The following corollary then provides a guarantee on the Polyak--Ruppert averaged TD(0) iterates~\eqref{eq:TD0cor}.
\begin{corollary}
 \label{corr:oracle-ineq-lstd-iid-discount}
 Under the set-up above, there are universal positive constants $(c, c_0)$ such
 that given a sample size $\numobs \geq \frac{c_0 \specpar^4 \smooth^2
   \usedim}{\strongconvex^2 (1 - \kappa (\SpecMat))^2} \log^2 \left(
 \frac{\vecnorm{\vvec_0 - \vbar}{2}^2 \smooth d}{\strongconvex (1 -
   \kappa (\SpecMat))} \right)$, then when the stochastic
 approximation scheme~\eqref{eq:lsa-lstd-iid-discount} is run with
 step size $\stepsize = \frac{1}{c_0 \specpar^2 \smooth \sqrt{ \usedim
     \numobs}}$, then the averaged iterates satisfy the bound
\begin{multline}
       \Exs \statnorm{\valuehat_\numobs - \valuestar}^2 \leq (1 +
       \offpar) \prefact (\SpecMat, \discount) \AppErr (\LinSpace, \valuestar) \\ + c \left(1 +
       \frac{1}{\offpar} \right) \left[ \EstErr_\numobs (\SpecMat,
         \Sigma_\Lmat + \Sigma_\bvec) + \big(1 +
         \statnorm{\valuebar}^2\big) \left( \frac{\specpar^2 \smooth}{(
           1- \kappa (\SpecMat))\strongconvex}
         \sqrt{\frac{\usedim}{\numobs}} \right)^3 \right]
\end{multline}
for any $\offpar > 0$.
\end{corollary}
\noindent See Appendix~\ref{App:proof-lstd-upper} for the proof of
this corollary.

In the worst case, the approximation factor $\prefact (\SpecMat,
\discount)$ scales as $\frac{1}{1 - \discount^2}$, recovering the
classical result~\eqref{eq:contraction-approx-factor-worst-case}, but
more generally gives a more fine-grained characterization of the
approximation factor depending on the one-step auto-covariance matrix
for the feature vectors. By
Lemma~\ref{lem:approx-factor-upper-bounds}, we have $\prefact
(\SpecMat, \discount) \leq O \big( \frac{1}{1 - \kappa (\SpecMat)}
\big)$, so intuitively, the approximation factor is large when the
Markov chain transitions slowly in the feature space along a certain
directions. On the other hand, if the one-step-transition is typically
a big jump, then approximation factor is smaller.

The statistical error term $ \EstErr_\numobs (\SpecMat, \Sigma_\Lmat +
\Sigma_\bvec)$ matches the Cram\'{e}r--Rao lower bound, and gives a
finer characterization than both worst-case upper
bounds~\cite{bhandari2018finite} and existing instance-dependent upper
bounds~\cite{lakshminarayanan2018linear}. Note that the final,
higher-order term depends on the condition number
$\frac{\smooth}{\strongconvex}$ of the covariance matrix $\Bmat$. This
ratio is $1$ when the basis vectors are orthonormal, but in general,
the speed of algorithmic convergence depends on this parameter.

\subsubsection{Approximation factor lower bounds for MRPs}

We conclude our discussion of discounted MRPs with an
information-theoretic lower bound for policy evaluation. This bound
involves technical effort over and above
Theorem~\ref{thm:linear-lower-bound} since our construction for MRPs
must make use only of operators $\Lmat$ that are constructed using a
valid transition kernel.
To set the stage,
we say that a Markov reward process $(\transition, \discount, \reward)$
and associated basis functions $\{ \psi_j\}_{j = 1}^d$ are in the
\emph{canonical set-up} if the following conditions hold:
\begin{itemize}
    \item The stationary distribution $\stationary$ of $\transition$ exists and is unique.
    \item The reward function and its observations are uniformly
      bounded. In particular, we have $\vecnorm{\reward}{\infty} \leq
      1$, and $\vecnorm{R}{\infty} \leq 1$ almost surely.
    \item The basis functions are orthonormal, i.e.,
      $\Exs_{\stationary} [\psi (s) \psi (s)^\top] = I_d$.
\end{itemize}
The three conditions are standard assumptions in Markov reward
processes.

Now given scalars $\kappamrp \in (0, 1]$ and $\discount \in (0, 1)$,
    integer $\bigdim > 0$ and scalar $\oracleErr \in (0, 1/2)$, we
    consider the following class of MRPs and associated feature
    vectors:
\begin{align*}
    \classMRP \left(\kappamrp, \discount, \bigdim, \oracleErr \right)
    \mydefn \left\{ (\transition, \discount, \reward, \psi)
    ~\Big|~ \begin{array}{c} \mbox{$ (\transition, \discount, \reward,
        \psi)$ is in the canonical setup,} \quad |\statespace| =
      \bigdim,\\ \AppErr (\LinSpace, \valuestar) \leq \oracleErr^2,
      \quad \kappa \left( \Exs_{\stationary} [\psi (s) \psi
        (s^+)^\top] \right) \leq \kappamrp .
    \end{array}
    \right\}.
\end{align*}
Note that under the canonical set-up, we have $\SpecMat = \discount
\Exs_{\stationary} [\psi (s) \psi (s^+)^\top]$, and consequently, a
problem instance in the class $\classMRP (\kappamrp, \discount,
\bigdim, \oracleErr)$ satisfies $\kappa (\SpecMat) \leq \kappamrp
\discount$ in the set-up of
Corollary~\ref{corr:oracle-ineq-lstd-iid-discount}. The condition
$\kappa \left( \Exs_{\stationary} [\psi (s) \psi (s^+)^\top] \right)
\leq \kappamrp$ can be seen as a ``mixing'' condition in the projected
space: when $\kappamrp$ is bounded away from $1$, the feature vector
cannot have too large a correlation with its next-step transition in
any direction.

We have the following minimax lower bound for this class, where we use
the shorthand $\classMRP \equiv \classMRP \left(\kappamrp, \discount,
\bigdim, \oracleErr \right)$ for convenience.
\begin{proposition}
\label{prop:mrp-approx-factor-lb}
There are universal positive constants $(c, c_1)$ such that if
$\bigdim \geq c_1 (\numobs^2 + \usedim)$, then for all scalars
$\kappamrp \in (0, 1]$ and $\discount \in (0, 1)$, we have
\begin{align}
\label{eq:lower-bound-AF-MRP}
\inf_{\widehat{\valuefunc}_\numobs \in \Vhatclass_\Xspace}
~\sup_{(\transition, \discount, \reward, \psi) \in \classMRP }
\statnorm{ \widehat{\valuefunc}_n - \valuestar }^2 \geq \frac{c}{1 -
  \kappamrp \discount} \oracleErr^2 \wedge 1.
    \end{align}
\end{proposition}

\noindent See Appendix~\ref{App:proof-lstd-lower} for the proof of
this proposition.

A few remarks are in order. First, in conjunction with
Corollary~\ref{corr:oracle-ineq-lstd-iid-discount} and the second
upper bound in Lemma~\ref{lem:approx-factor-upper-bounds}, we can
conclude that the TD algorithm for policy evaluation with linear
function approximation attains the minimax-optimal approximation
factor over the class $\classMRP$ up to universal constants. It is
also worth noting that Proposition~\ref{prop:mrp-approx-factor-lb}
also shows that the worst-case upper
bound~\eqref{eq:contraction-approx-factor-worst-case} due to
Tsitsiklis and Van Roy~\cite{tsitsiklis1997analysis} is indeed sharp
up to a universal constant; indeed, note that for all $\discount \in
(0, 1)$, we have $\frac{1}{1 - \discount^2} \asymp \frac{1}{1 -
  \discount}$, and that the latter factor can be obtained from the
lower bound~\eqref{eq:lower-bound-AF-MRP} by taking $\kappamrp = 1$.

Second, note that the class $\classMRP$ is defined in a more
``global'' sense, as opposed to the ``local'' class $\classApprox$
used in Theorem~\ref{thm:linear-lower-bound}. This class contains all
the MRP instances satisfying the approximation error bound and the
constraint on $\kappa (\SpecMat)$, and a minimax lower bound over this
larger class is weaker than the lower bound over the local class that
imposes restrictions on the projected matrix. That being said,
Proposition~\ref{prop:mrp-approx-factor-lb} still captures more
structure in the Markov transition kernel than the fact that it is
contractive in the $\xi$-norm. For example, when the Markov chain
makes ``local moves'' in the feature space, the correlation between
feature vectors can be large, leading to large value of $\kappamrp$
and larger values of optimal approximation factor. On the other hand,
if the one-step transition of the feature vector jumps a large
distance in all directions, the optimal approximation factor will be
small.

Finally, it is worth noticing that
Proposition~\ref{prop:mrp-approx-factor-lb} holds true only for the
$\mathrm{i.i.d.}$ observation models. If we are given the entire
trajectory of the Markov reward process, the approximation factor can
be made arbitrarily close to 1, using TD$(\lambda)$
methods~\cite{tsitsiklis1997analysis}. The trade-off inherent to the
Markov observation model is left for our companion paper.


\section{Proofs}
\label{SecProofs}

We now turn to the proofs of our main results.


\subsection{Proof of Theorem~\ref{thm:linear-oracle-ineq}}
\label{SecProofThmLinearOracleIneq}

We divide the proof into two parts, corresponding to the two
components in the mean-squared error of the estimator $\vhat_\numobs$.
The first term is the \emph{approximation error} $\statnorm{\vbar -
  \vstar}^2$ that arises from the difference between the exact
solution $\vstar$ to the original fixed point equation, and the exact
solution $\vbar$ to the projected set of equations.  The second term
is the \emph{estimation error} $\Exs \statnorm{\vhat_\numobs -
  \vbar}^2$, measuring the difficulty of estimating $\vbar$ on the
basis of $\numobs$ noisy samples.

In particular, under the conditions of the theorem, we prove that the
approximation error is upper bounded as
\begin{subequations}
  \begin{align}
 \label{eq:linear-approx-factor}    
    \statnorm{\vbar - \vstar}^2 & \leq \prefact(\SpecMat, \hilopnorm{\Lmat}) \,
    \inf_{\vvec \in \lspace} \statnorm{\vvec - \vstar}^2,
  \end{align}
whereas the estimation error is bounded as
\begin{align}  
\label{eq:linear-statistical-error}    
    \Exs \statnorm{\vhat_\numobs - \vbar}^2 & \leq c \frac{\trace
      \left( (I - \SpecMat)^{-1} \SigStar (I - \SpecMat)^{-\top}
      \right)}{\numobs} + c\frac{\sigmaA}{(1 - \kappa)^3} \left(
    \frac{\usedim}{\numobs} \right)^{3/2} \left( \statnorm{\vbar}^2
    \sigmaA^2 + \sigmab^2 \right).
\end{align}
\end{subequations}

Given these two inequalities, it is straightforward to prove the
bound~\eqref{eq:thm1-rate} stated in the theorem.  By expanding the
square, we have
\begin{align*}
  \Exs \statnorm{\vhat_\numobs - \vstar}^2 & = \Exs
  \statnorm{\vhat_\numobs - \vbar}^2 + \statnorm{\vbar - \vstar}^2 +
  2 \Exs \statinprod{\vhat_\numobs - \vbar}{\vbar - \vstar}\\
  & \stackrel{(i)}{\leq} \Exs \statnorm{\vhat_\numobs - \vbar}^2 +
  \statnorm{\vbar - \vstar}^2 + 2 \sqrt{\Exs \statnorm{\vhat_\numobs
      - \vbar}^2 \cdot \statnorm{\vbar - \vstar}^2}\\
  & \stackrel{(ii)}{\leq} \Exs \statnorm{\vhat_\numobs - \vbar}^2 +
  \statnorm{\vbar - \vstar}^2 + \tfrac{1}{\offpar} \Exs \statnorm{\vhat_\numobs
    - \vbar}^2 + \offpar  \statnorm{\vbar - \vstar}^2 \\
& = (1 + \offpar) \statnorm{\vbar - \vstar}^2 +   (1 +
  \tfrac{1}{\offpar}) \Exs \statnorm{\vhat_\numobs - \vbar}^2
\end{align*}
where step (i) follows from the Cauchy--Schwarz inequality; and step
(ii) follows from the arithmetic-geometric mean inequality, and is
valid for any $\offpar > 0$.  Substituting the bounds from
equations~\eqref{eq:linear-approx-factor}
and~\eqref{eq:linear-statistical-error} yields the claim of the
theorem. \\

\noindent The remainder of our argument is devoted to the proofs of
the bounds~\eqref{eq:linear-approx-factor}
and~\eqref{eq:linear-statistical-error}.


\subsubsection{Proof of approximation error bound~\eqref{eq:linear-approx-factor}}

We begin with some decomposition relations for vectors and
operators. Note that $\lspace$ is a finite-dimensional subspace, and
therefore is closed. We use
\begin{align*}
\lspace^\perp \mydefn \{u \in \Xspace \, \mid \, \statinprod{u}{v} = 0
\mid \mbox{for all $v \in \lspace$.} \big\}
\end{align*}
to denote its orthogonal complement. The pair
$(\lspace, \lspace^\perp)$ forms a direct product decomposition of
$\Xspace$, and the projection operators satisfy $\projecttolin +
\projecttoorth = I$. Also define the operators
$\Lmat_{\lspace, \lspace} = \projecttolin \Lmat \projecttolin$ and
$\Lmat_{\lspace, \perp} = \projecttolin \Lmat \projecttoorth$.  With
this notation, our proof can be broken down into two auxiliary lemmas,
which we state here:
\begin{lemma}
  \label{LemCandlelight}
  The error $\statnorm{\vbar - \vstar}$ between the projected fixed
  point $\vbar$ and the original fixed point $\vstar$ is bounded as
\begin{align}  
    \statnorm{\vbar - \vstar}^2 & \leq \left( 1 + \hilopnorm{(I -
      \Lmat_{\lspace, \lspace})^{-1} \Lmat_{\lspace, \perp}}^2 \right)
    \; \inf_{\vvec \in \LinSpace} \statnorm{\vvec - \vstar}^2.
\end{align}
\end{lemma}

\begin{lemma}
\label{LemDinosaur}
Under the set-up above, we have
\begin{align*}
\hilopnorm{(I - \Lmat_{\lspace, \lspace})^{-1} \Lmat_{\lspace,
    \perp}}^2 & \leq \lammax \Big( (I_\usedim - \SpecMat)^{-1} \Big(
\hilopnorm{\Lmat}^2 I_\usedim - \SpecMat \SpecMat^\top \Big) (I_\usedim -
\SpecMat)^{-\top} \Big).
    \end{align*}
\end{lemma}

\noindent The claimed bound~\eqref{eq:linear-approx-factor} on the
approximation error follows by combining these two lemmas, and
recalling our definition of $\prefact(\SpecMat, \Lmat)$.  We now
prove these two lemmas in turn.

\subsubsection{Proof of Lemma~\ref{LemCandlelight}}
\label{SecProofCandlelight}

For any vector $v \in \Xspace$, we perform the orthogonal
decomposition $v = v_\lspace + v_\perp$, where $v_\lspace \mydefn
\projecttolin(v)$ is a member of the set $\lspace$, and $v_\perp
\defn \Pi_{\lspace^{\perp}, \xi}$ is a member of the set
$\lspace^\perp$.  With this notation, the operator $\Lmat$ can be
decomposed as
\begin{align*}
\Lmat & = (\projecttolin + \projecttoorth) \Lmat (\projecttolin +
\projecttoorth) \; = \; \underbrace{\projecttolin \Lmat
  \projecttolin}_{=: \Lmat_{\lspace, \lspace}} \; + \;
\underbrace{\projecttolin \Lmat \projecttoorth}_{=: \Lmat_{\lspace,
    \perp}} \; + \; \underbrace{\projecttoorth \Lmat
  \projecttolin}_{=: \Lmat_{\perp, \lspace}} \; + \;
\underbrace{\projecttoorth \Lmat \projecttoorth}_{=: \Lmat_{\perp,
    \perp}}.
\end{align*}

The four operators $\Lmat_{\lspace, \lspace}, \Lmat_{\lspace, \perp},
\Lmat_{\perp, \lspace}, \Lmat_{\perp, \perp}$ defined in the equation
above are also bounded linear operators. By the properties of
projection operators, we note that $\Lmat_{\lspace, \lspace}$ and
$\Lmat_{\perp, \lspace}$ both map each element of $\lspace^\perp$ to
$0$, and $\Lmat_{\lspace, \perp}$ and $\Lmat_{\perp, \perp}$ both map
each element of $\lspace$ to $0$.

Decomposing the target vector $\vstar$ in an analogous manner yields
the two components
\begin{align*}
\vtil \mydefn \projecttolin (\vstar), \quad \mbox{and} \quad
\vvecperp \mydefn \vstar - \vtil.
\end{align*}
The fixed point equation $\vstar = \Lmat \vstar + \bvec$ can then be
written using $\lspace$ and its orthogonal complement as
  \begin{align}
\label{eq:approxA}    
        \vtil \stackrel{(a)}{=} \Lmat_{\lspace, \lspace} \vtil +
        \Lmat_{\lspace, \perp} \vvec^\perp + \bvec_\lspace, \quad
        \mbox{and} \quad \vvec^\perp \stackrel{(b)}{=}
        \Lmat_{\perp, \lspace} \vtil + \Lmat_{\perp, \perp}
        \vvec^{\perp} + b_\perp.
\end{align}
For the projected solution $\vbar$, we have the defining equation
\begin{align}
\label{eq:thetabar-defn}
    \vbar = \Lmat_{\lspace, \lspace} \vbar + b_\lspace.
\end{align}
Subtracting equation~\eqref{eq:approxA}(a) from
equation~\eqref{eq:thetabar-defn} yields
\begin{align*}
 (I - \Lmat_{\lspace, \lspace}) (\vtil - \vbar) = \Lmat_{\lspace,
    \perp} \vvecperp.
\end{align*}
Recall the quantity $\SpecMat = \PhiOp \Lmat \adjoint{\PhiOp}$, and
our assumption that $\kappa(\SpecMat) = \tfrac{1}{2} \lammax(\SpecMat
+ \SpecMat^T) < 1$.  This condition implies that $I -
\Lmat_{\lspace, \lspace}$ is invertible on the subspace $\lspace$.
Since this operator also maps each element of $\lspace^\perp$ to
itself, it is invertible on all of $\Xspace$, and we have $\vtil -
\vbar = (I - \Lmat_{\lspace, \lspace})^{-1} \Lmat_{\lspace, \perp}
\vvecperp$.

Applying the Pythagorean theorem then yields
\begin{align}
    \statnorm{\vbar - \vstar}^2 \; = \; \statnorm{\vbar - \vtil}^2 +
    \statnorm{\vtil - \vstar}^2 & = \statnorm{(I -
      \Lmat_{\lspace, \lspace})^{-1} \Lmat_{\lspace, \perp}
      \vvecperp}^2 + \statnorm{ \vvecperp}^2 \notag \\
 \label{eq:pythagorean-approx-factor-upper-bound}
    & \leq \left( 1 + \hilopnorm{(I -
   \Lmat_{\lspace, \lspace})^{-1} \Lmat_{\lspace, \perp}}^2 \right)
 \cdot \statnorm{\vvecperp}^2,
\end{align}
as claimed.


\subsubsection{Proof of Lemma~\ref{LemDinosaur}}
\label{SecProofLemDinosaur}

By the definition of operator norm for any vector $v \in \Xspace$ such
that $\statnorm{v} = 1$, we have
\begin{align*}
    \hilopnorm{\Lmat}^2 \geq \statnorm{\Lmat v}^2 =
    \statnorm{\Lmat_{\lspace, \lspace} v_\lspace + \Lmat_{\lspace,
        \perp} v_\perp}^2 + \statnorm{\Lmat_{\perp, \lspace}
      v_{\lspace} + \Lmat_{\perp, \perp} v_\perp}^2 \geq
    \statnorm{\Lmat_{\lspace, \lspace} v_\lspace + \Lmat_{\lspace,
        \perp} v_\perp}^2.
\end{align*}
Noting the fact that $\Lmat_{\lspace, \lspace} v_\perp = 0 =
\Lmat_{\lspace, \perp} v_\lspace$, we have the following norm bound on
the linear operator $\Lmat_{\lspace, \lspace} + \Lmat_{\lspace,
  \perp}$:
\begin{align*}
  \hilopnorm{\Lmat_{\lspace, \lspace} + \Lmat_{\lspace, \perp}} &
  = \sup_{\statnorm{v} = 1} \statnorm{(\Lmat_{\lspace, \lspace} +
    \Lmat_{\lspace, \perp}) v} \\ & = \sup_{\statnorm{v} = 1}
  \statnorm{ \Lmat_{\lspace, \lspace} v_\lspace + \Lmat_{\lspace,
      \perp} v_\perp} \leq \hilopnorm{\Lmat}.
\end{align*}
By definition, the operator $\adjoint{\Lmat_{\lspace, \perp}} =
\projecttoorth \adjoint{\Lmat} \projecttolin$ maps any vector to
$\lspace^\perp$, and the operator $\Lmat_{\lspace, \lspace}$ maps any
element of $\lspace^\perp$ to $0$. Therefore, we have the identity $
\Lmat_{\lspace,\lspace} \adjoint{\Lmat_{\lspace, \perp}} = 0$.  A
similar argument yields that $ \Lmat_{\lspace,\perp}
\adjoint{\Lmat_{\lspace, \lspace}} = 0$. Consequently, we have
\begin{align}
    \hilopnorm{\Lmat}^2 \geq \hilopnorm{\Lmat_{\lspace, \lspace}
      + \Lmat_{\lspace, \perp}}^2 & =
    \hilopnorm{(\Lmat_{\lspace, \lspace} + \Lmat_{\lspace,
        \perp}) \adjoint{(\Lmat_{\lspace, \lspace} + \Lmat_{\lspace,
          \perp})}} \nonumber \\
\label{eq:approx-factor-upper-G}    
& = \hilopnorm{\underbrace{\Lmat_{\lspace,\lspace}
    \adjoint{\Lmat_{\lspace,\lspace}} + \Lmat_{\lspace,\perp}
    \adjoint{\Lmat_{\lspace,\perp}}}_{=: G}}.
\end{align}

Note that the operator $G$ can be expressed as $G = \projecttolin
\left( \Lmat \projecttolin \adjoint{\Lmat} + \Lmat \projecttoorth
\adjoint{\Lmat} \right) \projecttolin$.  From this representation, we
see that:
\bcar
\item For any vector $x \in \Xspace$, we have $G x\in \lspace$.
\item For any vector $y \in \lspace^\perp$, we have $G y = 0$.
  \ecar
Consequently, there exists a matrix $\widetilde{G} \in \real^{d \times
  d}$, such that $G = \adjoint{\PhiOp} \widetilde{G} \PhiOp$. Since
$G$ is a positive semi-definite operator, the matrix $\widetilde{G}$
is positive semi-definite.  Equation~\eqref{eq:approx-factor-upper-G}
implies that
\begin{align}
\label{eq:mat1}
\lammax (\widetilde{G}) = \opnorm{\widetilde{G}} = \hilopnorm{G}
\leq \hilopnorm{\Lmat}^2.
\end{align}
Now defining $\mytau \mydefn \hilopnorm{(I -
  \Lmat_{\lspace, \lspace})^{-1} \Lmat_{\lspace, \perp}}$, note that
\begin{align}
\label{eq:approx-factor-upper-H}  
    \mytau^2 = \hilopnorm{ \underbrace{(I -
        \Lmat_{\lspace, \lspace})^{-1} \Lmat_{\lspace, \perp}
        \adjoint{\Lmat_{\lspace, \perp}} (I -
        \adjoint{\Lmat_{\lspace, \lspace}})^{-1}}_{=: H} }.
\end{align}
Moreover, the operator $H$ is self-adjoint, and we have the following
properties:
\bcar
    \item The operator $\Lmat_{\lspace, \perp}$ maps any vector to
      $\lspace$, and $(I - \Lmat_{\lspace, \lspace})^{-1}$ maps
      $\lspace$ to itself. Consequently, for any $x \in \Xspace$, the
      vector $H x = (I - \Lmat_{\lspace, \lspace})^{-1}
      \Lmat_{\lspace, \perp} \left( \adjoint{\Lmat_{\lspace, \perp}}
      (I - \adjoint{\Lmat_{\lspace, \lspace}} )^{-1}\right) x$ is a
      member of the set $\lspace$.
    \item The operator $\adjoint{\Lmat_{\lspace, \perp}} =
      \projecttoorth \adjoint{\Lmat} \projecttolin$ maps any vector
      from $\lspace^\perp$ to $0$. Consequently, for any $y
      \in \lspace^{\perp}$, we have $H y = (I -
      \Lmat_{\lspace, \lspace})^{-1} \Lmat_{\lspace, \perp} \left(
      \adjoint{\Lmat_{\lspace, \perp}} (I -
      \adjoint{\Lmat_{\lspace, \lspace}} )^{-1}\right) y = 0$.
\ecar

Owing to the facts above, there exists a matrix $\widetilde{H} \in
\real^{d \times d}$, such that $H = \adjoint{\PhiOp} \widetilde{H}
\PhiOp$. Since the operator $H$ is positive semi-definite, so is the
matrix $\widetilde{H}$. Consequently, by
equation~\eqref{eq:approx-factor-upper-H}, we obtain the identity
$\mytau^2 = \hilopnorm{H} = \opnorm{\widetilde{H}} = \lambda_{\max}
(H)$.  In particular, letting $u \in \sphere^{d - 1}$ be a maximal
eigenvector of $\widetilde{H}$, we have
\begin{align}
    \widetilde{H} \succeq \mytau^2 uu^\top.\label{eq:mat2}
\end{align}
Since $\SpecMat = \PhiOp \Lmat_{\lspace, \lspace} \adjoint{\PhiOp}$ by
definition, combining the above matrix inequalities~\eqref{eq:mat1}
and~\eqref{eq:mat2}, we arrive at the bound:
\begin{align*}
    &\hilopnorm{\Lmat}^2 I_d \succeq \widetilde{G} \\ &= \PhiOp
  \left(\Lmat_{\lspace, \lspace} \adjoint{\Lmat_{\lspace, \lspace}} +
  \Lmat_{\lspace, \perp} \adjoint{\Lmat_{\lspace, \perp}} \right)
  \adjoint{\PhiOp}\\ &= \PhiOp \Lmat_{\lspace, \lspace}
  \adjoint{\Lmat_{\lspace, \lspace}} \adjoint{\PhiOp} + \left( \PhiOp
  (I - \Lmat_{\lspace, \lspace}) \adjoint{\PhiOp} \right) \cdot \left(
  \PhiOp (I - \Lmat_{\lspace, \lspace})^{-1} \Lmat_{\lspace, \perp}
  \adjoint{\Lmat_{\lspace, \perp}} (I -
  \adjoint{\Lmat_{\lspace, \lspace}})^{-1} \adjoint{\PhiOp} \right)
  \cdot \left( \PhiOp (I - \adjoint{\Lmat_{\lspace, \lspace}})
  \adjoint{\PhiOp} \right)\\ &= \SpecMat \SpecMat^\top + (I -
  \SpecMat) \widetilde{H} (I - \SpecMat^\top)\\ &\succeq \SpecMat
  \SpecMat^\top + \mytau^2 (I - \SpecMat) uu^\top (I - \SpecMat^\top).
\end{align*}
Re-arranging and noting that $u \in \sphere^{\usedim - 1},$ we arrive
at the inequality
\begin{align*}
    \mytau^2 &\leq u^\top \left[ (I - \SpecMat)^{-1}
      (\hilopnorm{\Lmat}^2 I_d - \SpecMat \SpecMat^\top) (I -
      \SpecMat)^{- \top} \right] u \; \leq \; \lammax \Big( (I -
    \SpecMat)^{-1} (\hilopnorm{\Lmat}^2 I_d - \SpecMat \SpecMat^\top)
    (I - \SpecMat)^{- \top} \Big),
\end{align*}
which completes the proof of Lemma~\ref{LemDinosaur}.


\subsubsection{Proof of estimation error bound~\eqref{eq:linear-statistical-error}}

We now turn to the proof of our claimed bound on the estimation error.
Our analysis relies on two auxiliary lemmas.  The first lemma provides
bounds on the mean-squared error of the standard iterates $\{
\vvec_t\}_{t \geq 0}$---that is, without the averaging step:
\begin{lemma}
\label{LemLinearSABound}
Suppose that the noise conditions in
Assumption~\ref{assume-second-moment} hold.  Then for any stepsize
$\eta \in \big(0, \frac{1 - \kappa}{4 \sigma_\Lmat^2 \usedim + 1 +
  \hilopnorm{\Lmat}^2} \big)$, we have the bound
\begin{align}
  \Exs \statnorm{\vvec_t - \vbar}^2 \leq e^{- (1 - \kappa) \stepsize
    t / 2}\Exs \statnorm{\vvec_0 - \vbar}^2 + \frac{8 \stepsize
  }{1 - \kappa} (\statnorm{\vbar}^2 \sigma_\Lmat^2 \usedim +
  \sigma_\bvec^2 d) \qquad \mbox{valid for $t = 1, 2, \ldots$.}
\end{align}
\end{lemma}
\noindent See Section~\ref{SecProofLemLinearSABound} for the proof of
this claim. \\

Our second lemma provides a bound on the PR-averaged estimate
$\vhat_\numobs$ based on $\numobs$ observations in terms of a
covariance term, along with the error of the non-averaged sequences
$\{\vvec_t\}_{t \geq 1}$:
\begin{lemma}
  \label{LemPrisonTime}
  Under the setup above, we have the bound
\begin{multline}
\label{EqnPrisonBound}    
    \Exs \statnorm{\vhat_\numobs - \vbar}^2 \leq \frac{6}{\numobs -
      \numburn} \trace \left( (I - \SpecMat)^{-1} \SigStar  (I -
    \SpecMat)^{- \top} \right) \\
    + \frac{6}{(\numobs - \numburn)^2} \sum_{t = \numburn}^\numobs \Exs \vecnorm{(I -
      \SpecMat)^{-1} \PhiOp (\Lmat_{t + 1} - \Lmat) (\vvec_t -
      \vbar)}{2}^2 + \frac{3\Exs \statnorm{\vvec_\numobs -
        \vvec_{\numburn}}^2 }{\stepsize^2 (\numobs - \numburn)^2 (1 - \kappa)^2}.
\end{multline}
\end{lemma}
\noindent See Section~\ref{SecProofLemPrisonTime} for the proof of
this claim. \\

Equipped with these two lemmas, we can now complete the proof of the
claimed bound~\eqref{eq:linear-statistical-error} on the estimation
error.  Recalling that $\numburn = \numobs/2$, we see that the first
term in the bound~\eqref{EqnPrisonBound} matches a term in the
bound~\eqref{eq:linear-statistical-error}.  As for the remaining two
terms in equation~\eqref{EqnPrisonBound}, the second moment bounds
from Assumption~\ref{assume-second-moment} combined with the
assumption that $\kappa(\SpecMat) < 1$ imply that
\begin{align*}
    \Exs \vecnorm{(I - \SpecMat)^{-1} \PhiOp (\Lmat_{t + 1} - \Lmat)
      (\vvec_t - \vbar)}{2}^2 & \leq \frac{1}{(1 - \kappa)^2} \Exs
    \vecnorm{\PhiOp (\Lmat_{t + 1} - \Lmat) (\vvec_t - \vbar)}{2}^2 \\
& \leq \frac{1}{(1 - \kappa)^2}\sum_{j = 1}^\usedim \Exs
    \statinprod{\phi_j}{(\Lmat_{t + 1} - \Lmat) (\vvec_t -
      \vbar)}^2 \\
& \leq \frac{\sigmaA^2 \usedim \statnorm{\vvec_t - \vbar}^2}{(1 - \kappa)^2}.
\end{align*}

On the other hand, we can use Lemma~\ref{LemLinearSABound} to control
the third term in the bound~\eqref{EqnPrisonBound}.  We begin by
observing that
\begin{align*}
\statnorm{\vvec_\numobs - \vvec_{\numburn}}^2 \leq
2\statnorm{\vvec_\numobs - \vbar}^2 + 2 \statnorm{\vvec_{\numburn} -
  \vbar}^2 \leq 4 \sup_{\numburn \leq t \leq \numobs} \Exs
\statnorm{\vvec_t - \vbar}^2.
\end{align*}
If we choose a burn-in time $\numburn > \frac{c_0}{(1 - \kappa)
  \stepsize} \log \left(\frac{\statnorm{\vvec_0 - \vbar}^2 d}{1 -
  \kappa} \right)$, then Lemma~\ref{LemLinearSABound} ensures that
\begin{align*}
    \sup_{\numburn \leq t \leq \numobs} \Exs \statnorm{\vvec_t -
      \vbar}^2 \leq \frac{16 \stepsize}{1 - \kappa} \left(
    \statnorm{\vbar}^2 \sigmaA^2 \usedim + \sigmab^2 \usedim \right).
\end{align*}

Finally, taking the step size $\stepsize = \frac{1}{24 \sigmaA \sqrt{d
    \numobs}}$, recalling that $\numburn = \numobs/2$, and putting
together the pieces yields
\begin{align*}
    \Exs \statnorm{\vhat_\numobs - \vbar}^2 &\leq \frac{12}{\numobs} \trace
    \left( (I - \SpecMat)^{-1} \SigStar (I - \SpecMat)^{-\top} \right) + \frac{1}{(1
      - \kappa)^2} \left( \frac{12 \sigma_\Lmat^2 d }{\numobs} +
    \frac{48}{\stepsize^2 \numobs^2} \right)\sup_{\numburn \leq t \leq \numobs} \Exs
    \statnorm{\vvec_t - \vbar}^2 \\ &\leq \frac{12}{\numobs} \trace \left(
    (I - \SpecMat)^{-1} \SigStar (I - \SpecMat)^{-\top} \right) + \frac{48
      \sigmaA}{(1 - \kappa)^3} \left( \frac{d}{\numobs} \right)^{3/2} \left(
    \statnorm{\vbar}^2 \sigmaA^2 + \sigmab^2 \right),
\end{align*}
as claimed. \\

\noindent It remains to prove our two auxiliary lemmas, which we do in the following
subsections.


\subsubsection{Proof of Lemma~\ref{LemLinearSABound}}
\label{SecProofLemLinearSABound}

We now prove Lemma~\ref{LemLinearSABound}, which provides a bound on
the error of the non-averaged iterates $\{\vvec_t\}_{t \geq 1}$, as
defined in equation~\eqref{eq:lsa-iterates}.  Using the form of the
update, we expand the mean-squared error to find that
\begin{align}
  \Exs \statnorm{\vvec_{t + 1} - \vbar}^2 & = \Exs \statnorm{(I -
    \stepsize I + \stepsize \projecttolin \Lmat) (\vvec_t - \vbar) +
    \stepsize \projecttolin (\Lmat_{t + 1} - \Lmat) \vvec_t +
    \stepsize \projecttolin (\bvec_{t + 1} - \bvec) }^2 \nonumber \\
  & \overset{(i)}{=} \Exs \statnorm{(I - \stepsize I + \stepsize
    \projecttolin \Lmat) (\vvec_t - \vbar)}^2 + \stepsize^2 \Exs
  \statnorm{ \projecttolin (\Lmat_{t + 1} - \Amat) \vvec_t +
    \projection_\phi (\bvec_{t + 1} - \bvec)}^2 \nonumber \\
\label{eq:lsa-iterate-error-recursion}  
  & \overset{(ii)}{\leq} (1 - \stepsize (1 - \kappa)) \Exs
  \statnorm{\vvec_t - \vbar}^2 + 2 \stepsize^2 \Exs \statnorm{
    \projecttolin (\Lmat_{t + 1} - \Lmat) (\vvec_t - \vbar)}^2
  \nonumber\\ &\quad \quad + 2 \stepsize^2 \Exs \statnorm{
    \projecttolin (\Lmat_{t + 1} - \Lmat) \vbar + \projecttolin
    (\bvec_{t + 1} - \bvec)}^2. 
\end{align}
In step (i), we have made use of the fact that the noise is unbiased,
and in step (ii), we have used that for any $\Delta$ in the subspace
$\lspace$ and any stepsize $\stepsize \in \big(0, \frac{1 - \kappa}{1
  + \hilopnorm{\Lmat}^2}\big) $, we have
\begin{align*}
    \statnorm{(I - \stepsize I + \stepsize \projecttolin \Lmat)
      \Delta}^2 & = (1 - \stepsize)^2 \statnorm{\Delta}^2 +
    \stepsize^2 \statnorm{ \projecttolin \Lmat \Delta}^2 + 2 (1 -
    \stepsize) \stepsize \statinprod{\Delta}{\projecttolin \Lmat
      \Delta} \\
& \leq \Big \{ 1 - 2 \stepsize + \stepsize^2 + \stepsize^2
    \hilopnorm{\Lmat}^2 + 2 (1 - \stepsize) \stepsize \kappa \Big \}
    \statnorm{\Delta}^2 \\
& \leq \big( 1 - \stepsize (1 - \kappa) \big) \statnorm{\Delta}^2.
\end{align*}
Turning to the second term of
equation~\eqref{eq:lsa-iterate-error-recursion}, the moment
bounds in 
Assumption~\ref{assume-second-moment} imply that
\begin{align*}
    \Exs \statnorm{\projecttolin (\Lmat_{t + 1} - \Lmat) (\vvec_t -
      \vbar)}^2 = \sum_{j = 1}^\usedim \Exs
    \statinprod{\phi_j}{(\Lmat_{t + 1} - \Lmat) (\vvec_t - \vbar)}^2
    \leq \Exs \statnorm{\vvec_t - \vbar}^2 \sigma_\Lmat^2 \usedim.
\end{align*}
Finally, the last term of
equation~\eqref{eq:lsa-iterate-error-recursion} is also handled by
Assumption~\ref{assume-second-moment}, whence we obtain
\begin{multline*}
  \Exs \statnorm{ \projecttolin (\Lmat_{t + 1} - \Lmat) \vbar +
    \projecttolin (\bvec_{t + 1} - \bvec)}^2 \leq 2 \sum_{j =
    1}^\usedim \Exs \statinprod{\phi_j}{(\Lmat_{t + 1} - \Lmat)
    \vbar}^2 \\ + 2 \sum_{j = 1}^\usedim \Exs
  \statinprod{\phi_j}{\bvec_{t + 1} - \bvec}^2 \leq 2
  \statnorm{\vbar}^2 \sigma_\Lmat^2 \usedim + 2 \sigma_\bvec^2
  \usedim.
\end{multline*}
Putting together the pieces, we see that provided $\stepsize < \frac{1
  - \kappa}{4 \sigma_\Lmat^2 \usedim + 1 +
  \hilopnorm{\Lmat}^2}$, we have
\begin{align*}
    \Exs \statnorm{\vvec_{t + 1} - \vbar}^2 &\leq (1 - \stepsize (1 -
    \kappa) + 2 \stepsize^2 \sigma_\Lmat^2 \usedim) \Exs
    \statnorm{\vvec_t - \vbar}^2 + 4 \stepsize^2 ( \statnorm{\vbar}^2
    \sigma_\Lmat^2 \usedim + \sigma_\bvec^2 \usedim) \\
& \leq \left(1 - \frac{\stepsize (1 - \kappa)}{2} \right) \Exs
    \statnorm{\vvec_t - \vbar}^2 + 4 \stepsize^2 ( \statnorm{\vbar}^2
    \sigma_\Lmat^2 \usedim + \sigma_\bvec^2 \usedim).
\end{align*}
Finally, rolling out the recursion yields the bound
\begin{align*}
    \Exs \statnorm{\vvec_\numobs - \vbar}^2 \leq e^{- (1 - \kappa)
      \stepsize \numobs / 2}\Exs \statnorm{\vvec_0 - \vbar}^2 + \frac{8
      \stepsize }{1 - \kappa} (\statnorm{\vbar}^2 \sigma_\Lmat^2 d +
    \sigma_\bvec^2 d),
\end{align*}
which completes the proof.


\subsubsection{Proof of Lemma~\ref{LemPrisonTime}}
\label{SecProofLemPrisonTime}

Recall that $\vbar$ satisfies the fixed point equation $\vbar =
\projecttolin \Lmat \vbar + \projecttolin \bvec$.  Using this fact, we
can derive the following elementary identity:
\begin{multline}
 \frac{\vvec_{\numburn} - \vvec_\numobs}{\stepsize (\numobs -
   \numburn)} = \frac{1}{\numobs - \numburn} \sum_{t =
   \numburn}^{\numobs - 1} \left( \vvec_t - \projecttolin \Lmat_{t +
   1} \vvec_t - \projecttolin \bvec_{t + 1} \right) \\
 = (I - \projecttolin \Lmat) (\vhat_\numobs - \vbar) +
 \frac{1}{\numobs - \numburn} \underbrace{\sum_{t = \numburn}^{\numobs
     - 1} \projecttolin (\Lmat_{t + 1} - \Lmat) \vvec_t}_{=:
   \martingale^{(1)}_\numobs} + \frac{1}{\numobs - \numburn}
 \underbrace{\sum_{t = \numburn}^{\numobs - 1} \projecttolin (\bvec_{t
     + 1} - \bvec)}_{=:
   \martingale^{(2)}_\numobs}. \label{eq:martingale-decomposition-proof-of-linear-upper-bound}
\end{multline}
Re-arranging terms and applying the Cauchy--Schwarz inequality, we
have
\begin{align*}
 \statnorm{\vhat_\numobs - \vbar}^2 \leq \frac{3}{(\numobs -
   \numburn)^2} \left( \frac{1}{\stepsize^2}\statnorm{(I -
   \projecttolin \Lmat)^{-1} (\vvec_\numobs - \vvec_{\numburn})}^2 +
 \statnorm{(I - \projecttolin \Lmat)^{-1} \martingale^{(1)}_\numobs}^2
 + \statnorm{(I - \projecttolin \Lmat)^{-1}
   \martingale^{(2)}_\numobs}^2 \right).
\end{align*}
Note that the quantities $\martingale^{(1)}_\numobs$ and
$\martingale^{(2)}_\numobs$ are martingales adapted to the filtration
\mbox{$\filtration_\numobs \mydefn \sigma (\{ \Lmat_i, \bvec_i
  \}_{i=1}^\numobs)$,} so that
\begin{multline*}
    \Exs \statnorm{\vhat_\numobs - \vbar}^2 \leq \frac{3}{(\numobs -
      \numburn)^2} \sum_{t = \numburn}^{\numobs - 1} \Exs \statnorm{(I
      - \projecttolin \Lmat)^{-1} \projecttolin (\Lmat_{t + 1} -
      \Lmat) \vvec_t}^2 \\
+ \frac{3}{(\numobs - \numburn)^2} \sum_{t = \numburn}^{\numobs - 1}
\Exs \statnorm{(I - \projecttolin \Amat)^{-1} \projecttolin(\bvec_{t +
    1} - \bvec)}^2 \\
+ \frac{3}{(\numobs - \numburn)^2 \stepsize^2} \Exs \statnorm{(I -
  \projecttolin \Lmat)^{-1} (\vvec_\numobs - \vvec_{\numburn})}^2.
\end{multline*}

We claim that for any vector $v \in \Xspace$, we have
\begin{align}
\label{eq:equivalent-form-between-subspace-and-xspace}  
  (I - \projecttolin \Lmat)^{-1} \projecttolin v = \adjoint{\PhiOp}
\left( (I - \SpecMat)^{-1} \PhiOp v \right).
\end{align}
Taking this claim as given for the moment, by applying
equation~\eqref{eq:equivalent-form-between-subspace-and-xspace} with
$v = (\Lmat_{t + 1} - \Lmat) \vvec_t$ and $v = \bvec_{t + 1} - \bvec$,
we find that
\begin{multline*}
    \Exs \statnorm{(I - \projecttolin \Lmat)^{-1} \projecttolin
      (\Lmat_{t + 1} - \Lmat) \vvec_t}^2 = \Exs \vecnorm{(I -
      \SpecMat)^{-1} \PhiOp (\Lmat_{t + 1} - \Lmat)
      \vvec_t}{2}^2\\ \leq 2 \Exs \vecnorm{(I - \SpecMat)^{-1} \PhiOp
      (\Lmat_{t + 1} - \Lmat) \vbar}{2}^2 + 2 \Exs \vecnorm{(I -
      \SpecMat)^{-1} \PhiOp (\Lmat_{t + 1} - \Lmat) (\vvec_t -
      \vbar)}{2}^2,
\end{multline*}
and
\begin{align*}
      \Exs \statnorm{(I - \Lmat)^{-1} \projecttolin(\bvec_{t + 1} -
        \bvec)}^2 &= \Exs \vecnorm{(I - \SpecMat)^{-1} \PhiOp
        (\bvec_{t + 1} - \bvec)}{2}^2.
\end{align*}
Putting together the pieces, we obtain
\begin{align*}
    \Exs \statnorm{\vhat_\numobs - \vbar}^2 &\leq \frac{3}{\numobs -
      \numburn} \trace \left( (I - \SpecMat)^{-1} \cdot \cov (\PhiOp
    (\bvec_1 - \bvec)) \cdot (I - \SpecMat)^{- \top} \right)\\ &\quad
    \quad + \frac{6}{\numobs - \numburn} \trace \left( (I -
    \SpecMat)^{-1} \cdot \cov (\PhiOp (\Lmat_1 - \Lmat) \vbar) \cdot
    (I - \SpecMat)^{- \top} \right)\\ &\quad\quad\quad +
    \frac{6}{(\numobs - \numburn)^2} \sum_{t = \numburn}^\numobs \Exs
    \vecnorm{(I - \SpecMat)^{-1} \PhiOp (\Lmat_{t + 1} - \Lmat)
      (\vvec_t - \vbar)}{2}^2 + \frac{3\Exs \statnorm{\vvec_\numobs -
        \vvec_{\numburn}}^2 }{\stepsize^2 (\numobs - \numburn)^2 (1 -
      \kappa)^2},
\end{align*}
as claimed.

\noindent It remains to prove the
identity~\eqref{eq:equivalent-form-between-subspace-and-xspace}.

\paragraph{Proof of claim~\eqref{eq:equivalent-form-between-subspace-and-xspace}:}

Note that for any vector $v \in \Xspace$, the vector $z \mydefn (I -
\projecttolin \Lmat)^{-1} \projecttolin v$ is a member of $\lspace$,
since $z = \projecttolin \Lmat z + \projecttolin v$.  Furthermore,
since $\{\phi_j\}_{j = 1}^\usedim$ is a standard basis for $\lspace$,
we have $z = \projecttolin z = \adjoint{\PhiOp} \PhiOp z$, and
consequently,
\begin{align*}
  \PhiOp z = \PhiOp \Lmat z + \PhiOp v = (\PhiOp \Lmat
  \adjoint{\PhiOp}) \PhiOp z + \PhiOp v = \SpecMat \PhiOp z + \PhiOp
  v.
\end{align*}
Since the matrix $\SpecMat$ is invertible, we have $\PhiOp z =
(I_\usedim - \SpecMat)^{-1} \PhiOp v$. Consequently, we have the
identity $z = \adjoint{\PhiOp} \PhiOp z = \adjoint{\PhiOp} (I_\usedim
- \SpecMat)^{-1} \PhiOp v$, which proves the claim.


\subsection{Proof of Corollary~\ref{CorExtreme}}
\label{SecProofCorExtreme}
We begin by applying Theorem~\ref{thm:linear-oracle-ineq} with
$\offpar = 1$. Applying Lemmas~\ref{lem:approx-factor-upper-bounds}
and~\ref{lem:approx-factor-symmetric} yield the desired bounds on the
approximation error in parts (a) and (b). We also claim that
\begin{subequations}
\label{eq:err-cor}
\begin{align} \label{eq:est-error-cor}
    \BigEstErr \leq \frac{(\sigmaA^2 \statnorm{\vbar}^2 + \sigmab^2)
      \usedim}{(1 - \kappa)^2 \numobs},
\end{align}
and that given a sample size such that $\numobs \geq \frac{c \sigmaA^2
  \usedim}{(1 - \kappa)^2} \log^2 \left(\frac{\statnorm{\vvec_0 -
    \vbar}^2 \usedim}{ 1 - \kappa} \right) > \frac{c \sigmaA^2
  \usedim}{(1 - \kappa)^2}$, we have
\begin{align}
\label{eq:hot-error-cor}
    \BigResErr \leq \frac{\sigmaA}{1 - \kappa}
    \sqrt{\frac{\usedim}{\numobs}} \cdot \frac{(\sigmaA^2
      \statnorm{\vbar}^2 + \sigmab^2) \usedim}{(1 - \kappa)^2 \numobs}
    \leq \frac{(\sigmaA^2 \statnorm{\vbar}^2 + \sigmab^2) \usedim}{(1
      - \kappa)^2 \numobs},
\end{align}
\end{subequations}
Combining these two auxiliary claims establishes the corollary. It
remains to establish the bounds~\eqref{eq:err-cor}.

\paragraph{Proof of claim~\eqref{eq:err-cor}:} Let us first handle the
contribution to this error from the noise variables $\bvec_i$. We
begin with the following sequence of bounds:
\begin{align*}
    &\trace \left( (I - \SpecMat)^{-1} \cov (\PhiOp (\bvec_1 - \bvec))
  (I - \SpecMat)^{-\top} \right) \\ &\qquad \qquad \qquad \qquad=
  \trace \left((I - \SpecMat)^{-\top} (I - \SpecMat)^{-1} \cdot \cov
  (\PhiOp (\bvec_1 - \bvec)) \right)\\ &\qquad \qquad \qquad \qquad
  \leq \opnorm{(I - \SpecMat)^{-\top} (I - \SpecMat)^{-1}} \cdot
  \matsnorm{\cov (\PhiOp (\bvec_1 - \bvec))}{nuc} \\ &\qquad \qquad
  \qquad \qquad \leq \opnorm{(I - \SpecMat)^{-1}}^2 \trace \big(\cov
  (\PhiOp (\bvec_1 - \bvec)) \big).
\end{align*}
By the assumption $\kappa (\SpecMat) < 1$, for any vector $u \in
\real^\usedim$, we have that
\begin{align*}
    (1 - \kappa) \vecnorm{u}{2}^2 \leq \inprod{(I - \SpecMat)u}{u}
  \leq \vecnorm{(I - \SpecMat) u}{2} \cdot \vecnorm{u}{2}.
\end{align*}
Consequently, we have the bound $\opnorm{(I - \SpecMat)^{-1}} \leq
\frac{1}{1 - \kappa (\SpecMat)}$.  For the trace of the covariance, we
note by Assumption~\ref{assume-second-moment} that
\begin{align*}
    \trace \big(\cov (\PhiOp (\bvec_1 - \bvec)) \big) = \sum_{j =
      1}^\usedim \statinprod{\phi_j}{\bvec_1 - \bvec}^2 \leq \sigmab^2
    \usedim.
\end{align*}
Putting together the pieces yields $\trace \left( (I - \SpecMat)^{-1}
\cov (\PhiOp (\bvec_1 - \bvec)) (I - \SpecMat)^{-\top} \right) \leq
\frac{\sigmab^2 \usedim}{(1- \kappa)^2}$.

Turning now to the contribution to the error from the random
observation $\Lmat_i$, we have
\begin{align*}
    \trace \left( (I - \SpecMat)^{-1} \cov (\PhiOp (\Lmat_1 - \Lmat)
    \vbar) (I - \SpecMat)^{-\top} \right) \leq \opnorm{(I -
      \SpecMat)^{-1}}^2 \trace \big(\cov (\PhiOp (\Lmat_1 - \Lmat)
    \vbar) \big).
\end{align*}
Once again, Assumption~\ref{assume-second-moment} yields the bound
\begin{align*}
    \trace \big(\cov (\PhiOp (\Lmat_1 - \Lmat) \vbar) \big) = \sum_{j = 1}^\usedim \statinprod{\phi_j}{(\Lmat_1 - \Lmat) \vbar}^2 \leq \sigmaA^2 \statnorm{\vbar}^2 \usedim,
\end{align*}
and combining the pieces proves the claim.

\paragraph{Proof of claim~\eqref{eq:hot-error-cor}:} The proof of this claim is
immediate. Simply note that for $\numobs \geq \frac{c \sigmaA^2
  \usedim}{(1 - \kappa)^2} \log^2 \left(\frac{\statnorm{\vvec_0 -
    \vbar}^2 \usedim}{ 1 - \kappa} \right) > \frac{c \sigmaA^2
  \usedim}{(1 - \kappa)^2}$, we have
\begin{align*}
    \BigResErr \leq \frac{\sigmaA}{1 - \kappa}
    \sqrt{\frac{\usedim}{\numobs}} \cdot \frac{(\sigmaA^2
      \statnorm{\vbar}^2 + \sigmab^2) \usedim}{(1 - \kappa)^2 \numobs}
    \leq \frac{(\sigmaA^2 \statnorm{\vbar}^2 + \sigmab^2) \usedim}{(1
      - \kappa)^2 \numobs}.
\end{align*}


\subsection{Proof of Theorem~\ref{thm:linear-lower-bound}}
\label{subsec:proof-approx-factor-lower-bound}

At a high level, our proof of the lower bound proceeds by constructing
two ensembles of problem instances that are hard to distinguish from
each other, and such that the approximation error on at least one of
them is large. The two instances are indexed by values of a bit
$\plantedbit \in \{-1, 1 \}$, and each instance is, in turn, obtained
as a mixture over $2^{D - d}$ centers; each center is indexed by a
binary string $\rade \in \{-1, 1\}^{D - d}$. The problem is then
phrased as one of estimating the value of $\plantedbit$ from the
observations; this is effectively a reduction to testing and the use
of Le Cam's mixture-vs-mixture method.

Specifically, let $u \in \sphere^{\usedim - 1}$ be an eigenvector
associated to the largest eigenvalue of the matrix $(I -
\SpecMatZero)^{- 1} \big( \conmax^2 I - \SpecMatZero \SpecMatZero^\top
\big) (I - \SpecMatZero)^{- \top}$. By the definition of the
approximation factor $\prefact (\SpecMatZero, \conmax)$, we have:
\begin{align*}
    \big(\prefact (\SpecMatZero, \conmax) - 1 \big) \cdot (I - \SpecMatZero) uu^\top (I - \SpecMatZero)^\top \preceq \conmax^2 I - \SpecMatZero \SpecMatZero^\top.
\end{align*}

Based on the eigenvector $u$, we further define the
$\usedim$-dimensional vectors:
\begin{align}
\label{eq:wenlong-vectors}
w \mydefn \sqrt{\prefact (\SpecMatZero, \conmax) - 1} \cdot (I -
\SpecMatZero) u, \quad \mbox{and}\quad \plantedvec \mydefn
\sqrt{\prefact (\SpecMatZero, \conmax) - 1} \cdot \oracleErr u.
\end{align}
Substituting into the above PSD domination relation yields that
\begin{align}
\label{eq:w-key-relation-in-lb-proof}  
  w w^\top + \SpecMatZero \SpecMatZero^\top \preceq \conmax^2 I.
\end{align}

Now consider the following class of (population-level) problem
instances $(\Lmat^{(\rade, \plantedbit)}, \bvec^{(\rade,
  \plantedbit)}, \vstar_{\rade, \plantedbit})$ indexed by a binary
string $\rade \in \{ -1, 1 \}^{D - d}$ and a bit $\plantedbit \in
\{-1, 1\}$:
\begin{align}
\label{eq:approx-lower-bound-construction-population}  
\Lmat^{(\rade, \plantedbit)} \mydefn \begin{bmatrix} \SpecMatZero &
  \frac{ \sqrt{\usedim}}{\bigdim - \usedim} \rade_{\usedim + 1} w &
  \cdots & \frac{\sqrt{\usedim}}{\bigdim - \usedim} \rade_{\bigdim} w
  \\ 0 & 0 & \cdots & 0\\ \vdots & & \vdots & \\ 0& 0 & \cdots &0
    \end{bmatrix},~&
    \vstar_{\rade, \plantedbit} \mydefn
    \begin{bmatrix}
    \sqrt{2 \usedim} \left( \plantedbit \plantedvec + (I -
    \SpecMatZero)^{-1} \projectedbvecZero \right) \\ \sqrt{2}
    \plantedbit \oracleErr \rade_{\usedim + 1}\\ \vdots \\ \sqrt{2}
    \plantedbit \oracleErr \rade_{\bigdim}
    \end{bmatrix}, \nonumber\\
    \bvec^{(\rade, \plantedbit)} \mydefn (I - \Lmat^{(\rade,
      \plantedbit)}) \vstar_{\rade, \plantedbit} &= \begin{bmatrix}
      \sqrt{2\usedim} \projectedbvecZero\\ \sqrt{2} \plantedbit
      \oracleErr \rade_{\usedim + 1}\\ \vdots\\ \sqrt{2} \plantedbit
      \oracleErr \rade_{\bigdim}
    \end{bmatrix}. 
\end{align}

We take the weight vector $\stationary$ to be
\begin{align*}
    \stationary = \begin{bmatrix} \undermat{d}{\frac{1}{2 d} & \cdots
        & \frac{1}{2d}} & \undermat{(D - d)}{\frac{1}{2 (D - d)} &
        \cdots & \frac{1}{2(D - d)}} \end{bmatrix},
\end{align*}
and the weighted inner product $\langle \cdot, \cdot \rangle$ on the
space $\Xspace = \real^\bigdim$ is defined via
\begin{align*}
    \statinprod{p}{q} \mydefn \sum_{j = 1}^\bigdim p_j \stationary_j
    q_j \quad \text{ for each pair  $p, q \in \real^D$.}
\end{align*}
This choice of inner product then induces the vector norm
$\statnorm{\cdot}$ and operator norm $\hilopnorm{\cdot}$.

Next, we define the basis vectors via
\begin{align*}
\phi_i =
\begin{cases}
\sqrt{2d} e_i \quad &\text{ for } i = 1,2, \cdots, d, \text{ and }
\\ \sqrt{2 (D - d)} e_i &\text{ for } i = d +1 , \cdots D.
\end{cases}
\end{align*}
By construction, we have ensured that $\statnorm{\phi_i} = 1$ for each
$i \in [D]$.  We let the subspace $\lspace$ be the span of the first
$\usedim$ standard basis vectors, i.e., $\lspace \mydefn \myspan (e_1,
e_2, \cdots, e_d)$.

For each binary string $\rade \in \{-1, 1\}^{D - d}$ and signed bit
$\plantedbit \in \{-1, 1\}$, a straightforward calculation reveals
that the projected problem instance satisfies the identities
\begin{subequations}
  \begin{align}
\label{eq:projected-instances-in-lower-bound-construction}    
    \PhiOp \Lmat^{(\rade, \plantedbit)} \adjoint{\PhiOp} =
    \SpecMatZero, \quad \text{ and } \quad \PhiOp b^{(\rade,
      \plantedbit)} = \projectedbvecZero.
\end{align}
Also note that for any pair $(\rade, \plantedbit)$, we have by
construction that
\begin{align}
\label{eq:oracle-error-in-lower-bound-construction}  
    \inf_{\vvec \in \lspace} \statnorm{\vstar_{\rade, \plantedbit} -
      \vvec}^2 = \frac{1}{2 (D - d)} \sum_{j = d + 1}^D (\sqrt{2}
    \plantedbit \oracleErr \rade_j)^2 = \oracleErr^2.
\end{align}
\end{subequations}
In words, this shows that the $\statnorm{\cdot}$-error of
approximating $\vstar_{\rade, \plantedbit}$ with the linear subspace
$\lspace$ is always $\oracleErr$, irrespective of which $\rade \in
\{-1,1\}^{D - d}$ and $\plantedbit \in \{-1, 1\}$ are chosen.

Next, we construct the random observation models for the
$\mathrm{i.i.d.}$ observations, which are also indexed by the pair
$(\epsilon, \plantedbit)$.  In particular, we construct the random
matrix $\Lmat_i^{(\rade, \plantedbit)}$ and random vector
$b_i^{(\rade, \plantedbit)}$ via
\begin{align}
    \Lmat_i^{(\rade, \plantedbit)} \mydefn
    \begin{bmatrix} 
    \SpecMatZero & 0 & \cdots &0&\sqrt{d} \rade_{\tau_\Lmat^{(i)}} w
    &0& \cdots &0\\ 0 & 0 &&& \cdots &&&0\\ &&\vdots && & \vdots &&
    \\ 0 & 0 &&& \cdots &&&0
    \end{bmatrix},\quad
   \bvec_i^{(\rade, \plantedbit)}\mydefn \begin{bmatrix} \sqrt{2d}
     \projectedbvecZero \\ 0\\ \vdots\\ 0\\ \sqrt{2} (D - d)
     \plantedbit \oracleErr \rade_{\tau_b^{(i)}}\\ 0 \\ \vdots\\ 0
    \end{bmatrix} . \label{eq:approx-lower-bound-construction-random}
\end{align}
where the random indices $\tau_\Lmat^{(i)}$ and $\tau_\bvec^{(i)}$ are
chosen independently and uniformly at random from the set $\{ \usedim
+ 1, \usedim + 2, \cdots, \bigdim\}$. By construction, we have ensured
that for each $\rade \in \{-1, 1\}^{\bigdim - \usedim}$ and
$\plantedbit \in \{-1, 1\}$, the observations have mean
\begin{align*}
    \Exs \left[\Lmat_i^{(\rade, \plantedbit)} \right] = \Lmat^{(\rade,
      \plantedbit)}, \quad \mbox{and} \quad \Exs
    \left[\bvec_i^{(\rade, \plantedbit)} \right] = \bvec^{(\rade,
      \plantedbit)}.
\end{align*}

This concludes our description of the problem instances
themselves. Since our proof proceeds via Le Cam's lemma, we require
some more notation for product distributions and mixtures under this
observation model.  Let $\ProbInst_{\rade, \plantedbit}^{(\numobs)}$
denote the $\numobs$-fold product of the probability laws of the pair
$\big(\Lmat_i^{(\rade, \plantedbit)}, \bvec_i^{(\rade,
  \plantedbit)}\big)$. We also define the following mixture of product
measures for each $\plantedbit \in \{-1, 1\}$:
\begin{align*}
    \ProbInst^{(n)}_\plantedbit \mydefn \frac{1}{2^{\bigdim -
        \usedim}} \sum_{\rade \in \{\pm 1\}^{\bigdim - \usedim}}
    \ProbInst_{\rade, \plantedbit}^{(\numobs)}.
\end{align*}
We seek bounds on the total variation distance $\totalvariation \left(
\ProbInst^{(\numobs)}_1 , \ProbInst^{(\numobs)}_{-1} \right)$.

With this setup, the following lemmas assert that (a) Our construction
satisfies the conditions in
Assumption~\ref{assume-second-moment-strong}, and (b) The total
variation distance is small provided $\numobs \ll \sqrt{\bigdim -
  \usedim}$.

\begin{lemma}
  \label{lemma-approx-lower-bound-satisfies-conditions}
For each binary string $\rade \in \{-1, 1\}^{\bigdim - \usedim}$ and
bit $\plantedbit \in \{-1, 1\}$: \\ (a) The population-level matrix
$\Lmat^{(\rade, \plantedbit)}$ defined in
equation~\eqref{eq:approx-lower-bound-construction-population}
satisfies $\hilopnorm{\Lmat^{(\rade, \plantedbit)}} \leq \conmax$.
\\
(b) The random observations $\big( \Lmat^{(\rade, \plantedbit)}_i,
\bvec^{(\rade, \plantedbit)}_i \big)$ defined in
equation~\eqref{eq:approx-lower-bound-construction-random} satisfies
Assumption~\ref{assume-second-moment-strong}, for any scalar pair
$(\sigmaA, \sigmab)$ such that $\sigmaA \geq \conmax$ and $\sigma_b
\geq \oracleErr$.
\end{lemma}

\smallskip

\begin{lemma}
  \label{lem:full-TV}
Under the set-up above, we have $\totalvariation \left(
\ProbInst^{(\numobs)}_1 , \ProbInst^{(\numobs)}_{-1} \right) \leq
\frac{12 n^2}{D - d}$.
\end{lemma}

Part (a) of Lemma~\ref{lemma-approx-lower-bound-satisfies-conditions}
and
equations~\eqref{eq:projected-instances-in-lower-bound-construction}--\eqref{eq:oracle-error-in-lower-bound-construction}
together ensure that population-level problem instance $(\Lmat,
\bvec)$ we constructed belongs to the class $\classApprox
(\SpecMatZero, \projectedbvecZero, \bigdim, \oracleErr,
\conmax)$. Part (b) of
Lemma~\ref{lemma-approx-lower-bound-satisfies-conditions} further
ensures the probability distribution $\ProbInst_{\Lmat, \bvec}$
belongs to the class $\classNoise (\sigmaA,
\sigmab)$. Lemma~\ref{lem:full-TV} ensures that the two mixture
distributions corresponding to different choices of the bit
$\plantedbit$ are close provided $n$ is not too large. The final step
in applying Le Cam's mixture-vs-mixture result is to show that the
approximation error is large for at least one of the choices of the
bit $\plantedbit$. We carry out this step by splitting the rest of the
proof into two cases, depending on whether or not we enforce that our
estimator $\vhat$ is constrained to lie in the subspace
$\lspace$. Throughout, we use the decomposition $\vhat =
\left[\begin{smallmatrix} \vhat_1 \\ \vhat_2\end{smallmatrix}
    \right]$, where $\vhat_1 \in \real^\usedim$ and $\vhat_2 \in
\real^{\bigdim - \usedim}$. Also recall the definition of the vector
$\plantedvec$ from equation~\eqref{eq:wenlong-vectors}.

\paragraph{Case I: $\vhat \in \lspace$.} This corresponds to the ``proper learning'' case
where the estimator is restricted to take values in the subspace
$\lspace$ and $\vhat_2 = 0$. Note that for any $\rade \in \{-1,
1\}^{\bigdim - \usedim}$, we have
\begin{align*}
    \statnorm{\vstar_{\rade, \plantedbit} - \vhat}^2 =
    \statnorm{\vstar_{\rade, \plantedbit} - \projecttolin
      (\vstar_{\rade, \plantedbit})}^2 + \statnorm{\vstar_{\rade,
        \plantedbit} - \vhat}^2 = \oracleErr^2 + \frac{1}{2 \usedim}
    \vecnorm{\vhat_1 - \sqrt{2 \usedim} \plantedbit \plantedvec}{2}^2.
\end{align*}
Therefore, for any $\rade, \rade' \in \{-1, 1\}^{\bigdim - \usedim}$,
the following chain of inequalities holds:
\begin{align*}
    \frac{1}{2} \left( \statnorm{\vstar_{\rade, 1} - \vhat}^2 +
    \statnorm{\vstar_{\rade', -1} - \vhat}^2 \right) &= \oracleErr^2 +
    \frac{1}{4d} \left( \vecnorm{\vhat_1 - \sqrt{2d} \plantedvec}{2}^2
    + \vecnorm{\vhat_1 + \sqrt{2d} \plantedvec}{2}^2 \right)\\ &=
    \oracleErr^2 + \frac{1}{2d} \left( \vecnorm{\vhat_1}{2}^2 + 2d
    \vecnorm{\plantedvec}{2}^2 \right)\\ &\geq \oracleErr^2 +
    \vecnorm{\plantedvec}{2}^2\\ &= \prefact (\SpecMatZero, \conmax)
    \cdot \oracleErr^2.
\end{align*}
By Le Cam's lemma, we thus have
\begin{align*}
     \inf_{\vhat_n \in \Vhatclass_\lspace}~ \sup_{\substack{ (\Lmat,
         \bvec) \in \classApprox \\ \ProbInst_{\Lmat, \bvec} \in
         \classNoise (\sigmaA, \sigmab) }} \Exs \statnorm{\vhat_n -
       \vstar}^2 &\geq \prefact (\SpecMatZero, \conmax) \oracleErr^2
     \cdot \left(1 - \totalvariation (\ProbInst_{-1}^{(n)},
     \ProbInst_1^{(n)}) \right) \\ &\stackrel{(i)}{\geq} (1 - \offpar)
     \cdot \prefact (\SpecMatZero, \conmax) \cdot \oracleErr^2,
\end{align*}
where in step (i), we have applied Lemma~\ref{lem:full-TV} in
conjunction with the inequality $D \geq d + \frac{12n^2}{\offpar}$.

\paragraph{Case II: $\vhat \notin \lspace$.} This corresponds to the case
of ``improper learning'' where the estimator can take values in the
entire space $\Xspace$. In this case, for any pair $\rade, \rade' \in
\{-1, 1\}^{D - d}$, we obtain
\begin{align*}
  \statnorm{\vstar_{\rade, 1} - \vstar_{\rade', -1}} \geq
  \statnorm{\left[ \begin{matrix} 2 \sqrt{2\usedim} \plantedvec^\top &
        0 & \cdots &0 \end{matrix} \right]^\top} = 2
  \vecnorm{\plantedvec}{2} = 2 \oracleErr \sqrt{\prefact
    (\SpecMatZero, \conmax) - 1} .
\end{align*}
Applying triangle inequality and Young's inequality yields the bound
\begin{align*}
    \frac{1}{2} (\statnorm{\vhat - \vstar_{\rade, 1}}^2 +
    \statnorm{\vhat - \vstar_{\rade', 1}}^2) \geq \frac{1}{4}
    (\statnorm{\vhat - \vstar_{\rade, 1}} + \statnorm{\vhat -
      \vstar_{\rade', 1}})^2 \geq \frac{1}{4} \statnorm{\vstar_{\rade,
        1} - \vstar_{\rade', -1}}^2 \geq (\prefact (\SpecMatZero,
    \conmax) - 1) \cdot \oracleErr^2.
\end{align*}
By Le Cam's lemma, we once again have
\begin{align*}
    \inf_{\vhat_n \in \Vhatclass_\Xspace}~ \sup_{\substack{ (\Lmat,
        \bvec) \in \classApprox \\ \ProbInst_{\Lmat, \bvec} \in
        \classNoise (\sigmaA, \sigmab) }} \Exs \statnorm{\vhat_n -
      \vstar}^2 &\geq \big(\prefact (\SpecMatZero, \conmax) - 1 \big)
    \cdot \oracleErr^2 \cdot \left(1 - \totalvariation
    (\ProbInst_{-1}^{(n)}, \ProbInst_1^{(n)}) \right) \\ &\geq (1 -
    \offpar) \cdot \big(\prefact (\SpecMatZero, \conmax) - 1 \big)
    \cdot \oracleErr^2.
\end{align*}

\smallskip

Putting together the two cases completes the proof.


\subsubsection{Proof of Lemma~\ref{lemma-approx-lower-bound-satisfies-conditions}}

We prove the two parts of the lemma separately. Once again, recall our
definition of the pair $(w, \plantedvec)$ from
equation~\eqref{eq:wenlong-vectors}.

\paragraph{Proof of part (a):}
In order to study the operator norm of the matrix $\Lmat^{(\rade,
  \plantedbit)}$ in the Hilbert space $\Xspace$, we consider a vector
$p = \left[ \begin{smallmatrix} p^{(1)}\\p^{(2)} \end{smallmatrix}
  \right] \in \real^\bigdim$, with $ p^{(1)} \in \real^\usedim$ and $
p^{(2)} \in \real^{\bigdim - \usedim}$. Assuming $\statnorm{p} = 1$,
we have
\begin{align*}
    \statnorm{\Lmat^{(\rade, \plantedbit)} p}^2 = \frac{1}{2 \usedim}
    \vecnorm{\SpecMatZero p^{(1)} + w \cdot
      \tfrac{\sqrt{\usedim}}{\bigdim - \usedim}\sum_{j = \usedim +
        1}^{\bigdim} \rade_{j} p^{(2)}_j}{2}^2.
\end{align*}
By the Cauchy--Schwarz inequality, we have
\begin{align*}
  \abss{ \tfrac{\sqrt{\usedim}}{\bigdim - \usedim}\sum_{j = \usedim +
      1}^{\bigdim} \rade_{j} p^{(2)}_j}^2 \leq
  \tfrac{\usedim}{(\bigdim - \usedim)^2} \Big( \sum_{j = \usedim +
    1}^\bigdim \rade_j^2 \Big) \Big( \sum_{j = \usedim + 1}^\bigdim
  \big( p^{(2)}_j \big)^2 \Big) = \tfrac{d}{D - d}
  \vecnorm{p^{(2)}}{2}^2.
\end{align*}
Define the vector $a_1 \mydefn \frac{1}{\sqrt{2d}} p^{(1)} \in
\real^\usedim$ and $a_2 \mydefn \frac{1}{\sqrt{2 (D - d)}}
\vecnorm{p^{(2)}}{2}$. Clearly, we have $1 = \statnorm{p}^2 =
\vecnorm{a_1}{2}^2 + a_2^2$, and so
\begin{align*}
    \statnorm{\Lmat^{(\rade, \plantedbit)} p}^2 &\leq
    \frac{1}{2\usedim} \cdot \sup_{t \in [-1, 1]}
    \vecnorm{\SpecMatZero p^{(1)} + \sqrt{2d} a_2 t w}{2}^2\\ &=
    \frac{1}{2d} \cdot \max \left( \vecnorm{\SpecMatZero p^{(1)} +
      \sqrt{2 \usedim} a_2 w}{2}^2, \vecnorm{\SpecMatZero p^{(1)} -
      \sqrt{2 \usedim} a_2 w}{2}^2 \right)\\ &= \max
    \big(\vecnorm{\SpecMatZero a_1 + a_2 w}{2}^2,
    \vecnorm{\SpecMatZero a_1 - a_2 w}{2}^2 \big)\\ &\leq
    \opnorm{\begin{bmatrix} \SpecMatZero &w
      \end{bmatrix}}^2.
\end{align*}

Equation~\eqref{eq:w-key-relation-in-lb-proof} implies that
$\opnorm{\begin{bmatrix} \SpecMatZero & w
    \end{bmatrix}}^2 = \lambda_{\max} \left( \SpecMatZero \SpecMatZero^\top + w w^\top
    \right) \leq \conmax^2$, and therefore, for all $\rade \in
    \{-1,1\}^{D - d}$ and $\plantedbit \in \{-1,1\}$, we have
\begin{align*}
 \hilopnorm{\Lmat^{(\rade, \plantedbit)}} = \sup_{\statnorm{p} = 1}
 \statnorm{\Lmat^{(\rade, \plantedbit)} p} \leq \conmax,
\end{align*}
as desired.


\paragraph{Proof of part (b):}
Consider any pair of vectors $p, q \in \real^\bigdim$ such that
$\statnorm{p} = \statnorm{q} = 1$. Using the decompositions $p =
\left[ \begin{smallmatrix}p^{(1)}\\p^{(2)} \end{smallmatrix} \right]$
and $q = \left[ \begin{smallmatrix}q^{(1)}\\q^{(2)} \end{smallmatrix}
  \right]$, with $p^{(1)}, q^{(1)} \in \real^\usedim, ~ p^{(2)},
q^{(2)} \in \real^{\bigdim - \usedim}$, we have
\begin{align*}
\Exs \statinprod{p}{(\Lmat_i^{(\rade, \plantedbit)} - \Lmat^{(\rade,
    \plantedbit)}) q}^2 &\leq \frac{1}{(2d)^2} \Exs \left( \sqrt{d}
\rade_{\tau_\Lmat} q^{(2)}_{\tau_\Lmat^{(i)}} w^\top p^{(1)} \right)^2
\\ &= \frac{1}{4 d} \cdot \left( w^\top p^{(1)} \right)^2 \cdot \Exs
\left( q^{(2)}_{\tau_\Lmat^{(i)}} \right)^2\\ &\leq \frac{1}{4d}
\vecnorm{p^{(1)}}{2}^2 \cdot \vecnorm{w}{2}^2 \cdot \frac{1}{D - d}
\vecnorm{q^{(2)}}{2}^2 \\ &\leq \vecnorm{w}{2}^2 \cdot \statnorm{p}^2
\cdot \statnorm{q}^2 \leq \vecnorm{w}{2}^2.
\end{align*}

Recall that $\SpecMatZero \SpecMatZero^\top + ww^\top \leq \conmax^2
I_d$ by equation~\eqref{eq:w-key-relation-in-lb-proof}. Consequently,
we have $\vecnorm{w}{2}^4 \leq \vecnorm{\SpecMatZero^\top w}{2}^2 +
\vecnorm{w}{2}^4 \leq \conmax^2 \vecnorm{w}{2}^2$, which implies that
$\vecnorm{w}{2} \leq \conmax$. Therefore, the noise assumption in
equation~\eqref{eq:assume-second-moment-L-noise-strong} is satisfied
with parameter $\sigmaA = \conmax$.

For the noise on the vector $\bvec$, we note that
\begin{align*}
    \Exs \statinprod{\bvec_i^{(\rade, \plantedbit)} - \bvec}{p}^2 \;
    \leq \; \tfrac{1}{4 (\bigdim - \usedim)^2} \Exs \left( \sqrt{2}
    (\bigdim - \usedim) \plantedbit \oracleErr
    \rade_{\tau_\bvec^{(i)}} p^{(2)}_{\tau_\bvec^{(i)}} \right)^2 \leq
    \tfrac{\oracleErr^2}{2} \Exs \left(p^{(2)}_{\tau_\bvec^{(i)}}
    \right)^2 & \leq \tfrac{\oracleErr^2}{2} \cdot \tfrac{1}{D - d}
    \vecnorm{p^{(2)}}{2}^2\\ & \leq \delta^2 \statnorm{p}^2 =
    \oracleErr^2,
\end{align*}
showing the the the noise
assumption~\eqref{eq:assume-second-moment-b-noise-strong} is satisfied
with $\sigmab = \oracleErr$.


\subsubsection{Proof of Lemma~\ref{lem:full-TV}}

Recall that $\tau_\Lmat^{(i)}, \tau_\bvec^{(i)}$ are the random
indices in the $i$-th sample. We define $\Event$ to be the event that
the indices $\big(\tau_\Lmat^{(i)}\big)_{i = 1}^n,
\big(\tau_\bvec^{(i)}\big)_{i = 1}^n$ are not all distinct, i.e.,
\begin{align*}
    \Event \mydefn \left\{ \exists i_1, i_2 \in [n], ~\mathrm{s.t.}~
    \tau_\Lmat^{(i_1)} = \tau_\bvec^{(i_2)} \right\} \cup \left\{
    \exists i_1 \neq i_2, ~\mathrm{s.t.}~ \tau_\Lmat^{(i_1)} =
    \tau_\Lmat^{(i_2)} \right\} \cup \left\{ \exists i_1 \neq i_2,
    ~\mathrm{s.t.}~ \tau_\bvec^{(i_1)} = \tau_\bvec^{(i_2)} \right\}.
\end{align*}
We claim that
\begin{align}
\label{eq:lecam-proof-equal-in-distr-event}  
    \ProbInst^{(n)}_1 | \Event^C = \ProbInst^{(n)}_{-1}| \Event^C.
\end{align}

Assuming equation~\eqref{eq:lecam-proof-equal-in-distr-event}, we now
give a proof of the upper bound on the total variation distance
$\oracleErr$. We use the following lemma:
\begin{lemma}
\label{lemma:tv-bound-with-event}
Given two probability measures $\ProbInst_1, \ProbInst_2$ and an
event $\Event$ with $\ProbInst_1 (\Event), \ProbInst_2 (\Event) <
1/2$, we have
\begin{align*}
  \totalvariation (\ProbInst_1, \ProbInst_2) \leq
  \totalvariation (\ProbInst_1 | \Event^C, \ProbInst_2 |
  \Event^C) + 3 \ProbInst_1 (\Event) + 3 \ProbInst_2 (\Event).
\end{align*}
\end{lemma}

In order to bound the probability of $\Event$, we apply a union
bound. Under either of the probability measures $\ProbInst^{(n)}_1$
and $\ProbInst^{(n)}_{-1}$, we have the following bound:
\begin{align*}
 \Prob (\Event) \leq \sum_{i, j \in [n]} \Prob \left(\tau_\Lmat^{(i)}
 = \tau_\bvec^{(j)} \right) + \sum_{i_1 < i_2} \Prob
 \left(\tau_\Lmat^{(i_1)} = \tau_\Lmat^{(i_2)} \right) + \sum_{i_1 <
   i_2} \Prob \left(\tau_\bvec^{(i_1)} = \tau_\bvec^{(i_2)} \right)
 \leq \frac{2 n^2}{D - d}.
\end{align*}

Applying Lemma~\ref{lemma:tv-bound-with-event} in conjunction with
equation~\eqref{eq:lecam-proof-equal-in-distr-event} yields
\begin{align}
\label{eq:tv-upper-bound-in-proof-of-approx-factor-lb}  
  \totalvariation (\ProbInst_{1}^{(n)}, \ProbInst_{-1}^{(n)}) \leq
  \frac{12 n^2}{D - d}.
\end{align}
It remains to prove claim~\eqref{eq:lecam-proof-equal-in-distr-event}
and Lemma~\ref{lemma:tv-bound-with-event}.

\paragraph{Proof of equation~\eqref{eq:lecam-proof-equal-in-distr-event}:}
For $\bigdim \geq 2 \numobs + \usedim$, we define a probability
measure $\measureQ$ through the following sampling procedure:
\bcar
    \item Sample a subset $S \subseteq \{\usedim + 1, \cdots,
      \bigdim\}$ of size $2 \numobs$ uniformly at random over all
      possible $\binom{\bigdim - \usedim}{2 \numobs}$ possible
      subsets.
    \item Partition the set $S$ into two disjoint subsets $S = S_\Lmat
      \cup S_\bvec$, each of size $\numobs$. The partition is chosen
      uniformly at random over all $\binom{2 \numobs}{\numobs}$
      possible partitions. Let
    \begin{align*}
    S_\Lmat \mydefn \bigg\{ \widetilde{\tau}_\Lmat^{(1)},
    \widetilde{\tau}_\Lmat^{(2)}, \cdots, \widetilde{\tau}_\Lmat^{(n)}
    \bigg\} \quad \text{ and } S_\bvec \mydefn
    \bigg\{\widetilde{\tau}_\bvec^{(1)}, \widetilde{\tau}_\bvec^{(2)},
    \cdots, \widetilde{\tau}_\bvec^{(n)} \bigg\}.
    \end{align*}
  \item For each $i \in [n]$, sample two random bits
    $\zeta_\Lmat^{(i)}, \zeta_\bvec^{(i)} \simiid \mathcal{U} (\{-1,
    1\})$.
    \item Let $\measureQ$ be the probability distribution of the
      observations $(\Lmat_i, b_i)_{i = 1}^n$, that are constructed
      from the tuple $(\widetilde{\tau}_\Lmat^{(i)},
      \widetilde{\tau}_\bvec^{(i)}, \zeta_\Lmat^{(i)},
      \zeta_\bvec^{(i)})$ defined above. Specifically, we let
    \begin{align*}
            \Lmat_i \mydefn \left[ \begin{matrix} \SpecMatZero & 0 &
                \cdots & 0&\sqrt{d} \zeta_\Lmat^{(i)} w &0& \cdots
                &0\\ 0 & 0 & \cdots & 0& 0 &0& \cdots &0\\ 0 & 0 &&&
                \cdots &&&0\\ &&\vdots && & \vdots && \\ 0 & 0 &&&
                \cdots &&&0
    \end{matrix} \right],\quad
   \bvec_i \mydefn \left[ \begin{matrix} \sqrt{2d} \projectedbvecZero
       \\ 0\\ \vdots\\ 0\\ \sqrt{2} (\bigdim - \usedim)
       \zeta_\bvec^{(i)} \oracleErr\\ 0\\ \ldots\\ 0
    \end{matrix} \right],
    \end{align*}
    where the vector $\sqrt{\usedim} \zeta_\Lmat^{(i)} w$ appears at
    the $\widetilde{\tau}_\Lmat^{(i)}$-th column of the matrix
    $\Lmat_i$, and the scalar $\sqrt{2} (\bigdim - \usedim)
    \zeta_\bvec^{(i)} \delta$ appears at the
    $\widetilde{\tau}_\bvec^{(i)}$-th row of the vector $\bvec_i$.
    \ecar
For either choice of the bit $\plantedbit \in \{ \pm 1\}$, we claim
that the probability measure $\ProbInst_z^{(n)} | \Event^C$ is
identical to the distribution $\measureQ$. To prove this claim, we
first note that conditioned on the event $\Event^C$, the indices
$(\tau_\Lmat^{(i)}, \tau_\bvec^{(i)})_{i = 1}^n$ actually form a
uniform random subset of $\{d + 1, \cdots, D\}$ with cardinality $2n$,
and the partition into $(\tau_\Lmat^{(i)})_{i = 1}^n$ and
$(\tau_\bvec^{(i)})_{i = 1}^n$ is a uniform random partition, i.e.,
\begin{align}
\label{eq:indices-equal-in-distribution}  
\left(\tau_\Lmat^{(i)}, \tau_\bvec^{(i)} \right)_{i = 1}^n \big|
\Event^C \overset{d}{=} \left(\widetilde{\tau}_\Lmat^{(i)},
\widetilde{\tau}_\bvec^{(i)} \right)_{i = 1}^n, \quad \mbox{under
  both $\ProbInst_1^{(n)}$ and
  $\ProbInst_{-1}^{(n)}$.} 
\end{align}

Given an index subset $\big( t_\Lmat^{(i)}, t_\bvec^{(i)} \big)_{i =
  1}^n \subseteq \{d + 1, \cdots, D\}$ that are mutually distinct,
conditioned on the value of $\big( \tau_\Lmat^{(i)}, \tau_\bvec^{(i)}
\big)_{i = 1}^n = \big( t_\Lmat^{(i)}, t_\bvec^{(i)} \big)_{i = 1}^n$,
the observed random bits under the probability distribution
$\ProbInst_z^{(n)}$ are given by
\begin{align*}
  \rade_{\tau_\Lmat^{(1)}}, \rade_{\tau_\Lmat^{(2)}}, \cdots,
  \rade_{\tau_\Lmat^{(n)}}, \plantedbit \rade_{\tau_\bvec^{(1)}},
  \cdots, \plantedbit \rade_{\tau_\bvec^{(n)}},
\end{align*}
which are $2n$ independent Rademacher random variables.

On the other hand, the random bits $\zeta_{\Lmat}^{(1)},
\zeta_{\Lmat}^{(2)}, \cdots, \zeta_{\Lmat}^{(n)}, \zeta_{\bvec}^{(1)},
\zeta_{\bvec}^{(2)}, \cdots, \zeta_{\bvec}^{(n)}$ are also $2n$
independent Rademacher random variables.  Consequently, for any index
subset $\big( t_\Lmat^{(i)}, t_\bvec^{(i)} \big)_{i = 1}^n \subseteq
\{d + 1, \cdots, D\}$ that are mutually distinct, we have the
following equality-in-distribution:
\begin{align}
\label{eq:bits-equal-in-distribution}  
  \left( \rade_{\tau_\Lmat^{(i)}}, \plantedbit
  \rade_{\tau_\bvec^{(i)}} \right)_{i = 1}^n \Big| (\tau_\Lmat^{(i)} =
  t_\Lmat^{(i)},\tau_\bvec^{(i)} = t_\bvec^{(i)})_{i = 1}^n
  \overset{d}{=} \left( \zeta_\Lmat^{(i)}, \zeta_\bvec^{(i)}
  \right)_{i = 1}^n \Big| (\widetilde{\tau}_\Lmat^{(i)} =
  t_\Lmat^{(i)}, \widetilde{\tau}_\bvec^{(i)} = t_\bvec^{(i)})_{i =
    1}^n.
\end{align}
Putting equations~\eqref{eq:indices-equal-in-distribution} and
\eqref{eq:bits-equal-in-distribution} together completes the proof.
\qed

\paragraph{Proof of Lemma~\ref{lemma:tv-bound-with-event}:}

Given a function $f$ with range contained in $[0,1]$, we have
\begin{align*}
    &\abss{\int f (x) \ProbInst_1 (dx) - \int f(x) \ProbInst_2
    (dx)}\\ &\leq \abss{\int_\Event f (x) \ProbInst_1 (dx)} +
  \abss{\int_\Event f (x) \ProbInst_2 (dx)} + \abss{\int_{\Event^C} f
    (x) \ProbInst_1 (dx) - \int_{\Event^C} f (x) \ProbInst_2 (dx)}\\ &
  \leq \ProbInst_1 (\Event) + \ProbInst_2 (\Event) + \ProbInst_1
  (\Event^C) \cdot \abss{\frac{\int_{\Event^C} f (x) \ProbInst_1
      (dx)}{\ProbInst_1 (\Event^C)} - \frac{\int_{\Event^C} f (x)
      \ProbInst_2 (dx)}{\ProbInst_2 (\Event^C)}} + \frac{|\ProbInst_1
    (\Event^C) - \ProbInst_2 (\Event^C)|}{\ProbInst_2 (\Event^C)}\\ &
  \leq \ProbInst_1 (\Event) + \ProbInst_2 (\Event) + \totalvariation
  (\ProbInst_1 | \Event^C, \ProbInst_2 | \Event^C) + 2 |\ProbInst_1
  (\Event) - \ProbInst_2 (\Event)|\\ &\leq 3 (\ProbInst_1 (\Event) +
  \ProbInst_2 (\Event)) + \totalvariation (\ProbInst_1 | \Event^C,
  \ProbInst_2 | \Event^C),
\end{align*}
which completes the proof.  \qed

\subsection{Proof of Theorem~\ref{thm:linear-stat-error-lower-bound}}\label{subsec:proof-est-error-lower-bound}

In order to prove our local minimax lower bound, we make use of the
Bayesian Cram\'{e}r--Rao bound, also known as the van Trees
inequality.  In particular, we use a functional version of this
inequality.  It applies to a parametric family of densities $\{p_\eta,
\eta \in \Theta \}$ w.r.t Lebesgue measure, with sufficient regularity
so that the Fisher information matrix $I (\eta) \mydefn \Exs_\eta
\big[ ( \frac{\partial}{\partial \eta} \log p_\eta (X_1) ) (
  \frac{\partial}{\partial \eta} \log p_\eta (X_1) )^\top \big]$ is
well-defined.
\begin{proposition}[Theorem 1 of~\cite{gill1995applications}, special case]
  \label{prop:van-tree-functional}
  Given a prior distribution $\rho$ with bounded support contained
  within $\Theta$, let $T: \support (\rho) \mapsto \real^p$ denote a
  locally smooth functional. Then for any estimator $\widehat{T}$
  based on i.i.d. samples $X_1^n = \{X_i\}_{i=1}^\numobs$ and for any
  smooth matrix-valued function $C: \real^d \rightarrow \real^{p
    \times d}$, we have
\begin{align*}
     \underset{\eta \sim \rho}{\Exs} \underset{X_1^n \sim
       p_\eta}{\Exs} \vecnorm{\widehat{T} (X_1^n) - T (\eta)}{2}^2
     \geq \tfrac{\left( \int \trace \left(C (\eta) \frac{\partial
         T}{\partial \eta} (\eta) \right) \rho (\eta) \usedim \eta
       \right)^2}{n \int \trace \left(C (\eta) I (\eta) C (\eta)^\top
       \right) \rho (\eta) d \eta + \int \vecnorm{\nabla \cdot C
         (\eta) + C (\eta) \cdot \nabla \log \rho (\eta)}{2}^2 \rho
       (\eta) d \eta}.
\end{align*}
\end{proposition}

Recall that our lower bound is local, and holds for problem instances
$(\Lmat, \bvec)$ such that the pair $(\PhiOp \Lmat \adjoint{\PhiOp},
\PhiOp b)$ is within a small $\ell_2$ neighborhood of a fixed pair
$(\SpecMatZero, \projectedbvecZero)$.  We proceed by constructing a
careful prior on such instances that will allow us to apply
Proposition~\ref{prop:van-tree-functional}; note that it suffices to
construct our prior over the $d \times d$ matrix $\PhiOp \Lmat
\adjoint{\PhiOp}$ and $d$-dimensional vector $\PhiOp \bvec$.  For the
rest of the proof, we work under the orthonormal basis $\{\phi_j
\}_{j= 1}^\usedim$. We also use the convenient shorthand $\vbarZero
\defn \PhiOp \vbar_0$.

As a building block for our construction, we consider the
one-dimensional density function \mbox{$\mu(t) \mydefn \cos^2 \left(
  \frac{\pi t}{2} \right) \cdot \bm{1}_{t \in [-1, 1]}$,} borrowed
from Section 2.7 of Tsybakov~\cite{tsybakov2008introduction}. It can
be verified that $\mu$ defines a probability measure supported on the
interval $[-1, 1]$. We denote by $\mu^{\otimes d}$ the $d$-fold
product measure of $\mu$. Let $Z$ and $Z'$ denote two random vectors
drawn i.i.d. from the distribution $\mu^{\otimes d}$.

We use an auxiliary pair of $\real^d$-valued random variables $(\psi,
\lambda)$ given by
\begin{align}
  \psi \mydefn \frac{1}{\sqrt{\numobs}} \Sigma_\bvec^{\frac{1}{2}} Z
  \quad \text{ and } \quad \lambda \mydefn \frac{1}{\sqrt{\numobs}}
  \Sigma_\Lmat^{\frac{1}{2}} Z'. \label{eq:prior-bayes-risk-lb}
\end{align}
Our choice of this pair is motivated by the fact that the Fisher
information matrix of this distribution takes a desirable form. In
particular, we have the following lemma.
\begin{lemma}
\label{lemma:fisher-info-in-bayes-risk-lb}
Let $\rho : \real^{2d} \rightarrow \real_+$ denote the density of
$(\psi, \lambda)$ defined in
equation~\eqref{eq:prior-bayes-risk-lb}. Then
\begin{align*}
 I(\rho) = n \pi \left[\begin{matrix} \Sigma_\bvec^{-1} & 0 \\ 0 &
     \Sigma_\Lmat^{-1}
        \end{matrix}
        \right].
    \end{align*}
\end{lemma}

We now use the pair $(\psi, \lambda)$ in order to define the ensemble
of population-level problem instances
\begin{align}
  \label{eq:M-beta-def}
  \SpecMatPsilam \mydefn \SpecMatZero + \vecnorm{\vbarZero}{2}^{-2}
  \lambda (\vbarZero)^\top \quad \text{ and } \quad
  \projectedbvecPsilam \mydefn \projectedbvecZero + \psi.
\end{align}
In order to define the problem instance in the Hilbert space
$\Xspace$, we simply let $\LmatPsilam \mydefn \adjoint{\Phi}_d
\SpecMatPsilam \PhiOp$ and $\bvecPsilam \mydefn \adjoint{\PhiOp}
\projectedbvecPsilam$, for a given basis $(\phi_i)_{i = 1}^d$ in the
space $\lspace$.

The matrix-vector pair $(\LmatPsilam, \bvecPsilam)$ induces the fixed
point equation $\vbarpsilam = \SpecMatPsilam \vbarpsilam +
\projectedbvecPsilam$, and its solution is given by
\begin{align*}
    \vbarpsilam = (I - \SpecMatPsilam)^{-1} \projectedbvecPsilam = (I
    - \SpecMatZero - \vecnorm{\vbarZero}{2}^{-2} \lambda
    (\vbarZero)^\top)^{-1} (\projectedbvec_0 + \psi).
\end{align*}
Note that by construction, the Jacobian matrix formed by taking the
partial derivative of $\vbarpsilam$ with respect to $\psi$ and
$\lambda$ is given by
\begin{align*}
    \nabla_{\psi, \lambda} \vbarpsilam=
     \begin{bmatrix} (I - \SpecMatPsilam)^{-1} &
        \vecnorm{\vbarZero}{2}^{-2} (\vbarZero)^\top (I -
        \SpecMatPsilam)^{-1} (\projectedbvec_0 + \psi) \cdot (I -
        \SpecMatPsilam)^{-1} \end{bmatrix}.
\end{align*}

Now define the observation model via $\Lmat_i^{(\psi, \lambda)}
\mydefn \adjoint{\PhiOp} \SpecMat_i^{(\psi, \lambda)} \PhiOp$ and
$b_i^{(\psi, \lambda)} \mydefn \adjoint{\PhiOp}
\projectedbvec_i^{(\psi, \lambda)}$, where
\begin{align}
    \SpecMat_i^{(\psi, \lambda)} \mydefn \SpecMatPsilam +
    \vecnorm{\vbarZero}{2}^{-2} w_i (\vbarZero)^\top \quad \text{ and
    } \quad \projectedbvec_i^{(\psi, \lambda)} \mydefn
    \projectedbvecPsilam + w_i',\label{eq:bayes-risk-lb-obs}
\end{align}
where $w_i \sim \Normal(0, \Sigma_\Lmat)$ and $w_i' \sim \Normal(0,
\Sigma_\bvec)$ are independent.

The following lemma certifies some basic properties of observation
model constructed above.
\begin{lemma}
\label{lemma:constructed-instance-in-class-bayes-cramer-rao}
Consider the ensemble of problem instances defined in
equations~\eqref{eq:M-beta-def} and~\eqref{eq:bayes-risk-lb-obs}. For
each pair $(\psi, \lambda)$ in the support of $\rho$, each index $j
\in [\usedim]$ and each unit vector $u \in \sphere^{\usedim - 1}$, we
have
\begin{subequations}
  \begin{align}
    \matsnorm{\SpecMatPsilam - \SpecMatZero}{F} \leq \sigmaA
    \sqrt{\frac{\usedim}{\numobs}}, \quad &\mbox{and} \quad
    \vecnorm{\projectedbvecPsilam - \projectedbvec_0}{2} \leq \sigmab
    \sqrt{\frac{\usedim}{\numobs}},\label{eq:bayes-risk-lb-construction-locality}\\ \cov
    \left( \big(\SpecMat_1^{(\psi, \lambda)} - \SpecMatPsilam \big)
    \vbarZero \right) = \Sigma_\Lmat, \quad &\mbox{and} \quad \cov
    \left( \projectedbvec_1^{(\psi, \lambda)} - \projectedbvecPsilam
    \right) =
    \Sigma_\bvec,\label{eq:bayes-risk-lb-construction-cov}\\ \Exs
    \left( e_j^\top \big(\SpecMat_1^{(\psi, \lambda)} -
    \SpecMatPsilam\big) u \right)^2 \leq \sigmaA^2, \quad
    &\mbox{and}\quad \Exs \left( e_j^\top \projectedbvec_1^{(\psi,
      \lambda)} - \projectedbvecPsilam \right)^2 \leq
    \sigmab^2.\label{eq:bayes-risk-lb-construction-noise}
      \end{align}
    \end{subequations}
\end{lemma}
Lemma~\ref{lemma:constructed-instance-in-class-bayes-cramer-rao}
ensures that our problem instance lies in the desired class In
particular, equation~\eqref{eq:bayes-risk-lb-construction-locality}
guarantees that the population-level problem instance
$\big(\LmatPsilam, \bvecPsilam \big)$ lies in the class $\classEst$;
on the other hand, equation~\eqref{eq:bayes-risk-lb-construction-cov}
and~\eqref{eq:bayes-risk-lb-construction-noise} guarantees that the
probability distribution $\ProbInst_{\Lmat, \bvec}$ we constructed
lies in the class $\classCov$.

Some calculations yield that the Fisher information matrix for this
observation model is given by
\begin{align*}
I(\psi, \lambda) = \begin{bmatrix} \Sigma_\bvec^{-1} & 0 \\
    0 & \Sigma_\Lmat^{-1} \end{bmatrix},
\end{align*}
for any $\psi, \lambda \in \real^d$.

We will apply Proposition~\ref{prop:van-tree-functional} shortly, for
which we use the following matrix $C$:
\begin{align*}
    C (\psi, \lambda) \mydefn \nabla_{\psi, \lambda} \vbarpsilam
    \big|_{(0, 0)} \cdot I (\psi, \lambda)^{-1} = \begin{bmatrix} (I -
      \SpecMatZero)^{-1} \Sigma_\bvec &(I - \SpecMatZero)^{-1}
      \Sigma_\Lmat \end{bmatrix}.
\end{align*}
Note that by construction, the matrix $C$ does not depend on the pair
$(\psi, \lambda)$.

We also claim that if $n \geq 16 \sigmaA^2 \opnorm{(I -
  \SpecMatZero)^{-1}}^2 d$, then the following inequalities hold for
our construction:
\begin{subequations}
\label{eq:algebra-claims}
\begin{align}
T_\bvec \mydefn \Exs_\rho \left[ \trace \left( (I - \SpecMatZero)^{-1}
  \Sigma_\bvec (I - \SpecMatPsilam)^{-\top} \right) \right] &\geq
\frac{1}{2} \trace \left( (I - \SpecMatZero)^{-1} \Sigma_\bvec (I -
\SpecMatZero)^{- \top} \right) \text{ and } \label{eq:algebra-claim1}
\\ T_\Lmat \mydefn \Exs_\rho \left[ \frac{(\vbarZero)^\top
    \vbarpsilam}{\vecnorm{\vbarZero}{2}^2 } \trace \left( (I -
  \SpecMatZero)^{-1} \Sigma_\Lmat (I - \SpecMatPsilam)^{-\top} \right)
  \right] &\geq \frac{1}{3} \trace \left( (I - \SpecMatZero)^{-1}
\Sigma_\Lmat (I - \SpecMatZero)^{-
  \top}\right). \label{eq:algebra-claim2}
\end{align}
\end{subequations}
Taking these two claims as given for the moment, let us complete the
proof of the theorem. First, note that
\begin{align}
  &\Exs_\rho \left[\trace \left( C (\psi, \lambda) \cdot \nabla_{\psi,
      \lambda}\vbarpsilam^\top \right) \right] \notag \\ &\quad =
  \Exs_\rho \left[ \trace \left( (I - \SpecMatZero)^{-1} \Sigma_\bvec
    (I - \SpecMatPsilam)^{-\top} \right) \right] + \Exs_\rho \left[
    \frac{(\vbarZero)^\top \vbarpsilam}{\vecnorm{\vbarZero}{2}^2}
    \trace \left( (I - \SpecMatZero)^{-1} \Sigma_\Lmat (I -
    \SpecMatPsilam)^{-\top} \right) \right] \notag \\ &\quad \geq
  \frac{5}{6} \trace \left( (I - \SpecMatZero)^{-1} \Sigma_\bvec (I -
  \SpecMatZero)^{- \top}
  \right).  \label{eq:bayes-risk-lower-bound-term12}
\end{align}
Second, since $I (\cdot, \cdot)$ and $C (\cdot, \cdot)$ are both
constant functionals, we have that
\begin{align}
  \Exs \left[ \trace \left( C(\psi, \lambda) I (\psi, \lambda) C(\psi,
    \lambda) \right) \right] = \trace \left( (I - \SpecMatZero)^{-1}
  (\Sigma_\Lmat + \Sigma_\bvec)(I - \SpecMatZero)^{-\top}
  \right). \label{eq:bayes-risk-lower-bound-denom-1}
\end{align}
Additionally, Lemma~\ref{lemma:fisher-info-in-bayes-risk-lb} yields
\begin{align}
  \Exs \vecnorm{\nabla \cdot C (\psi, \lambda) + C (\psi, \lambda)
    \cdot \nabla \log \rho (\psi, \lambda)}{2}^2 &= \trace \left( C
  (0, 0) \cdot\Exs \left[\nabla \log \rho (\psi, \lambda)\nabla \log
    \rho (\psi, \lambda)^\top \right] C(0, 0)^\top \right)
  \nonumber\\ &= n \pi \cdot \trace \left( (I - \SpecMatZero)^{-1}
  (\Sigma_\Lmat + \Sigma_\bvec)(I - \SpecMatZero)^{-\top}
  \right). \label{eq:bayes-risk-lower-bound-denom-2}
\end{align}

We are finally in a position to put together the pieces.  Applying
Proposition~\ref{prop:van-tree-functional} and combining
equations~\eqref{eq:bayes-risk-lower-bound-term12},~\eqref{eq:bayes-risk-lower-bound-denom-1},
and~\eqref{eq:bayes-risk-lower-bound-denom-2}, we obtain the lower
bound
\begin{align}
    \int \Exs \vecnorm{\widehat{x}_n (\Lmat_1^n, b_1^n) -
      \vbarpsilam}{2}^2 \rho (d\psi, d\lambda) \geq \frac{\trace
      \left( (I - \SpecMatZero)^{-1} (\Sigma_\Lmat + \Sigma_\bvec) (I
      - \SpecMatZero)^{- \top}\right)}{9 (1 + \pi)
      n}, \label{eq:bayes-risk-lower-bound}
\end{align}
for any estimator $\widehat{x}_n$ that takes values in $\real^\usedim$.

For the problem instances we construct, note that
\begin{align*}
    \vbar^{(\psi, \lambda)} = (I - \projecttolin\Lmat^{(\psi,
      \lambda)})^{-1} \projecttolin \bvec^{(\psi, \lambda)} =
    \adjoint{\PhiOp} (I - \SpecMat^{(\psi, \lambda)})^{-1}
    \projectedbvec^{(\psi, \lambda)} = \adjoint{\PhiOp} \vbarpsilam.
\end{align*}
For any estimator $\vhat_n \in \Vhatclass_\Xspace$, we note that
\begin{align*}
    \statnorm{\vbar^{(\psi, \lambda)} - \vhat_n}^2 \geq
    \statnorm{\projecttolin (\vbar^{(\psi, \lambda)} - \vhat_n)}^2 =
    \vecnorm{\PhiOp \vhat_n - \vbarpsilam}{2}^2.
\end{align*}

Recall by
Lemma~\ref{lemma:constructed-instance-in-class-bayes-cramer-rao} that
on the support of the prior distribution $\rho$, the population-level
problem instance $\big(\LmatPsilam, \bvecPsilam \big)$ lies in the
class $\classEst$, and that the probability distribution
$\ProbInst_{\Lmat, \bvec}$ we constructed lies in the class
$\classCov$.  We thus have the minimax lower bound
\begin{multline*}
    \inf_{\vhat_n \in \Vhatclass_n} \sup_{ \substack{ (\Lmat, \bvec)
        \in \classEst \\ \ProbInst_{\Lmat, \bvec}\in \classCov }} \Exs
    \statnorm{\vhat_n (\Lmat_1^n, b_1^n) - \vbar}^2 \geq
    \inf_{\widehat{x}_n} \sup_{(\psi, \lambda) \in \support (\rho)}
    \Exs \vecnorm{\widehat{x}_n (\Lmat_1^n, b_1^n) -
      \vbarpsilam}{2}^2\\ \geq \int \Exs \vecnorm{\widehat{x}_n
      (\Lmat_1^n, b_1^n) - \vbarpsilam}{2}^2 \rho (d\psi, d\lambda)
    \geq c \cdot \frac{\trace \left( (I - \SpecMatZero)^{-1} \SigStar
      (I - \SpecMatZero)^{- \top}\right)}{\numobs}
\end{multline*}
for $c = \frac{1}{9 (1 + \pi)} > 0$, which completes the proof of the
theorem.


\subsubsection{Proof of Lemma~\ref{lemma:fisher-info-in-bayes-risk-lb}}

We first note that $\lambda$ is independent of $\psi$, and
consequently $\rho = \rho_\bvec \otimes \rho_a$, where $\rho_\bvec,
\rho_\Lmat$ are the marginal densities for $\psi$ and $\lambda$
respectively. Since the Fisher information tensorizes over product
measures, it suffices to compute the Fisher information of
$\rho_\Lmat$ and $\rho_\bvec$ separately.

By a change of variables, we have
\begin{align*}
  \rho_\Lmat (\lambda) = n^{\frac{d}{2}} \det (\Sigma_\Lmat)^{- \frac{1}{2}}
  \cdot \mu^{\otimes d} \left( \sqrt{\numobs} \Sigma_\Lmat^{- 1/2} \lambda
  \right).
\end{align*}
Substituting this into the expression for Fisher information, we obtain
\begin{align*}
    I (\rho_a) &= \int (\nabla \log \rho_\Lmat (\lambda)) (\nabla \log
    \rho_a (\lambda))^\top \rho_\Lmat (\lambda) d\lambda\\ &= \int
    \left( \sqrt{\numobs} \Sigma_\Lmat^{-1/2} \nabla \log \mu^{\otimes
      d} (\plantedvec) \right) \cdot \left( \sqrt{\numobs}
    \Sigma_\Lmat^{-1/2} \nabla \log \mu^{\otimes d} (z) \right)^\top
    \mu^{\otimes d} (z) dz\\ &= n \Sigma_\Lmat^{-1/2} \cdot
    \underbrace{\Exs_{Z\sim \mu^{\otimes d}} \left[ (\nabla \log
        \mu^{\otimes d} (Z)) (\nabla \log \mu^{\otimes d} (Z))^\top
        \right]}_{I \left(\mu^{\otimes d} \right)} \cdot
    \Sigma_\Lmat^{-1/2}.
\end{align*}

Finally, since $\mu^{\otimes d}$ is a product measure, we have $I
\left(\mu^{\otimes d} \right) = I (\mu) \cdot I_d = \pi I_d$, and
hence $I (\rho_\Lmat) = \pi n \Sigma_\Lmat^{-1}$.  Reasoning similarly
for $\rho_\bvec$, we have that $I (\rho_\bvec) = \pi n
\Sigma_\bvec^{-1}$. This completes the proof.


\subsubsection{Proof of Lemma~\ref{lemma:constructed-instance-in-class-bayes-cramer-rao}}

We prove the three facts in sequence.

\paragraph{Proof of equation~\eqref{eq:bayes-risk-lb-construction-locality}:}

Note that the scalars $\sigmaA$ and $\sigmab$ satisfies the
compatibility condition~\eqref{eq:compatibility-condition-cov}, we
therefore have the bounds
\begin{align*}
  \matsnorm{\SpecMatPsilam - \SpecMatZero}{F} &=
  \vecnorm{\vbarZero}{2}^{-1} \cdot \vecnorm{\lambda}{2} \leq n^{-1/2}
  \vecnorm{\vbarZero}{2}^{-1} \cdot \sqrt{\trace (\Sigma_\Lmat)} \leq
  \sigmaA
  \sqrt{\frac{\usedim}{\numobs}},\\ \vecnorm{\projectedbvecPsilam -
    \projectedbvec_0}{2} &= \vecnorm{\psi}{2} \leq \sqrt{n^{-1} \trace
    (\Sigma_\bvec)} \leq \sigmab \sqrt{\frac{\usedim}{\numobs}},
\end{align*}
which completes the proof of the first bound.

\paragraph{Proof of equation~\eqref{eq:bayes-risk-lb-construction-cov}:}

Straightforward calculation leads to the following identities
\begin{align*}
    \cov \left( \big(\SpecMat_1^{(\psi, \lambda)} - \SpecMatPsilam
    \big) \vbarZero \right) &= \cov \left( \vecnorm{\vbarZero}{2}^{-2}
    w_1 \vbarZero^\top \vbarZero \right) = \cov (w_1) =
    \Sigma_\Lmat,\\ \cov \left( \projectedbvec_1^{(\psi, \lambda)} -
    \projectedbvecPsilam \right) & = \cov (w_1') = \Sigma_\bvec.
\end{align*}

\paragraph{Proof of equation~\eqref{eq:bayes-risk-lb-construction-noise}:}
Given any index $j \in [\usedim]$ and vector $u \in \sphere^{\usedim -
  1}$, we note that:
\begin{align*}
  \Exs \left( e_j^\top \big(\SpecMat_1^{(\psi, \lambda)} -
  \SpecMatPsilam\big) u \right)^2 = \frac{1}{\vecnorm{\vbarZero}{2}^4}
  \Exs \left(e_j^\top w_1 \cdot \vbarZero^\top u\right)^2 \leq
  \frac{1}{\vecnorm{\vbarZero}{2}^2} \Exs \left( e_j^\top w_1
  \right)^2 = \frac{1}{\vecnorm{\vbarZero}{2}^2} e_j^\top \Sigma_\Lmat
  e_j \leq \sigmaA^2,
\end{align*}
where the last inequality is due to the compatibility
condition~\eqref{eq:compatibility-condition-cov}.

Similarly, for the noise $ \projectedbvec_1^{(\psi, \lambda)}  - \projectedbvecPsilam$, we have:
\begin{align*}
    \Exs \left(e_j^\top \big(\projectedbvec_1^{(\psi, \lambda)} -
    \projectedbvecPsilam\big)\right)^2 = \Exs \left( e_j^\top w_1'
    \right)^2 = e_j^\top \Sigma_\bvec e_j \leq \sigmab^2,
\end{align*}
which completes the proof of the last condition.

\subsubsection{Proof of claims~\eqref{eq:algebra-claims}}

We prove the two bounds separately, using the convenient shorthand
$I_\bvec$ for the LHS of claim~\eqref{eq:algebra-claim1} and $I_a$ for the
LHS of claim~\eqref{eq:algebra-claim2}.

\paragraph{Proof of claim~\eqref{eq:algebra-claim1}:}

We begin by applying the matrix inversion formula, which yields
\begin{align*}
    (I - \SpecMatPsilam)^{-1} = (I - \SpecMatZero)^{-1} -
  \underbrace{\frac{1}{\vecnorm{\vbarZero}{2}^2 - (\vbarZero)^\top (I -
      \SpecMatZero) \lambda}(I - \SpecMatZero)^{-1} \lambda (\vbarZero)^\top (I -
    \SpecMatZero)^{-1}}_{=: H}.
\end{align*}
For $n \geq 16 \sigmaA^2 d$, we have
\begin{align*}
    \abss{(\vbarZero)^\top (I - \SpecMatZero) \lambda} \leq 2
    \vecnorm{\vbarZero}{2} \cdot \vecnorm{\lambda}{2} \leq 2
    \vecnorm{\vbarZero}{2} \sqrt{n^{-1} \trace (\Sigma_\Lmat)} \leq 2
    \sigmaA \vecnorm{\vbarZero}{2}^2 \sqrt{\frac{\usedim}{\numobs}} \leq \frac{1}{2}
    \vecnorm{\vbarZero}{2}^2,
\end{align*}

To bound $T_\bvec$ from below, we note that
\begin{align*}
    T_\bvec &= \trace \left( (I - \SpecMatZero)^{-1} \Sigma_\bvec (I - \SpecMatZero)^{-
      \top} \right) - \trace \left( (I - \SpecMatZero)^{-1} \Sigma_\bvec \Exs
    [H^\top] \right)\\ &\geq \trace \left( (I - \SpecMatZero)^{-1}
    \Sigma_\bvec (I - \SpecMatZero)^{- \top} \right) - \matsnorm{ (I - \SpecMatZero)^{-1}
      \Sigma_\bvec (I - \SpecMatZero)^{- \top}}{nuc} \cdot \opnorm{(I - \SpecMatZero)^\top
      \Exs[H]^\top}\\ &= \trace \left( (I - \SpecMatZero)^{-1} \Sigma_\bvec (I
    - \SpecMatZero)^{- \top} \right) \cdot \left(1 - \opnorm{\Exs[H] (I - \SpecMatZero)}
    \right).
\end{align*}

When $n \geq 4 \sigmaA^2 d$, we have
\begin{align*}
    \opnorm{\Exs [H] (I - \SpecMatZero)} &\leq \sum_{k = 0}^{\infty}
    \frac{1}{\vecnorm{\vbarZero}{2}^2} \opnorm{\Exs \left[ \left(
        \frac{(\vbarZero)^\top (I - \SpecMatZero)
          \lambda}{\vecnorm{\vbarZero}{2}^2} \right)^k (I - \SpecMatZero)^{-1}
        \lambda (\vbarZero)^\top\right]}\\ &\leq
    \frac{1}{\vecnorm{\vbarZero}{2}^2} \sum_{k = 1}^{\infty} \Exs
    \left( \abss{ \frac{(\vbarZero)^\top (I - \SpecMatZero)
        \lambda}{\vecnorm{\vbarZero}{2}^2} }^k \cdot \matsnorm{ (I -
      \SpecMatZero)^{-1} \lambda (\vbarZero)^\top}{F} \right)\\ &\leq 2\sigmaA^2
    \opnorm{(I - \SpecMatZero)^{-1}} \frac{\usedim}{\numobs}.
\end{align*}
Therefore, for $n \geq 16 \sigmaA^2 \opnorm{(I - \SpecMatZero)^{-1}} d$, we
obtain the lower bound
\begin{align*}
    T_\bvec \geq \frac{1}{2} \trace \left( (I - \SpecMatZero)^{-1}
    \Sigma_\bvec (I - \SpecMatZero)^{- \top} \right),
\end{align*}
as desired.

\paragraph{Proof of claim~\eqref{eq:algebra-claim2}:}
We note that
\begin{align*}
    \vbarpsilam = \vbarZero - H \projectedbvec_0 + (I - \SpecMatZero) \psi - H
    \psi.
\end{align*}
Consequently, we have
\begin{align*}
    T_\Lmat = \Exs \left[\frac{(\vbarZero)^\top (\vbarZero - H \projectedbvec_0 + (I -
        \SpecMatZero) \psi - H \psi) }{\vecnorm{\vbarZero}{2}^2} \cdot
      \trace \left( (I - \SpecMatZero)^{-1} \Sigma_\Lmat ((I - \SpecMatZero)^{- \top} -
      H^\top) \right)\right].
\end{align*}
Since $\lambda$ is independent of $\psi$ and $H$ is dependent only
upon $\lambda$, by taking expectation with respect to $\psi$, we have
that
\begin{align*}
    T_\Lmat = \Exs \left[\frac{(\vbarZero)^\top (\vbarZero - H \projectedbvec_0)
      }{\vecnorm{\vbarZero}{2}^2} \cdot \trace \left( (I -
      \SpecMatZero)^{-1} \Sigma_\Lmat ((I - \SpecMatZero)^{- \top} - H^\top) \right)\right].
\end{align*}
We note that
\begin{align*}
    \frac{|(\vbarZero)^\top H \projectedbvec_0|}{\vecnorm{\vbarZero}{2}^2} =
    \frac{|(\vbarZero)^\top (I - \SpecMatZero)^{-1}
      \lambda|}{\vecnorm{\vbarZero}{2}^2 - (\vbarZero)^\top (I - \SpecMatZero)}
    \leq 2 \opnorm{(I - \SpecMatZero)^{-1}} \sigmaA \sqrt{\frac{\usedim}{\numobs}}.
\end{align*}
Therefore, for $n \geq 16 \sigmaA^2 \opnorm{(I - \SpecMatZero)^{-1}}^2 d$, we
have the bound $\frac{1}{2}\leq \frac{(\vbarZero)^\top (\vbarZero - H
  \projectedbvec_0) }{\vecnorm{\vbarZero}{2}^2} \leq \frac{3}{2}$, and
consequently, we have
\begin{align*}
    T_\Lmat &\geq \frac{1}{2} \trace \left( (I - \SpecMatZero)^{-1} \Sigma_\Lmat
    (I - \SpecMatZero)^{- \top}\right) - \frac{3}{2} \Exs \abss{\trace
      \left( (I - \SpecMatZero)^{-1} \Sigma_\Lmat H^\top\right) }\\ &\geq
    \frac{1}{2} \trace \left( (I - \SpecMatZero)^{-1} \Sigma_\Lmat (I -
    \SpecMatZero)^{- \top}\right) - \frac{3}{2} \matsnorm{(I - \SpecMatZero)^{-1}
      \Sigma_\Lmat (I - \SpecMatZero)^{- \top}}{\text{nuc}} \cdot \Exs \opnorm{H (I
      - \SpecMatZero)}\\ &\geq \frac{1}{2} \trace \left( (I - \SpecMatZero)^{-1}
    \Sigma_\Lmat (I - \SpecMatZero)^{- \top}\right) \cdot \left(1 - 3 \Exs
    \opnorm{H (I - \SpecMatZero)} \right).
\end{align*}
For the matrix $H (I - \SpecMatZero)$, we have the almost-sure upper bound
\begin{align*}
    \opnorm{H (I - \SpecMatZero)} \leq \frac{2}{\vecnorm{\vbarZero}{2}^2} \opnorm{(I - \SpecMatZero)^{-1} \lambda (\vbarZero)^\top} \leq \frac{2}{\vecnorm{\vbarZero}{2}} \cdot \vecnorm{(I - \SpecMatZero)^{-1} \lambda}{2}  \leq 2 \opnorm{(I - \SpecMatZero)^{-1}} \sigmaA \sqrt{\frac{\usedim}{\numobs}}.
\end{align*}
Thus, provided $n \geq 18 \sigmaA^2 \opnorm{(I - \SpecMatZero)^{-1}}^2 d$, putting together the pieces yields the lower bound
\begin{align*}
    T_\Lmat \geq \frac{1}{3} \trace \left(  (I - \SpecMatZero)^{-1} \Sigma_\Lmat (I - \SpecMatZero)^{- \top}\right).
\end{align*}


\subsection{Proof of Corollary~\ref{cor:lower-bound-final}}
\label{subsec:proof-of-corollary-combined}

Define the terms $\Delta_1 \mydefn \frac{2}{3} \prefact (\SpecMatZero,
\conmax) \delta^2$ and $\Delta_2 \mydefn c \cdot \EstErr_\numobs
(\SpecMatZero, \Sigma_\Lmat + \Sigma_\bvec)$.  We split our proof into
two cases.

\paragraph{Case I: if $\Delta_1 \leq \Delta_2$.} Consider the function class:
\begin{align*}
    \widetilde{\class} \mydefn \bigcup_{(\SpecMat', \projectedbvec')
      \in \neighborhood (\SpecMatZero, \projectedbvecZero)}
    \classApprox (\SpecMat', \projectedbvec', \bigdim, 0, \conmax)
\end{align*}
Clearly, we have the inclusion $\widetilde{\class} \subseteq
\classFinal$. Moreover, for a problem instance in
$\widetilde{\class}$, we have $\AppErr (\LinSpace, \vstar) = 0$, and
consequently $\vstar = \vbar$. Note that the construction of problem
instances in Theorem~\ref{thm:linear-stat-error-lower-bound} can be
embedded in $\Xspace$ of any dimension $\bigdim$, and the linear
operator $\Lmat$ constructed in the proof of
Theorem~\ref{thm:linear-stat-error-lower-bound} satisfies the bound
\begin{align*}
    \hilopnorm{\LmatPsilam} \leq \opnorm{\SpecMatPsilam} \leq
    \opnorm{\SpecMatZero} + \sigmaA \sqrt{\frac{\usedim}{\numobs}}
    \leq \conmax.
\end{align*}
Consequently, the class $\widetilde{\class}$ contains the
population-level problem instances $(\SpecMatPsilam,
\projectedbvecPsilam)$ constructed in the proof of
Theorem~\ref{thm:linear-stat-error-lower-bound}, for any choice of
$\psi, \lambda \in \real^d$.  Invoking
Theorem~\ref{thm:linear-stat-error-lower-bound}, we thus obtain the
sequence of bounds
\begin{align*}
\inf_{\vhat_n \in \Vhatclass_\Xspace } ~\sup_{\substack{(\Lmat, \bvec)
    \in \classFinal\\ \ProbInst_{\Lmat, \bvec} \in \classCov}} \Exs
\statnorm{\vhat_n - \vstar}^2 \geq \inf_{\vhat_n \in
  \Vhatclass_\Xspace } ~\sup_{\substack{(\Lmat, \bvec) \in
    \widetilde\class \\ \ProbInst_{\Lmat, \bvec} \in \classCov}} \Exs
\statnorm{\vhat_n - \vstar}^2 & = \inf_{\vhat_n \in \Vhatclass_\Xspace
} ~\sup_{\substack{(\Lmat, \bvec) \in \widetilde{\class}
    \\ \ProbInst_{\Lmat, \bvec} \in \classCov}} \Exs \statnorm{\vhat_n
  - \vbar}^2 \\
& = \inf_{\vhat_n \in \Vhatclass_\Xspace } ~\sup_{\substack{(\Lmat,
    \bvec) \in \classEst \\ \ProbInst_{\Lmat, \bvec} \in \classCov}}
\Exs \statnorm{\vhat_n - \vbar}^2 \\
& \geq \Delta_2.
\end{align*}

\paragraph{Case II: if $\Delta_1 > \Delta_2$.} In this case, we consider the class of
noise distributions
\begin{align*}
\widetilde{\probClass} \mydefn \classCov (0, 0, \sigmaA, \sigmab).
\end{align*}
Clearly, $\widetilde{\probClass}$ is a sub-class of $\classCov$. Note
that the observation model $\big(\Lmat_i^{(\rade, \plantedvec)},
\bvec_i^{(\rade, \plantedvec)}\big)_{i = 1}^n$ constructed in the
proof of Theorem~\ref{thm:linear-lower-bound} satisfies the following
identities almost surely:
\begin{align*}
    \PhiOp \Lmat_i^{(\rade, \plantedvec)} \adjoint{\PhiOp} = \PhiOp
    \Lmat^{(\rade, \plantedvec)} \adjoint{\PhiOp}, \quad \PhiOp
    \bvec_i^{(\rade, \plantedvec)} = \PhiOp \bvec^{(\rade,
      \plantedvec)} .
\end{align*}
So the problem instances constructed in the proof of
Theorem~\ref{thm:linear-lower-bound} belongs to class
$\widetilde{\probClass}$. Invoking
Theorem~\ref{thm:linear-lower-bound}, we obtain the bound
\begin{align*}
     \inf_{\vhat_n \in \Vhatclass_\Xspace } ~\sup_{\substack{(\Lmat,
         \bvec) \in \classFinal\\ \ProbInst_{\Lmat, \bvec} \in
         \classCov}} \Exs \statnorm{\vhat_n - \vstar}^2 \geq
     \inf_{\vhat_n \in \Vhatclass_\Xspace } ~\sup_{\substack{(\Lmat,
         \bvec) \in \classFinal\\ \ProbInst_{\Lmat, \bvec} \in
         \widetilde{\probClass}}} \Exs \statnorm{\vhat_n - \vstar}^2
     \geq \frac{2}{3} (\prefact (\SpecMatZero, \conmax) - 1) \oracleErr^2
     = \Delta_1.
\end{align*}

Combining the results in two cases, we arrive at the lower bound:
\begin{multline*}
       \inf_{\vhat_n \in \Vhatclass_\Xspace } ~\sup_{\substack{(\Lmat,
           \bvec) \in \classFinal\\ \ProbInst_{\Lmat, \bvec} \in
           \classCov}} \Exs \statnorm{\vhat_n - \vstar}^2 \geq \max
       (\Delta_1, \Delta_2) \\ \geq \frac{1}{2} \Delta_1 + \frac{1}{2}
       \Delta_2 \geq \frac{1}{3} (\prefact (\SpecMatZero, \conmax) - 1)
       \oracleErr^2 + \frac{c}{2} \EstErr_\numobs (\SpecMat,
       \Sigma_\Lmat + \Sigma_\bvec),
\end{multline*}
which completes the proof of the corollary.


\section{Discussion}
\label{SecDiscussion}

In this paper, we studied methods for computing approximate solutions
to fixed point equations in Hilbert spaces, using methods that search
over low-dimensional subspaces of the Hilbert space, and operate on
stochastic observations of the problem data.  We analyzed a standard
stochastic approximation scheme involving Polyak--Ruppert averaging,
and proved non-asymptotic instance-dependent upper bounds on its
mean-squared error.  This upper bound involved a pure approximation
error term, reflecting the discrepancy induced by searching over a
finite-dimensional subspace as opposed to the Hilbert space, and an
estimation error term, induced by the noisiness in the observations.
We complemented this upper bound with an information-theoretic
analysis, that established instance-dependent lower bounds for both
the approximation error and the estimation error. A noteworthy
consequence of our analysis is that the optimal approximation factor
in the oracle inequality is neither unity nor constant, but a quantity
depending on the projected population-level operator.  As direct
consequences of our general theorems, we showed oracle inequalities for
three specific examples in statistical estimation: linear regression
on a linear subspace, Galerkin methods for elliptic PDEs, and value
function estimation via temporal difference methods in Markov reward processes. \\

\noindent The results of this paper leave open a number of directions
for future work:
\bcar
\item This paper focused on the case of independently drawn
  observations.  Another observation model, one which arises naturally
  in the context of reinforcement learning, is the Markov observation
  model. As discussed in Section~\ref{ex:td-discounted}, we consider
  the problem with $\Lmat = \discount \transition$ and $\bvec =
  \reward$, where $\transition$ is a Markov transition kernel,
  $\discount$ is the discount factor and $\reward$ is the reward
  function. The observed states and rewards in this setup are given by
  a single trajectory of the Markov chain $\transition$, instead of
  $\mathrm{i.i.d.}$ from the stationary distribution. It is
  known~\cite{tsitsiklis1997analysis} that the resolvent formalism
  (a.k.a. TD$(\lambda)$) leads to an improved approximation factor
  with larger $\lambda \in [0, 1)$. On the other hand, larger choice
    of $\lambda$ may lead to larger variance and slower convergence
    for the stochastic approximation estimator, and a model selection
    problem exists (See Section 2.2 in the
    monograph~\cite{szepesvari2010algorithms} for a detailed
    discussion). It is an important future work to extend our
    fine-grained risk bounds to the case of TD$(\lambda)$ methods with
    Markov data. Leveraging the instance-dependent upper and lower
    bounds, one can also design and analyze estimators that achieve
    the optimal trade-off.
  \item This paper focused purely on oracle inequalities defined with
    respect to a subspace.  However, the framework of oracle
    inequalities is far more general; in the context of statistical
    estimation, one can prove oracle inequalities for any star-shaped set
    with bounds on its metric entropy. (See Section 13.3 in the
    monograph~\cite{wainwright2019high} for the general mechanism and
    examples.) For all the three examples considered in
    Section~\ref{SecExamples}, one might imagine approximating
    solutions using sets with nonlinear structure, such as those
    defined by $\ell_1$-constraints, Sobolev ellipses, or the function
    class representable by a given family neural networks.  An
    interesting direction for future work is to understand the complexity
    of projected fixed point equations defined by such approximating
    classes.
\ecar

\section*{Acknowledgement}
We would like to thank Peter Bartlett for helpful discussions.


{\small{ \bibliography{main-linear} }}

\appendix

\section{Proof of Theorem~\ref{thm:lower-bound-approx-factor-improved}}
\label{AppIntermediate}

We begin by defining some additional notation needed in the proof.
For a non-negative integer $k$, we use $\hadamard{k} \in \real^{2^k
  \times 2^k}$ to denote the Hadamard matrix of order $k$, recursively
defined as:
\begin{align*}
\hadamard{0} \mydefn 1, \quad \hadamard{k} \mydefn \begin{bmatrix}
  \hadamard{k - 1} & \hadamard{k - 1}\\ \hadamard{k - 1} & -
  \hadamard{k - 1}
\end{bmatrix}, ~\mbox{for}~ k = 1,2,\cdots
\end{align*}
For any integer $q \geq 2$, we define
\begin{align*}
    J_q \mydefn \begin{bmatrix} 
    0 &1 & 0 &\cdots& 0\\
    0 & 0 & 1 &\cdots & 0\\
    & \vdots && \vdots &\\
    0 & 0& \cdots & 0 & 1\\
    0 & 0& \cdots & 0 & 0
    \end{bmatrix},
\end{align*}
which is a $q \times q$ Jordan block with zeros in the diagonal.\\

Now we turn to the proof of the theorem. We assume that $\bigdim$ is
an integer multiple of $q$, and that $m \mydefn \frac{\bigdim -
  \usedim}{q}$ is an integer power of $2$; the complementary case can
be handled by adjusting the constant factors in our bounds.  Similarly
to the proof of Theorem~\ref{thm:linear-lower-bound}, we let $u \in
\sphere^{\usedim - 1}$ be an eigenvector associated to the largest
eigenvalue of the matrix $(I - \SpecMatZero)^{- 1} \big( \conmax^2 I
- \SpecMatZero \SpecMatZero^\top \big) (I - \SpecMatZero)^{- \top}$,
and define the $\usedim$-dimensional vectors:
\begin{align*}
    w \mydefn \sqrt{\prefact (\SpecMatZero, \conmax) - 1} \cdot (I -
    \SpecMatZero) u, \quad \mbox{and}\quad \plantedvec \mydefn
    \sqrt{\prefact (\SpecMatZero, \conmax) - 1} \cdot \oracleErr u.
\end{align*}
We first construct the following $(\bigdim - \usedim) \times (\bigdim-
\usedim)$ block matrix, indexed by the bits $(\rade_{ij})_{1 \leq i
  \leq q, 1 \leq j \leq m}$:
\begin{subequations}
  \begin{align}
\label{eq:lb-refined-construction-J}    
    J^{(\rade)} \mydefn
    \begin{bmatrix}
    I_m & \mathrm{diag} (\rade_{1j} \cdot \rade_{2j})_{j = 1}^m & 0 &
    \cdots&0&0\\ 0 & I_m & \mathrm{diag} (\rade_{2j} \cdot
    \rade_{3j})_{j = 1}^m & \cdots&0&0\\ & \vdots &\ddots &\ddots & &
    \vdots\\ 0 & 0 & \cdots& & I_m & \mathrm{diag} (\rade_{(q - 1) j}
    \cdot \rade_{qj})_{j = 1}^m \\ 0 & 0 & \cdots & & 0 & I_m
    \end{bmatrix}.
\end{align}
Each submatrix depicted above is an $m \times m$ matrix, and the
diagonal blocks are given by identity matrices.  We use this
construction to define the population-level instance $(\Lmat^{(\rade,
  \plantedbit)}, \bvec^{(\rade, \plantedbit)})$ as follows:
\begin{align}
    \Lmat^{(\rade, \plantedbit)} = \begin{bmatrix} \SpecMatZero &
      \frac{\sqrt{d / 2}}{D - d} \plantedbit\rade_{11}w& \frac{\sqrt{d
          / 2}}{D - d} \plantedbit\rade_{12}w & \cdots& \frac{\sqrt{d
          / 2}}{D - d} \plantedbit\rade_{1m}w & 0 &\cdots& 0\\ 0
      \\ 0\\ \vdots&&& \frac{1}{2}J^{(\rade)}\\ 0
    \end{bmatrix} , \quad
    \bvec^{(\rade, \plantedbit)} = \begin{bmatrix} \sqrt{2d} \projectedbvecZero \\ 0\\ \vdots \\0
      \\ \frac{\oracleErr}{\sqrt{2}}
      \rade_{q1}\\ \vdots\\ \frac{\oracleErr}{\sqrt{2}} \rade_{qm}
    \end{bmatrix}. \label{eq:lb-refined-construction-A}
\end{align}
It can then be verified that the solution to the fixed point equation
$\vstar_{\rade, \plantedbit} = (I - \Lmat^{(\rade, \plantedbit)})^{-1}
b^{(\rade, \plantedbit)}$ is given by
\begin{align}
    \label{eq:lb-refined-construction-theta-b}  
    \vstar_{\rade, \plantedbit} = \begin{bmatrix} \frac{\sqrt{d}}{q}
      \plantedbit \plantedvec + \sqrt{2d} (I - \SpecMatZero)^{-1}
      \projectedbvecZero\\ \sqrt{2} \oracleErr \rade_{11}\\ \sqrt{2}
      \oracleErr \rade_{12}\\ \vdots\\ \sqrt{2} \oracleErr \rade_{q
        1}\\ \vdots\\ \sqrt{2} \oracleErr \rade_{qm}
    \end{bmatrix}.
\end{align}
\end{subequations}
Similarly to the proof of Theorem~\ref{thm:linear-lower-bound}, we
take the subspace $\lspace$ to be the one spanned by first $d$
coordinates, and take the weight vector be
\begin{align*}
    \stationary = \begin{bmatrix} \undermat{d}{\frac{1}{2 d} & \cdots
        & \frac{1}{2d}} & \undermat{(D - d)}{\frac{1}{2 (D - d)} &
        \cdots & \frac{1}{2(D - d)}} \end{bmatrix}.
\end{align*}
The inner product is then given by $\statinprod{p}{p'} \mydefn \sum_{j
  = 1}^{\bigdim} p_j \stationary_j p_j'$ for vectors $p, p' \in
\real^\bigdim$.

It remains to define our basis vectors. For $j \in [d]$, we let
$\phi_j \mydefn \sqrt{2d} e_j$.  For the orthogonal complement
$\lspace^\perp$, we use the basis vectors
\begin{align*}
    \begin{bmatrix}\phi_{d + 1} & \phi_{d + 2} & \cdots &\phi_{d + qm} \end{bmatrix} = \begin{bmatrix} 0\\
    \sqrt{2q}~ \hadamard{\log_2 m} \otimes I_q \end{bmatrix}.
\end{align*}
Recall that $\hadamard{k} \in \real^{2^k \times 2^k}$ denotes the
Hadamard matrix of order $k$, and $\otimes$ denotes Kronecker
product. The first $\usedim$ rows of the matrix are zeros, while the
following $\bigdim - \usedim = m q$ columns are given by the Kronecker
product.  By the definition of Hadamard matrix, we have
$\statnorm{\phi_i} = 1$ and $\statinprod{\phi_i}{\phi_j} = 0$ for $i
\neq j$.

As before, the construction of
equations~\eqref{eq:lb-refined-construction-J}-\eqref{eq:lb-refined-construction-theta-b}
ensures that for any choice of the binary string $\rade \in
\{-1,1\}^{mq}$ and bit $\plantedvec \in \{-1, 1\}$, the oracle
approximation error is equal to
\begin{subequations}
  \begin{align}
\label{eq:oracle-error-in-refined-lb-construction}    
    \AppErr (\LinSpace, \vstar_{\rade, \plantedbit}) = \inf_{\vvec
      \in \lspace} \statnorm{\vstar_{\rade, \plantedbit} - \vvec}^2 =
    \frac{1}{2 (D - d)} \sum_{i = 1}^q \sum_{j = 1}^m (\sqrt{2}
    \oracleErr \rade_{ij})^2 = \oracleErr^2.
\end{align}

Furthermore, straightforward calculation yields that the projected
matrix-vector pair takes the form
\begin{align}
\label{eq:projected-instances-in-refined-lb-construction}  
    \PhiOp \Lmat^{(\rade, \plantedbit)} \adjoint{\PhiOp} =
    \SpecMatZero, \quad \mbox{and} \quad \PhiOp \bvec^{(\rade,
      \plantedbit)} = \projectedbvecZero.
\end{align}
\end{subequations}

Now, we construct our observation model from which samples
$\big(\Lmat_i^{(\rade,\plantedbit)}, \bvec_i^{(\rade,
  \plantedbit)}\big)_{i = 1}^n$ are generated. For each $i \in [n]$
and $j \in [m]$, we sample independently Bernoulli random variables
$\chi_{0j}^{(i)}, \chi_{1j}^{(i)}, \cdots, \chi_{qj}^{(i)} \simiid
\mathrm{Ber} (1/m)$. The random observations are then generated by the
random matrix
\begin{subequations}
\begin{align}
    J^{(\rade)}_{i} \mydefn
    \begin{bmatrix}
    I_m & \mathrm{diag} \left(m \chi_{1j}^{(i)} \rade_{1j} \rade_{2j}
    \right)_{j = 1}^m & 0 & \cdots&0&0\\ 0& I_m & \mathrm{diag}
    \left(m \chi_{2j}^{(i)} \rade_{2j} \rade_{3j} \right)_{j = 1}^m &
    \cdots&0&0\\ & \vdots &\ddots &\ddots&&\vdots\\ 0& 0 & \cdots& &
    I_m &\mathrm{diag} \left(m \chi_{(q - 1) j}^{(i)} \rade_{(q - 1)
      j} \rade_{q j} \right)_{j = 1}^m \\ 0&0&\cdots&& 0 & I_m
    \end{bmatrix}, \label{eq:lb-refined-construction-J-obs}
\end{align}
where once again, the diagonal blocks correspond to $m \times m$
identity matrices.  We use this random matrix to generate the
observations
\begin{align}
    \Lmat_i^{(\rade, \plantedbit)} = \begin{bmatrix} \SpecMatZero &
      \chi_{01} \frac{\sqrt{d / 2}}{q} \plantedbit \rade_{11}w&
      \chi_{02} \frac{\sqrt{d / 2}}{q} \plantedbit \rade_{12}w &
      \cdots& \chi_{0m} \frac{\sqrt{d / 2}}{q} \plantedbit \rade_{1m}w
      & 0 &\cdots& 0\\ 0 \\ 0\\ \vdots&&& \frac{1}{2}J_i^{(\rade)}\\ 0
    \end{bmatrix} \label{eq:lb-refined-construction-A-obs}
\end{align}
and
\begin{align}
     b_i^{(\rade, \plantedbit)} = \begin{bmatrix} \sqrt{2d}
       \projectedbvecZero^\top & \undermat{D - d - m}{ 0 & \cdots &0
       }& m \chi_{q1} \frac{\oracleErr}{\sqrt{2}} \rade_{q1}& \cdots&
       m \chi_{qm }\frac{\oracleErr}{\sqrt{2}} \rade_{qm}
    \end{bmatrix}^\top. \label{eq:lb-refined-construction-b-obs}
\end{align}
\vspace{0.1cm}
\end{subequations}

This concludes our description of the problem instances themselves. As
before, our proof proceeds via Le Cam's lemma, and we use similar
notation for product distributions and mixtures under this observation
model.  Let $\ProbInst_{\rade, \plantedbit}^{(\numobs)}$ denote the
$\numobs$-fold product of the probability laws of $(\Lmat_i^{(\rade,
  \plantedbit)}, \bvec_i^{(\rade, \plantedbit)})$. We also define the
following mixture of product measures for each $\plantedbit \in \{-1,
1\}$:
\begin{align*}
    \ProbInst^{(n)}_\plantedbit \mydefn \frac{1}{2^{\bigdim -
        \usedim}} \sum_{\rade \in \{\pm 1\}^{m \times q}}
    \ProbInst_{\rade, \plantedbit}^{(\numobs)}.
\end{align*}
We seek bounds on the total variation distance
\begin{align*}
\Delta \mydefn \totalvariation \left( \ProbInst^{(\numobs)}_1 ,
\ProbInst^{(\numobs)}_{-1} \right).
\end{align*}

With this setup, the following lemmas assert that (a) Our construction
satisfies the operator norm condition and the noise conditions in
Assumption~\ref{assume-second-moment-strong} with the associated
parameters bounded by dimension-independent constants, and (b) The
total variation distance $\Delta$ is small provided $n \ll m^{1 +
  1/q}$.

\begin{lemma}
\label{lemma:lb-refined-satisfies-assumptions}
    For $q \in \left[2, \frac{1}{\sqrt{2 (1 - 1 \wedge\conmax)}}
      \right]$, and any $\rade \in \{-1, 1\}^{m \times q}$ and
    $\plantedbit \in \{-1, 1\}$,\\ (a) The construction in
    equation~\eqref{eq:lb-refined-construction-A} satisfies the bound
    $\hilopnorm{\Lmat^{(\rade, \plantedbit)}} \leq \conmax$.\\ (b) The
    observation model defined in
    equation~\eqref{eq:lb-refined-construction-J-obs}-\eqref{eq:lb-refined-construction-b-obs}
    satisfies Assumption~\ref{assume-second-moment-strong} with
    $\sigmaA = \conmax + 1$ and $\sigmab = \oracleErr / q$.
\end{lemma}

\begin{lemma}\label{lemma:tv-bound-in-refined-approx-factor-lb-proof}
    Under the setup above, we have
    \begin{align*}
        \Delta \leq \frac{12 n^{q + 1}}{m^{q}}.
    \end{align*}
\end{lemma}

Part (a) of Lemma~\ref{lemma:lb-refined-satisfies-assumptions} and
equation~\eqref{eq:oracle-error-in-refined-lb-construction}-\eqref{eq:projected-instances-in-refined-lb-construction}
together ensure that population-level problem instance $(\Lmat,
\bvec)$ we constructed belongs to the class $\classApprox
(\SpecMatZero, \projectedbvecZero, \bigdim, \oracleErr,
\conmax)$. Part (b) of
Lemma~\ref{lemma:lb-refined-satisfies-assumptions} further ensures the
probability distribution $\ProbInst_{\Lmat, \bvec}$ belongs to the
class $\classNoise (\sigmaA,
\sigmab)$. Lemma~\ref{lemma:tv-bound-in-refined-approx-factor-lb-proof}
ensures that the two mixture distributions corresponding to different
choices of the bit $\plantedbit$ are close provided $\numobs$ is not
too large. The final step in applying Le Cam's mixture-vs-mixture
result is to show that the approximation error is large for at least
one of the choices of the bit $\plantedbit$.

Given any pair $\rade, \rade' \in \{-1, 1\}^{q \times m}$, we note
that
\begin{align*}
  \statnorm{\vstar_{\rade, 1} - \vstar_{\rade', -1}} \geq
  \statnorm{\big[\begin{matrix} 2 \sqrt{\frac{\usedim}{q}}
        \plantedvec^\top & 0 & \cdots &0 \end{matrix} \big]^\top} =
  \frac{\sqrt{2}}{q} \vecnorm{\plantedvec}{2} =\frac{\sqrt{2}}{q}
  \sqrt{\prefact (\SpecMatZero, \conmax) - 1} \cdot \oracleErr.
\end{align*}
Applying triangle inequality and Young's inequality, we have the bound
\begin{align*}
    \frac{1}{2} (\statnorm{\vhat - \vstar_{\rade, 1}}^2 +
    \statnorm{\vhat - \vstar_{\rade', 1}}^2) \geq \frac{1}{4}
    (\statnorm{\vhat - \vstar_{\rade, 1}} + \statnorm{\vhat -
      \vstar_{\rade', 1}})^2 \geq \frac{1}{4} \statnorm{\vstar_{\rade,
        1} - \vstar_{\rade', -1}}^2 \geq \frac{\prefact (\SpecMatZero,
      \conmax) - 1}{2 q^2} \oracleErr^2.
\end{align*}
Finally, applying Le Cam's lemma yields
\begin{align*}
     \inf_{\vhat_n \in \Vhatclass_{\Xspace}} ~\sup_{\substack{ (\Lmat,
         \bvec) \in \classApprox \\ \ProbInst_{\Lmat, \bvec} \in
         \classNoise (\sigmaA, \sigmab) }} \Exs \statnorm{\vhat_n -
       \vstar}^2 \geq \frac{\prefact (\SpecMatZero, \conmax) - 1}{2
       q^2} \oracleErr^2 \cdot \left(1 - \totalvariation
     (\ProbInst_{-1}^{(n)}, \ProbInst_1^{(n)}) \right)
\end{align*}
and using Lemma~\ref{lemma:tv-bound-in-refined-approx-factor-lb-proof}
in conjunction with the condition $D \geq d + 3 q n^{1 + 1/q}$, we
arrive at the final bound
\begin{align*}
     \inf_{\vhat_n \in \Vhatclass_\Xspace} ~\sup_{\substack{ (\Lmat,
         \bvec) \in \classApprox \\ \ProbInst_{\Lmat, \bvec} \in
         \classNoise (\sigmaA, \sigmab) }} \Exs \statnorm{\vhat_n -
       \vstar}^2 &\geq \frac{\prefact (\SpecMatZero, \conmax) - 1}{2
       q^2} \oracleErr^2
\end{align*}
as desired.


\subsection{Proof of Lemma~\ref{lemma:lb-refined-satisfies-assumptions}}

We prove the two parts of the lemma separately.

\paragraph{Proof of part $(a)$:}
We first show the upper bound on the operator norm. For any vector $p
\in \real^\bigdim$, we employ the decomposition $p =
\left[ \begin{smallmatrix} p^{(1)}\\p^{(2)} \end{smallmatrix} \right]$
with $p^{(1)} \in \real^\usedim$ and $p^{(2)} \in
\real^{qm}$. Assuming that $\statnorm{p}^2 = \frac{1}{2\usedim}
\vecnorm{p^{(1)}}{2}^2 + \frac{1}{2 (\bigdim - \usedim)}
\vecnorm{p^{(2)}}{2}^2 = 1$, we have that
\begin{align*}
    \statnorm{\Lmat^{(\rade, \plantedbit)} p}^2 &= \frac{1}{2d}
    \vecnorm{\SpecMatZero p^{(1)} + w \cdot \frac{\plantedbit
        \sqrt{d}}{D - d} \sum_{j = 1}^m \rade_{1j} p_j^{(2)}}{2}^2 +
    \frac{1}{2 (D - d)} \vecnorm{\frac{1}{2} J^{(\rade)}
      p^{(2)}}{2}^2.
\end{align*}
Define the vector $a_1 \mydefn \frac{1}{\sqrt{2d}} p^{(1)}$ and scalar
$a_2 \mydefn \frac{1}{\sqrt{2 (\bigdim - \usedim)}}
\vecnorm{p^{(2)}}{2}$ for convenience; we have the identity
$\vecnorm{a_1}{2}^2 + a_2^2 = 1$. Following the same arguments as in
the proof of
Lemma~\ref{lemma-approx-lower-bound-satisfies-conditions}, we then
have
\begin{multline*}
   \frac{1}{2 \usedim} \vecnorm{\SpecMat p^{(1)} + w \cdot
     \frac{\plantedbit \sqrt{d}}{D - d} \sum_{j = 1}^m \rade_{1j}
     p_j^{(2)}}{2}^2 \leq \frac{1}{2 \usedim} \sup_{t \in [-1, 1]}
   \vecnorm{\SpecMatZero p^{(1)} + \frac{\sqrt{\usedim}}{q} a_2 t w
   }{2}^2\\ \leq \max \left( \vecnorm{\SpecMatZero a_1 + \frac{1}{q
       \sqrt{2}} a_2 w}{2}^2, \vecnorm{\SpecMatZero a_1 - \frac{1}{q
       \sqrt{2}} a_2 w}{2}^2 \right) \leq \opnorm{\begin{bmatrix}
       \SpecMatZero & w
      \end{bmatrix}}^2 \left( \vecnorm{a_1}{2}^2 + \frac{1}{2 q^2} a_2^2 \right).
\end{multline*}
By the definition of the vector $w$, we have the bound
\begin{align*}
    \opnorm{\begin{bmatrix} \SpecMatZero & w
      \end{bmatrix}}^2 = \lammax \left( \SpecMatZero \SpecMatZero^\top + ww^\top \right) \leq \conmax^2.
\end{align*}
On the other hand, note that
\begin{align*}
    \frac{1}{2 (D - d)} \vecnorm{\frac{1}{2} J^{(\rade)} p^{(2)}}{2}^2
    \leq \frac{1}{8 (\bigdim - \usedim)} \opnorm{J^{(\rade)}}^2 \cdot
    \vecnorm{p^{(2)}}{2}^2 = \frac{1}{4}\opnorm{J^{(\rade)}}^2 a_2^2,
\end{align*}
and consequently, that
\begin{align}
\label{eq:statnorm-in-lb-refined-assumption-proof}  
     \statnorm{\Lmat^{(\rade, z)} p}^2 \leq \conmax^2 \left(
     \vecnorm{a_1}{2}^2 + \frac{1}{2 q^2} a_2^2 \right) + \frac{1}{4}
     \opnorm{J^{(\rade)}}^2 \cdot a_2^2.
\end{align}
In order to bound the operator norm of the matrix $J^{(\rade)}$, we
use the following fact about operator norm, proved at the end of this
section for convenience.  For any block matrix $T = [T_{ij}]_{1 \leq
  i, j \leq q}$, with each block $T_{ij} \in \real^{m \times m}$, we
have
\label{eq:fact-opnorm-block-matrix}
\begin{align}
  \opnorm{[T_{ij}]_{1 \leq i, j \leq q}} \leq \opnorm{\big[
      \opnorm{T_{ij}} \big]_{1 \leq i, j \leq q}}.
\end{align}

Applying equation~\eqref{eq:fact-opnorm-block-matrix} to matrix
$J^{(\rade)}$ yields the bound
\begin{align*}
    \opnorm{J^{(\rade)}} \leq\opnorm{I_q + J_q}.
\end{align*}
Recall that $J_q$ is the Jordan block of size $q$ with zeros in the diagonal.
Straightforward calculation yields the bound
\begin{align*}
    (I_q + J_q) (I_q + J_q)^\top \preceq \left[ \begin{matrix} 2 &1 &
      0 &\cdots & 0& 0\\ 1 & 2 & 1 &\cdots &0& 0\\ & \vdots && \vdots
      &&\\ 0 & 0& \cdots &1 & 2 & 1\\ 0 & 0& \cdots & 0 & 1 & 2
    \end{matrix} \right] =: C_q.
\end{align*}
Note that $C_q$ is a tridiagonal Toeplitz matrix, whose norm admits
the closed-form expression
\begin{align*}
    \opnorm{C_q} = 2 + 2 \cos \left( \frac{\pi }{q + 1} \right) \leq 4 - \frac{4}{q^2}.
\end{align*}
Therefore, we have $\opnorm{J^{(\rade)}} \leq \sqrt{\opnorm{C_q}} \leq
\sqrt{4 - 4 / q^2}$. Substituting into
equation~\eqref{eq:statnorm-in-lb-refined-assumption-proof}, we obtain
\begin{align*}
     \statnorm{\Lmat^{(\rade, \plantedbit)} p}^2 \leq \conmax^2 \left(
     \vecnorm{a_1}{2}^2 + \frac{1}{2 q^2} a_2^2 \right) + \left(1 -
     \frac{1}{q^2} \right) \cdot a_2^2.
\end{align*}
Invoking the condition $q \leq \frac{1}{\sqrt{2 (1 - 1
    \wedge\conmax)}}$, we have the bound
\begin{align*}
      \statnorm{\Lmat^{(\rade, \plantedbit)} p}^2 \leq \conmax^2 \left( \vecnorm{a_1}{2}^2 +
      \frac{a_2^2}{2q} \right) + \left(1 - \frac{1}{q^2} \right) a_2^2
      \leq \conmax^2 (\vecnorm{a_1}{2}^2 + a_2^2) = \conmax^2,
\end{align*}
Since the choice of the vector $p$ is arbitrary, this yields
\begin{align*}
    \hilopnorm{\Lmat^{(\rade, \plantedbit)}} \leq \conmax,
\end{align*}
which completes the proof.


\paragraph{Proof of part $(b)$:}
Next, we verify the noise conditions in
Assumption~\ref{assume-second-moment-strong}. For a vector $p
\in \Xspace$ such that $\statnorm{p} = 1$, denote it with $p =
\left[ \begin{smallmatrix} p^{(1)}\\p^{(2)}
  \end{smallmatrix}\right]$, where $p^{(1)} \in \real^\usedim$ and $p^{(2)} \in \real^{\bigdim - \usedim}$.
We have the identity $\frac{1}{2d} \vecnorm{p^{(1)}}{2}^2 + \frac{1}{2
  (\bigdim - \usedim)} \vecnorm{p^{(2)}}{2}^2 = 1$.

For the noise $\bvec_i^{(\rade, \plantedbit)} - \bvec^{(\rade,
  \plantedbit)}$, we note that
\begin{multline*}
    \Exs \statinprod{p}{\bvec_i^{(\rade, \plantedbit)} -
      \bvec^{(\rade, \plantedbit)}}^2 \leq \frac{1}{4 (\bigdim -
      \usedim)^2} \sum_{j = 1}^m \Exs \left( (m \chi_{qj} - 1)
    \frac{\oracleErr}{\sqrt{2}} \rade_{qj} \right)^2 \big(p_{(q - 1)m
      + j}^{(2)}\big)^2\\ \leq \frac{1}{4 (\bigdim - \usedim)^2} \cdot
    \frac{m \oracleErr^2}{2} \vecnorm{p^{(2)}}{2}^2 \leq
    \frac{\oracleErr^2}{q^2}.
\end{multline*}
Consequently, equation~\eqref{eq:assume-second-moment-b-noise-strong}
is satisfied for $\sigmab = \oracleErr / q$.

In order to bound the noise in the $\Lmat$ component, we consider the
first $\usedim$ basis vectors and the last $(\bigdim - \usedim)$ basis
vectors separately. First, note that for each $k \in [\usedim]$, we
have
\begin{align*}
    \Exs \statinprod{\phi_k}{\big(\Lmat_i^{(\rade, \plantedbit)} -
      \Lmat^{(\rade, \plantedbit)} \big) p}^2&= \frac{1}{4 d^2} \Exs
    \left( \frac{\sqrt{d/2}}{D - d} \sum_{j = 1}^m z (m
    \chi_{1j}^{(i)} - 1)\rade_{1j} p^{(2)}_j \cdot \sqrt{2 \usedim}
    w^\top e_k \right)^2\\ &\leq \frac{1}{4} \vecnorm{w}{2}^2 \cdot
    \frac{m}{(D - d)^2} \sum_{j = 1}^m \big(p^{(2)}_j\big)^2 \\ &\leq
    \frac{\vecnorm{w}{2}^2}{4 q}.
\end{align*}
Following the derivation in
Lemma~\ref{lemma-approx-lower-bound-satisfies-conditions}, we have
$\vecnorm{w}{2} \leq \conmax$, and consequently, we have the bound
\begin{align*}
    \Exs \statinprod{\phi_k}{\big(\Lmat_i^{(\rade, \plantedbit)} -
      \Lmat^{(\rade, \plantedbit)} \big) p}^2 \leq \conmax^2.
\end{align*}

On the other hand, for $k \geq d + 1$, the basis vector $\phi_k$ is
constructed through the Hadamard matrix. Let $\psi^{(k)} \defn
(2q)^{-1/2} \cdot \phi_k$, and note that the entries of $\psi^{(k)}$
are uniformly bounded by~$1$. Letting $k - d = (k_0 - 1) m + k_1$ for
some choice of integers $k_0 \in [q]$ and $k_1 \in [m]$, we have
\begin{align*}
    &\Exs \statinprod{\phi_k}{(\Lmat_i^{(\rade, \plantedbit)} -
    \Lmat^{(\rade, \plantedbit)}) p}^2\\ &\leq \frac{1}{4 (D - d)^2}
  \Exs \left( \sum_{j = 1}^m (m \chi_{k_0 j}^{(i)} - 1) \rade_{k_0 j}
  \rade_{(k_0 + 1) j} \cdot \sqrt{2 q} \psi^{(k)}_{d + (k_0 - 1) m +
    j} \cdot p^{(2)}_{k_0 m + j} \right)^2\\ &\leq \frac{2m q}{4 (D -
    d)^2} \sum_{j = 1}^m \big(\psi^{(k)}_{d + (k_0 - 1) m + j}\big)^2
  \cdot \big( p^{(2)}_{k_0 m + j} \big)^2 \\ &\leq \frac{1}{2 (D - d)}
  \vecnorm{p^{(2)}}{2}^2 = \statnorm{p^{(2)}}^2 \leq 1.
\end{align*}
This verifies that
equation~\eqref{eq:assume-second-moment-L-noise-strong} is satisfied
with parameter $\sigmaA = \conmax + 1$.


\paragraph{Proof of equation~\eqref{eq:fact-opnorm-block-matrix}:}

For any vector $x \in \real^{mq}$, consider the decomposition $x^\top
= \begin{bmatrix}x_1^\top & \cdots &x_q^\top \end{bmatrix}$ with $x_j
\in \real^m$ for each $j \in [q]$.  We have the bound
\begin{align*}
    \vecnorm{T x}{2}^2 &= \sum_{i = 1}^q \vecnorm{\sum_{j = 1}^q
      T_{ij} x_j}{2}^2 \leq \sum_{i = 1}^q \left( \sum_{j = 1}^q
    \opnorm{T_{ij}} \vecnorm{ x_j }{2} \right)^2\\ &= \vecnorm{\big[
        \opnorm{T_{ij}} \big]_{1 \leq i, j \leq q} \cdot \big[
        \vecnorm{x_j}{2} \big]_{1 \leq j \leq q}}{2}^2 \leq
    \opnorm{\big[ \opnorm{T_{ij}} \big]_{1 \leq i, j \leq q}}^2 \cdot
    \vecnorm{x}{2}^2,
\end{align*}
which proves this inequality.

\subsection{Proof of Lemma~\ref{lemma:tv-bound-in-refined-approx-factor-lb-proof}}

For each $j \in [m]$, define the event
\begin{align}
    \Event_j \mydefn \left\{ \text{ for all } i \in \{0, 1, \cdots,
    q\}, \text{ there exists } \ell_i \in [n] \text{ such that }
    \chi_{ij}^{(\ell_i)} = 1
    \right\}. \label{eq:bad-event-generalized-birthday}
\end{align}

Let $\Event \mydefn \bigcup_{j = 1}^m \Event_j$.  We note that under
both $\ProbInst_1^{(n)}$ and $\ProbInst_{-1}^{(n)}$, we have the
inequality
\begin{align*}
    \Prob (\Event) \leq \sum_{j = 1}^m \Prob (\Event_j) = \sum_{j =
      1}^m \prod_{i = 0}^q \left\{ 1 - \Prob \left( \chi_{ij}^{(\ell)}
    = 1 \text{ for all } \ell \in [n] \right) \right\} = m \cdot
    \left(1 - \left(\frac{m - 1}{m} \right)^n \right)^{q + 1} \leq
    \frac{n^{q + 1}}{m^q}.
\end{align*}
As before, our choice of the event $\Event$ was guided by the fact
that the two mixture distributions are identical on its complement. In
particular, we claim that
    \begin{align} \label{eq:identical}
        \ProbInst_1^{(n)} | \Event^C = \ProbInst_{-1}^{(n)} | \Event^C.
    \end{align}
Taking this claim as given for the moment and applying
Lemma~\ref{lemma:tv-bound-with-event}, we arrive at the bound
\begin{align*}
    \totalvariation (\ProbInst_1^{(n)}, \ProbInst_{-1}^{(n)}) \leq
    \frac{12 n^{q + 1}}{m^q}.
\end{align*}
This completes the proof of this lemma. It remains to prove
equation~\eqref{eq:identical}.

\paragraph{Proof of equation~\eqref{eq:identical}:}

We first note that both $\ProbInst_1^{(n)}$ are $\ProbInst_{-1}^{(n)}$
are $m$-fold $\mathrm{i.i.d.}$ product distributions: given $z = 1$ or
$z = -1$, the random objects $\left( (\rade_{ij})_{1 \leq i \leq q},
(\chi_{ij}^{(\ell)})_{0 \leq i \leq q, 1 \leq \ell \leq n} \right)_{j
  = 1}^m$ are independent and identically distributed.  Now for each
$j \in [m]$, let $\measureQ_{z, j}^{(n)}$ be the joint law of the
random object $(\rho^{(\ell)})_{\ell = 1}^n$, where $\rho^{(\ell)} =
\left( z \chi_{0j}^{(\ell)} \rade_{1j}, \chi_{1j}^{(\ell)} \rade_{1j}
\rade_{2j}, \cdots, \chi_{(q - 1)j}^{(\ell)} \rade_{(q - 1)j}
\rade_{qj}, \chi_{qj}^{(\ell)} \rade_{qj} \right)$. It suffices to
show that
\begin{align}
    \forall j \in [m], \quad \measureQ_{1, j}^{(n)} | \Event_j^C =
    \measureQ_{-1, j}^{(n)} |
    \Event_j^C. \label{eq:generalized-birthday-equal-in-distr-factorized}
\end{align}
To prove
equation~\eqref{eq:generalized-birthday-equal-in-distr-factorized}, we
construct a distribution $\measureQ_{*, j}^{(n)}$, and show that it is
equal to both of the conditional laws above. In analogy to the proof
of Theorem~\ref{thm:linear-lower-bound}, we construct $\measureQ_{*,
  j}^{(n)}$ according to the following sampling procedure:

{\small{
    \bcar
    \item Sample the indicators $(\chi_{ij}^{(\ell)})_{0 \leq i \leq
      q, 1 \leq \ell \leq n}$, each from the Bernoulli distribution
      $\mathrm{Ber} (1/m)$, conditioned\footnote{Note that $\Prob
    (\Event_j^C) > 0$ in this sampling procedure. So the conditional
    distribution is well-defined.} on the event $\Event_j^C$.
    \item For each $i \in \{0, 1, \cdots, q\}$, sample a random bit
      $\zeta^{(i)} \sim \mathcal{U} (\{-1, 1\})$ independently.
    \item For each $\ell \in [n]$, generate the random object
    \begin{align*}
        \rho^{(\ell)} = \left(\chi_{0j}^{(\ell)}
      \zeta^{(0)}, \chi_{1j}^{(\ell)} \zeta^{(1)}, \cdots, \chi_{(q -
        1)j}^{(\ell)} \zeta^{(q - 1)}, \chi_{qj}^{(\ell)} \zeta^{(q)}
      \right).
    \end{align*}
    \ecar }}
In the following, we construct a coupling between
$\measureQ_{\plantedbit, j}^{(n)} | \Event_j^C$ and $\measureQ_{*,
  j}^{(n)}$, for any $\plantedbit \in \{-1, 1\}$, and show that they
are actually the same.

First, we couple the random indicators $(\chi_{ij}^{(\ell)})_{0 \leq i
  \leq q, 1 \leq \ell \leq n}$ under $\measureQ_{\plantedbit, j}^{(n)}
| \Event_j^C$ and $\measureQ_{*, j}^{(n)}$ directly so that they are
equal almost surely. By the first step in the sampling procedure, we
know that the conditional law of these indicators are the same under
both probability distributions.

By definition~\eqref{eq:bad-event-generalized-birthday}, we note that
\begin{align*}
    \Event_j^C = \left\{ \text{there exists } i \in \{0, 1, \cdots, q\},~\mbox{such
      that}~\chi_{ij}^{(\ell)} = 0 \text{ for all } \ell \in [n]\right\}.
\end{align*}
Let the random variable $\iota \in \{0, 1, \cdots, q\}$ be the
smallest such index\footnote{Note that under the joint distribution we
construct, $\iota$ is well-defined almost surely.} $i$. 
We construct the joint distribution by conditioning on different
values of $\iota$.  First, note that the value of the random variable
$\zeta^{(\iota)}$ is never observed, so we may set it to be an
independent Rademacher random variable without loss of generality, and
this does not affect the law of the random object
$(\rho^{(\ell)})_{\ell = 1}^n$ under consideration. Now consider the
following three cases:

\paragraph{Case I: $\iota = 0$:} In this case,
we have $\chi_{0j}^{(\ell)} = 0$ for all $\ell \in [n]$. So the first
coordinate of each $\rho^{(\ell)}$ is always zero.  We define the
following random variables in the probability space of
$(\rade_{ij})_{i = 1}^q$:
\begin{align*}
    {\zeta^{(q)}}' \mydefn \rade_{qj}, \quad\mbox{and}\quad
    {\zeta^{(i)}}' \mydefn \rade_{ij} \rade_{(i + 1) j} ~ \mbox{for}~
    i \in \{1,2, \cdots, q - 1\}.
\end{align*}
Since $(\rade_{ij})_{i = 1}^q$ are $\mathrm{i.i.d.}$ Rademacher random
variables, it is easy to show by induction that the random sequence
$\big({\zeta^{(i)}}'\big)_{i = 1}^q$ is also $\mathrm{i.i.d.}$
Rademacher, which has the same law as $({\zeta^{(i)}})_{i =
  1}^q$. Consequently, we can construct the coupling such that
$\big({\zeta^{(i)}}\big)_{i = 1}^q = \big({\zeta^{(i)}}'\big)_{i =
  1}^q$ almost surely. Under this coupling, when $\iota = 0$, the
random objects $(\rho^{(\ell)})_{\ell = 1}^n$ generated under both
probability distributions are almost-surely the same.

\paragraph{Case II: $\iota \in \{1, 2, \cdots, q - 1\}$:}
In this case, we have $\chi_{\iota j}^{(\ell)} = 0$ for any $\ell \in
[n]$.  We define the following random variables in the probability
space of $(\rade_{ij})_{i = 1}^q$:
\begin{align*}
    {\zeta^{(0)}}' \mydefn \plantedbit \rade_{1j}, \quad \mbox{and}\quad
    {\zeta^{(i)}}' \mydefn \rade_{ij} \rade_{(i + 1) j} ~ \mbox{for}~
    i \in \{1,2, \cdots, q - 1\} \setminus \{\iota\},
    \quad\mbox{and}\quad {\zeta^{(q)}}' \mydefn \rade_{qj}.
\end{align*}
Note that the tuple of random variables $({\zeta^{(i)}}')_{i =
  0}^{\iota - 1}$ lives in the sigma-field $\sigma (\rade_{1j},
\cdots, \rade_{\iota j})$, and $({\zeta^{(i)}}')_{i = \iota + 1}^q$,
on the other hand, lives in the sigma-field $\sigma (\rade_{(\iota +
  1)j}, \cdots, \rade_{q j})$, so these two tuples are
independent. For any fixed bit $\plantedbit \in \{-1, 1\}$, it is easy
to show by induction that $\big({\zeta^{(i)}}'\big)_{i = 0}^{\iota -
  1}$ is an $\mathrm{i.i.d.}$ Rademacher random sequence. Similarly,
applying the induction backwards from $q$ to $\iota + 1$, we can also
show that $\big({\zeta^{(i)}}'\big)_{i = \iota + 1}^q$ is also an
$\mathrm{i.i.d.}$ Rademacher random sequence. Putting together the
pieces, we see that the random variables $\big({\zeta^{(i)}}'\big)_{0
  \leq i \leq q, i \neq \iota}$ are $\mathrm{i.i.d.}$ Rademacher, and
have the same law as the tuple $\big(\zeta^{(i)}\big)_{0 \leq i \leq
  q, i \neq \iota}$. We can therefore construct the coupling such that
they are equal almost surely. Under this coupling, the random objects
$(\rho^{(\ell)})_{\ell = 1}^n$ generated under both probability
distributions are almost-surely the same.

\paragraph{Case III: $\iota = q$:}
In this case, we have $\chi_{\iota j}^{(\ell)} = 0$ for any $\ell \in
[n]$.  We define the following random variables in the probability
space of $(\rade_{ij})_{i = 1}^q$:
\begin{align*}
    {\zeta^{(0)}}' \mydefn \plantedbit \rade_{1j}, \quad \mbox{and}\quad
    {\zeta^{(i)}}' \mydefn \rade_{ij} \rade_{(i + 1) j} ~ \mbox{for}~
    i \in \{1,2, \cdots, q - 1\}.
\end{align*}
Note that $(\rade_{ij})_{i = 1}^q$ are $\mathrm{i.i.d.}$ Rademacher
random variables. For each choice of the bit $\plantedbit \in \{-1,
1\}$, we can show by induction that the random sequence
$\big({\rade^{(i)}}'\big)_{i = 1}^q$ is also $\mathrm{i.i.d.}$
Rademacher, which has the same law as the tuple $({\rade^{(i)}})_{i =
  1}^q$.  Making them equal almost surely in the coupling leads to the
corresponding random object $(\rho^{(\ell)})_{\ell = 1}^n$ being
almost-surely equal.

\medskip

Therefore, we have constructed a coupling between $\measureQ_{z,
  j}^{(n)} | \Event_j^C$ and $\measureQ_{*, j}^{(n)}$ so that the
generated random objects are always the equal, for any $\plantedbit
\in \{-1, +1\}$. This shows that
\begin{align*}
    \measureQ_{1, j}^{(n)} | \Event_j^C = \measureQ_{*, j}^{(n)} = \measureQ_{-1, j}^{(n)} |
    \Event_j^C,
\end{align*}
which completes the proof of
equation~\eqref{eq:generalized-birthday-equal-in-distr-factorized},
and hence, the lemma.


\section{Proof of the bounds on the approximation factor}

In this section, we prove the claims on the quantity $\alpha(M,
\conmax)$ that defines the optimal approximation factor.

\subsection{Proof of Lemma~\ref{lem:approx-factor-upper-bounds}}
\label{app:proof-of-approx-factor-upper-bounds}

Recall that
\begin{align}
  \prefact (\SpecMat, s) = 1 + \lambda_{\max} \left( (I -
  \SpecMat)^{- 1} ( s^2 I_d - \SpecMat \SpecMat^\top ) (I -
  \SpecMat)^{- \top} \right).
\end{align}
In the following, we prove upper bounds for the two different cases
separately.

\paragraph{Bounds in the general case:} By assumption, we have
$\opnorm{M} \leq s$, and consequently,
\begin{align*}
    0 \preceq s^2 I_d - MM^\top \preceq s^2.
\end{align*}
Thus, we have the sequence of implications
\begin{align*}
    \prefact (\SpecMat, s) - 1 & = \lambda_{\max} \left( (I -
    \SpecMat)^{- 1} ( s^2 I - \SpecMat \SpecMat^\top ) (I -
    \SpecMat)^{- \top} \right)\\ &= \opnorm{(I - \SpecMat)^{- 1} ( s^2
      I - \SpecMat \SpecMat^\top ) (I - \SpecMat)^{- \top}}\\ &\leq
    \opnorm{(I - \SpecMat)^{-1}} \cdot \opnorm{s^2 I_d - \SpecMat
      \SpecMat^\top } \cdot \opnorm{(I - \SpecMat)^{-1}}\\ &\leq
    \opnorm{(I - \SpecMat)^{-1}}^2 \cdot s^2,
\end{align*}
which proves the bound.

\paragraph{Bounds under non-expansive condition:} When $s \leq 1$, we have
\begin{align*}
    s^2 I - \SpecMat \SpecMat^\top \preceq I - \SpecMat \SpecMat^\top
    = \frac{1}{2} (I - \SpecMat) (I + M^\top) + \frac{1}{2} (I + M) (I
    - \SpecMat^\top).
\end{align*}
Consequently, we have the chain of bounds
\begin{align*}
    \prefact (\SpecMat, s) - 1 &\leq \lambda_{\max} \left( (I -
    \SpecMat)^{- 1} (I - \SpecMat \SpecMat^\top ) (I - \SpecMat)^{-
      \top} \right)\\ &= \frac{1}{2} \lambda_{\max} \left( (I +
    M)^\top (I - \SpecMat^\top)^{-1} + (I - \SpecMat)^{-1} (I + M)
    \right)\\ &\leq \frac{1}{2} \opnorm{(I + M)^\top (I -
      \SpecMat^\top)^{-1}} + \frac{1}{2} \opnorm{(I - \SpecMat)^{-1}
      (I + M) }\\ &\leq \opnorm{(I - \SpecMat)^{-1}} + \opnorm{(I -
      \SpecMat^{\top})^{-1}}\\ &= 2 \opnorm{(I - \SpecMat)^{-1}}.
\end{align*}
Finally, we note that if $\kappa (\SpecMat) < 1$, then for any $u \in \real^d$, we have
\begin{align*}
    (1 - \kappa (\SpecMat)) \vecnorm{u}{2}^2 \leq \inprod{(I -
    \SpecMat)u}{u} \leq \vecnorm{(I - \SpecMat) u}{2} \cdot
  \vecnorm{u}{2}.
\end{align*}
Consequently, we have $\opnorm{(I - \SpecMat)^{-1}} \leq \frac{1}{1 -
  \kappa (\SpecMat)}$, which completes the proof of this lemma.


\subsection{Proof of Lemma~\ref{lem:approx-factor-symmetric}}
\label{app:proof-of-approx-factor-sym}

Once again, recall the definition
\begin{align*}
    \prefact (\SpecMat, s) = 1 + \lambda_{\max} \left( (I -
    \SpecMat)^{- 1} ( s^2 I_d - MM^\top ) (I - \SpecMat)^{- \top}
    \right).
\end{align*}
Since $M$ is symmetric, let $\SpecMat = P \Lambda P^\top$ be its
eigen-decomposition, where $\Lambda = \mathrm{diag} (\lambda_1,
\lambda_2, \cdots, \lambda_d)$, and note that
\begin{align*}
    \prefact (\SpecMat, s) &= 1 + \lambda_{\max} \left( P (I -
    \Lambda)^{-1} (s^2 - \Lambda^2) (I - \Lambda)^{-1} P^\top
    \right)\\ &= 1 + \lambda_{\max} \left( (I - \Lambda)^{-2} (s^2 -
    \Lambda^2) \right)\\ &= 1 + \max_{1 \leq i \leq d} \left(
    \frac{\conmax^2 - \lambda_i^2}{(1 - \lambda_i)^2} \right),
\end{align*}
which completes the proof.



\section{Proofs for the examples}

In this section, we provide proofs for the results related to three
examples discussed in Section~\ref{SecConsequences}. Note that
Corollary~\ref{corr:oracle-ineq-linear-regression} follows directly
from Theorem~\ref{thm:linear-oracle-ineq}. Moreover, the proof of
Corollary~\ref{corr:elliptic} builds on some technical results, and so
we postpone the proof of all results related to elliptic equations to
Appendix~\ref{app:elliptic}. We begin this section with proofs of
results related to temporal difference methods, i.e.,
Corollary~\ref{corr:oracle-ineq-lstd-iid-discount} and
Proposition~\ref{prop:mrp-approx-factor-lb}.

\subsection{Proof of Corollary~\ref{corr:oracle-ineq-lstd-iid-discount}}
\label{App:proof-lstd-upper}

Recall our definition of the positive definite matrix $B$, with
$B_{ij} = \inprod{\psi_i}{\psi_j}$. Letting $\theta_t \mydefn
\Bmat^{1/2} \vartheta_t$, the
iterates~\eqref{eq:lsa-lstd-iid-discount} can be equivalently written
as
\begin{align}
  \theta_{t + 1} = \theta_t - \stepsize \Bmat \cdot \left( \phi (s_{t
    + 1}) \phi(s_{t + 1})^\top \theta_t - \discount \phi (s_{t + 1})
  \phi (s_{t + 1}^+)^\top \theta_t + R_{t + 1} (s_{t + 1}) \phi (s_{t
    + 1}) \right), \label{eq:lsa-lstd-in-orthogonal-coordinate}
\end{align}
and the Polyak--Ruppert averaged iterate is given by
$\thetahat_\numobs \mydefn \frac{2}{\numobs} \sum_{t =
  \numobs/2}^{\numobs - 1} \theta_t$. We also define $\thetabar
\mydefn \PhiOp \vbar$, which is the solution to projected linear
equations under the orthogonal basis. Clearly, we have $\thetabar =
\Bmat^{1/2} \bar{\vartheta}$.

We now claim that if $\numobs \geq \frac{c_0 \specpar^4 \smooth^2
}{\strongconvex^2 (1 - \kappa (\SpecMat))^2} d \log^2 \left(
\frac{\vecnorm{\vartheta_0 - \bar{\vartheta}}{2} d
  \smooth}{\strongconvex (1 - \kappa (\SpecMat))}\right)$, then
\begin{subequations}
\begin{align}
   \statnorm{\adjoint{\PhiOp} \bar{\theta} - \valuestar}^2 &\leq
   \prefact (\SpecMat, \discount) \AppErr (\LinSpace, \valuestar),
   \text{ and } \label{eq:lstd-orthogonal-basis-approx-err}\\ \Exs
   \vecnorm{\thetahat_\numobs - \bar{\theta}}{2}^2 &\leq c
   \EstErr_\numobs (\SpecMat, \Sigma_\Lmat + \Sigma_\bvec) + c \big(1
   + \statnorm{\valuebar}^2\big) \left( \frac{\specpar^2 \smooth}{(1 -
     \kappa (\SpecMat)) \strongconvex} \sqrt{\frac{\usedim}{\numobs}}
   \right)^3. \label{eq:lstd-orthogonal-basis-stat-error}
\end{align}
\end{subequations}
Taking both inequalities as given for now, we proceed with the proof
of this corollary. Combining
equation~\eqref{eq:lstd-orthogonal-basis-approx-err} and
equation~\eqref{eq:lstd-orthogonal-basis-stat-error} via Young's
inequality, we arrive at the bound
\begin{align*}
    \Exs \statnorm{\valuehat_n - \valuestar}^2 &\leq (1 + \offpar)
    \statnorm{\adjoint{\PhiOp} \bar{\theta} - \valuestar}^2 + \left(1
    + \frac{1}{\offpar} \right)\Exs \vecnorm{\thetahat_\numobs -
      \bar{\theta}}{2}^2\\ &\leq (1 + \offpar)\AppErr (\LinSpace,
    \valuestar) + c \left(1 + \frac{1}{\offpar} \right) \left[
      \EstErr_\numobs (\SpecMat, \Sigma_\Lmat + \Sigma_\bvec) + \big(1
      + \statnorm{\valuebar}^2\big) \left( \frac{\specpar^2 \smooth}{(1
        - \kappa (\SpecMat)) \strongconvex}
      \sqrt{\frac{\usedim}{\numobs}} \right)^3 \right],
\end{align*}
which completes the proof of this corollary.

\paragraph{Proof of equation~\eqref{eq:lstd-orthogonal-basis-approx-err}:}

By equation~\eqref{eq:lstd-in-low-dimensional-space} and the
definition of $\thetabar$, we have
\begin{align*}
    \thetabar = \discount \SpecMat \thetabar + \Exs_\stationary [R (s)
      \phi (s)].
\end{align*}
It is easy to see that $\adjoint{\PhiOp}\thetabar$ solves the
projected Bellman equation~\eqref{eq:projected-bellman-discount}. Note
furthermore that the projected linear operator is given by
\begin{align*}
    \PhiOp \Lmat \adjoint{\PhiOp} = \discount \PhiOp \transition \adjoint{\PhiOp} = \SpecMat.
\end{align*}
Invoking the bound in equation~\eqref{eq:linear-approx-factor}, we
complete the proof of this inequality.

\paragraph{Proof of equation~\eqref{eq:lstd-orthogonal-basis-stat-error}:}

Following the proof strategy for the
bound~\eqref{eq:linear-statistical-error}, we first show an upper
bound on the iterates $\Exs \statnorm{\vartheta_t -
  \bar{\vartheta}}^2$ under the non-orthogonal basis $(\psi_j)_{j \in
  [\usedim]}$, and then use this bound to establish the final
estimation error guarantee under $\statnorm{\cdot}$-norm.

Recall the stochastic approximation procedure under the non-orthogonal basis:
\begin{align*}
  \vartheta_{t + 1} = \vartheta_t - \stepsize \left( \psi (s_{t +
    1}) \psi (s_{t + 1})^\top \vartheta_t - \discount \psi (s_{t +
    1}) \psi (s_{t + 1}^+)^\top \vartheta_t - R_{t + 1} (s_{t + 1})
  \psi (s_{t + 1}) \right).
\end{align*}

Let $\widetilde{\SpecMat} \mydefn I_d - \frac{1}{\smooth} \Bmat^{1/2}
\left(I_d - \SpecMat \right) \Bmat^{1/2}$ and
$\widetilde{\projectedbvec} \mydefn \frac{1}{\smooth} \Exs [R (s) \psi
  (s)]$.  We can view equation~\eqref{eq:lsa-lstd-iid-discount} as a
stochastic approximation procedure for solving the linear fixed-point
equation $\bar{\vartheta} = \widetilde{\SpecMat} \bar{\vartheta} +
\widetilde{\projectedbvec}$, with stochastic observations
\begin{align*}
    \widetilde{\SpecMat}_t \mydefn I_d - \smooth^{-1} \left( \psi
    (s_t) \psi (s_t)^\top - \discount \psi (s_{t}) \psi (s_t^+)^\top
    \right), \quad \mbox{and} \quad \widetilde{\projectedbvec}_t
    \mydefn \smooth^{-1} R(s_t) \psi (s_t).
\end{align*}
To verify Assumption~\ref{assume-second-moment}, we note that for $p, q \in \sphere^{\usedim - 1}$, the following bounds directly follows from the condition~\eqref{eq:lstd-moment-assumption}:
\begin{align*}
    \Exs \left( p^\top \big( \widetilde{\SpecMat}_t -
    \widetilde{\SpecMat} \big)q\right)^2 &\leq 2 \smooth^{-2} \Exs
    \left( (p^\top \psi (s_t)) \cdot (\psi (s_t)^\top q) \right)^2 + 2
    \smooth^{-2} \Exs \left( (p^\top \psi (s_t)) \cdot (\psi
    (s_t^+)^\top q) \right)^2\\ &\leq 2 \smooth^{-2} \sqrt{\Exs \left(
      p^\top \psi (s_t) \right)^4 \cdot \Exs \left( \psi (s_t)^\top q
      \right)^4} + 2 \smooth^{-2} \sqrt{\Exs \left( p^\top \psi (s_t)
      \right)^4 \cdot \Exs \left( \psi (s_t^+)^\top q
      \right)^4}\\ &\leq \frac{4 \specpar^4}{\smooth^2}
    \vecnorm{\Bmat^{1/2} p}{2}^2 \cdot \vecnorm{\Bmat^{1/2} q}{2}^2
    \leq 4 \specpar^4,
\end{align*}
and
\begin{align*}
  \Exs \left( p^\top \big( \widetilde{\projectedbvec}_t -
  \widetilde{\projectedbvec} \big) \right)^2 &\leq \beta^{-2} \Exs
  \left( R (s_t) \cdot p^\top \psi (s_t) \right)^2\\ & \leq \beta^{-2}
  \sqrt{\Exs \left[ R (s_t)^4 \right] \cdot \Exs \left( p^\top
    \Bmat^{1/2} \phi (s_t) \right)^4} \leq \specpar^4 / \smooth.
\end{align*}
Consequently, for the stochastic approximation procedure in
equation~\eqref{eq:lsa-lstd-iid-discount},
Assumption~\ref{assume-second-moment} is satisfied with $\sigmaA = 2 \specpar^2$ and $\sigmab = \specpar^2 / \sqrt{\smooth}$.

To establish an upper bound on $\kappa (\widetilde{\SpecMat})$, we note that
\begin{align*}
  1 - \kappa (\widetilde{\SpecMat}) &= \frac{1}{\smooth}
  \lambda_{\min} \left( \Bmat - \Bmat^{1/2} \frac{\SpecMat +
    \SpecMat^\top}{2} \Bmat^{1/2} \right) \\ &= \frac{1}{\smooth}
  \inf_{u \in \sphere^{\usedim - 1}} (\Bmat^{1/2} u)^\top \left(
  I_\usedim - \frac{\SpecMat + \SpecMat^\top}{2} \right) (\Bmat^{1/2}
  u)\\ &\geq \frac{\strongconvex}{\smooth} \inf_{u \in
    \sphere^{\usedim - 1}} u^\top \left( I_\usedim - \frac{\SpecMat +
    \SpecMat^\top}{2} \right) u \geq \frac{\strongconvex}{\smooth}
  \left(1 - \kappa (\SpecMat) \right).
\end{align*}
Invoking Lemma~\ref{LemLinearSABound}, for $\stepsize < \frac{c_0 (1 -
  \kappa (\SpecMat)) \strongconvex}{ (\specpar^4 \usedim + 1)
  \smooth^2}$, we have
\begin{align}
\label{eq:lstd-iterate-bound-in-non-orthogonal-coordinate}  
\Exs \vecnorm{\vartheta_t - \bar{\vartheta}}{2}^2 \leq e^{-
  \frac{\strongconvex}{2} (1 - \kappa (\SpecMat)) \stepsize t} \Exs
\vecnorm{\vartheta_0 - \bar{\vartheta}}{2}^2 + \frac{8 \stepsize
  \smooth}{(1 - \kappa (\SpecMat)) \strongconvex} \left(
\vecnorm{\vartheta}{2}^2 \specpar^4 d + \specpar^4 d / \smooth \right).
\end{align}
On the other hand, applying Lemma~\ref{LemPrisonTime} to the
stochastic approximation
procedure~\eqref{eq:lsa-lstd-in-orthogonal-coordinate} under the
orthogonal coordinates, we have the bound
\begin{multline}
  \Exs \statnorm{\vhat_\numobs - \vbar}^2 \leq \frac{6}{\numobs -
    \numburn} \trace \left( (I - \SpecMat)^{-1} \SigStar (I -
  \SpecMat)^{- \top} \right) \\
    + \frac{6}{(\numobs - \numburn)^2} \sum_{t = \numburn}^\numobs \Exs \vecnorm{(I -
     \Bmat^{1/2} \widetilde{\SpecMat} \Bmat^{-1/2})^{-1} \Bmat^{1/2} (\widetilde{\SpecMat}_{t + 1} - \widetilde{\SpecMat}) \Bmat^{-1/2} (\theta_t -
      \thetabar)}{2}^2\\
      + \frac{3\Exs \vecnorm{\big( I_d - \Bmat^{1/2} \widetilde{\SpecMat} \Bmat^{-1/2}\big)^{-1} (\theta_\numobs -
        \theta_{\numburn})}{2}^2 }{\stepsize^2 \smooth^2 (\numobs - \numburn)^2 }.\label{eq:prison-in-lstd-with-non-orthogonal}
\end{multline}
Straightforward calculation yields
\begin{align*}
    \Exs \vecnorm{(I -
     \Bmat^{1/2} \widetilde{\SpecMat} \Bmat^{-1/2})^{-1} \Bmat^{1/2} (\widetilde{\SpecMat}_{t + 1} - \widetilde{\SpecMat}) \Bmat^{-1/2} (\theta_t -
      \thetabar)}{2}^2 = \smooth^2 \Exs \vecnorm{(I - \SpecMat)^{-1} \Bmat^{- 1/2}(\widetilde{\SpecMat}_{t + 1} - \widetilde{\SpecMat}) (\vartheta_t - \bar{\vartheta})}{2}^2.
\end{align*}
For any vector $p \in \real^d$, using condition~\eqref{eq:lstd-moment-assumption}, we note that
\begin{align*}
    \Exs \vecnorm{ \Bmat^{- 1/2}(\widetilde{\SpecMat}_{t} - \widetilde{\SpecMat}) p}{2}^2 &\leq 2 \smooth^{-2} \Exs \vecnorm{\phi (s_{t}) \phi (s_t)^\top \Bmat^{1/2} p}{2}^2 + 2 \smooth^{-2} \Exs \vecnorm{\phi (s_{t}) \phi (s_t^+)^\top \Bmat^{1/2} p}{2}^2\\
    &\leq 2 \smooth^{-2} \sqrt{\Exs \vecnorm{\phi (s_t)}{2}^4} \cdot \sqrt{\Exs \big( \phi (s_t)^\top \Bmat^{1/2} p \big)^4} + 2 \beta^{-2} \sqrt{\Exs \vecnorm{\phi (s_t)}{2}^4} \cdot \sqrt{\Exs \big( \phi (s_t^+)^\top \Bmat^{1/2} p \big)^4}\\
    &\leq 4 \smooth^{-1} \specpar^4 \usedim.
\end{align*}
Substituting into the identity above, we obtain
\begin{align*}
    \Exs \vecnorm{(I -
     \Bmat^{1/2} \widetilde{\SpecMat} \Bmat^{-1/2})^{-1} \Bmat^{1/2} (\widetilde{\SpecMat}_{t + 1} - \widetilde{\SpecMat}) \Bmat^{-1/2} (\theta_t -
      \thetabar)}{2}^2  \leq \frac{4 \smooth \specpar^4 \usedim}{\big( 1 - \kappa (\SpecMat) \big)^2} \Exs \vecnorm{\vartheta_t - \bar{\vartheta}}{2}^2.
\end{align*}
For the third term in equation~\eqref{eq:prison-in-lstd-with-non-orthogonal}, we note that
\begin{align*}
    \Exs \vecnorm{\big( I_d - \Bmat^{1/2} \widetilde{\SpecMat} \Bmat^{-1/2}\big)^{-1} (\theta_\numobs -
        \theta_{\numburn})}{2}^2 &= \smooth^2 \Exs \vecnorm{(I - \SpecMat)^{-1} \Bmat^{- 1/2} (\vartheta_\numobs - \vartheta_\numburn)}{2}^2 \\
        &\leq \frac{2\smooth^2}{\strongconvex \big(1 - \kappa (\SpecMat) \big)^2} \left( \Exs \vecnorm{\vartheta_\numobs - \bar{\vartheta}}{2}^2 + \Exs \vecnorm{\vartheta_\numburn - \bar{\vartheta}}{2}^2 \right).
\end{align*}
Putting together the pieces and invoking the bound~\eqref{eq:lstd-iterate-bound-in-non-orthogonal-coordinate}, we see that if $\numburn \geq c_0 \frac{1}{\strongconvex \stepsize (1 - \kappa)} \log \big( \frac{\usedim \smooth}{\strongconvex (1 - \kappa)} \big)$, then
\begin{align*}
     \Exs \statnorm{\vhat_\numobs - \vbar}^2 &\leq 6 \BigEstErr + \left[ \frac{24 \smooth \specpar^4 \usedim}{\big( 1 - \kappa (\SpecMat) \big)^2 \numobs} + \frac{48 }{\strongconvex \big(1 - \kappa (\SpecMat) \big)^2 \stepsize^2  \numobs^2} \right] \cdot \sup_{\numburn \leq t \leq \numobs} \Exs \vecnorm{\vartheta_t - \bar{\vartheta}}{2}^2\\
     &\leq  6 \BigEstErr + c \frac{\smooth^3}{\strongconvex^2 \big(1 - \kappa (\SpecMat) \big)^3} \left[ \frac{\specpar^4 \stepsize \usedim}{\numobs} + \frac{1}{\stepsize \smooth^2 \numobs^2} \right] \left( \vecnorm{ \bar{\vartheta} }{2}^2 \specpar^4 d + \specpar^4 d / \smooth \right).
\end{align*}
 Now note that $\vecnorm{\bar{\vartheta}}{2}^2 = \vecnorm{\Bmat^{-1/2} \bar{\theta}}{2}^2 \leq \strongconvex^{-1} \statnorm{\vbar}^2$, and so choosing the step size $\stepsize \mydefn \frac{1}{c_0 \specpar^2 \smooth \sqrt{\usedim \numobs}}$ yields
 \begin{align*}
      \Exs \statnorm{\vhat_\numobs - \vbar}^2 \leq 6 \BigEstErr + c \frac{\smooth^3 \specpar^6}{\strongconvex^3 \big(1 - \kappa (\SpecMat) \big)^3} \left( \frac{d}{n} \right)^{3/2}.
 \end{align*}
This completes the proof of
equation~\eqref{eq:lstd-orthogonal-basis-stat-error}, and thus the
corollary.


\subsection{Proof of Proposition~\ref{prop:mrp-approx-factor-lb}}\label{App:proof-lstd-lower}

Our construction and proof is inspired by the proof of Theorem~\ref{thm:linear-lower-bound},
with some crucial differences in the analysis that result from the specific noise model
in the MRP setting.
Letting $D$ and $d$ be integer multiples of four without loss of
generality, we denote the state space by $\statespace = \{1, 2, \cdots,
D\}$.  We decompose the state space into $\statespace = \statespace_0
\cup \statespace_1 \cup \statespace_2$, with $\statespace_0 \mydefn
\{1,2, \cdots, 2d\}$, $\statespace_1 \mydefn \{2d + 1, \cdots, d +
\frac{D}{2}\}$, and $\statespace_2 \mydefn \{d + \frac{D}{2} + 1,
\cdots, D\}$. Define the scalars $\rho = \min (\discount, \kappamrp)
\in (0, 1)$ and $\tau \mydefn \frac{\oracleErr}{\sqrt{2 (1 - \rho)}} \wedge 1$.

\begin{figure}[htb]
\begin{center}
    \widgraph{0.9\linewidth}{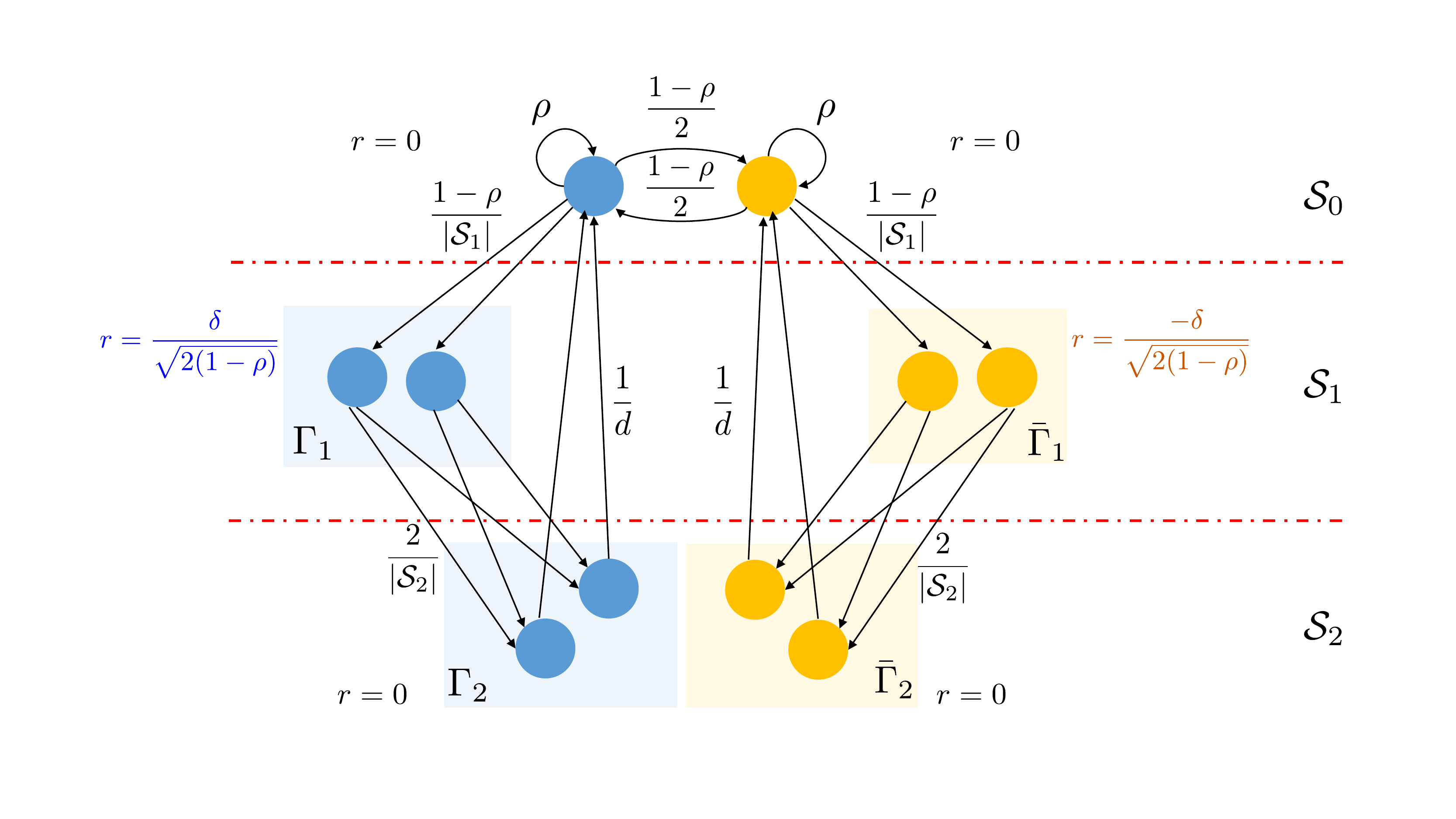}
    \caption{A graphical illustration of the MRP instance constructed
      above. For this instance, we let $d = 1$, $|\statespace_1| = 4$
      and $|\statespace_2| = 4$, so that the total number of states is
      $\bigdim = 10$. In the graph, solid rounds stand for states, and
      arrows stand for the possible transitions. The numbers
      associated to the arrows stand for the probability of the
      transitions, and the equations $r = \cdots$ standard for the
      reward at a state. The sets $\statespace_0$, $\statespace_1$ and
      $\statespace_2$ are separated by red dotted lines, and the sets
      $\Gamma_1$, $\bar{\Gamma}_1$, $\Gamma_2$, and $\bar{\Gamma}_2$
      are marked by transparent rectangles. A blue round stands for a
      state with positive value function, and an orange round stands
      for a state with negative value function.}
    \label{fig:mdp-illustration}
\end{center}
\end{figure}

Given a sign $z \in \{-1, 1\}$ and subsets $\Gamma_1
\subseteq \statespace_1$ and $\Gamma_2 \subseteq \statespace_2$ such that
$|\Gamma_i| = \frac{1}{2} |\statespace_i|$ for each $i \in \{1, 2\}$, we let 
$\bar{\Gamma}_i \mydefn \statespace_i \setminus \Gamma_i$ for $i \in \{1,2\}$. We then
construct Markov reward processes
$(\transition^{(\Gamma_1, \Gamma_2, z)}, \reward^{(\Gamma_1, \Gamma_2,
  z)})$ and feature vectors $(\psi^{(\Gamma_1, \Gamma_2,
  z)} (s_i))_{i = 1}^D$, indexed by the tuple $(\Gamma_1, \Gamma_2, z)$. Entry $(i, j)$
  of the transition matrix is given by
\begin{subequations}
\begin{align}
    \transition^{(\Gamma_1, \Gamma_2, z)} (i, j)
    &\mydefn \begin{cases} \rho & i = j \in \statespace_0,\\ \frac{1 -
        \rho}{2} & i, j \in \statespace_0,~ |i - j| = d,\\ \frac{1 -
        \rho}{|\statespace_1|} & (i, j) \in \left( \{1,\cdots, d\}
      \times \Gamma_1 \right) \cup \left( \{d + 1,\cdots, 2d\} \times
      \bar{\Gamma}_1 \right),\\ \frac{2}{|\statespace_2|} & (i, j) \in
      \left( \Gamma_1 \times \Gamma_2 \right) \cup \left(
      \bar{\Gamma}_1 \times \bar{\Gamma}_2 \right),\\ \frac{1}{d} &
      (i, j) \in \left( \Gamma_2 \times \{1,2,\cdots, d\} \right) \cup
      \left( \bar{\Gamma}_2 \times \{d + 1,\cdots, 2 d\} \right)\\ 0 &
      \mbox{otherwise}.    
      \end{cases} \label{eq:mrp-approx-factor-lb-transition}
      \end{align}
      The reward function at state $i$ is given by
      \begin{align}
    \reward^{(\Gamma_1, \Gamma_2, z)} (i) &\mydefn \begin{cases}
    z\tau & i \in \Gamma_1,\\
    -z\tau & i \in \bar{\Gamma}_1,\\
    0 & \mbox{otherwise}.
    \end{cases} \label{eq:mrp-approx-factor-lb-reward}
    \end{align}
    This MRP is illustrated in Figure~\ref{fig:mdp-illustration} for convenience.
    It remains to specify the feature vectors, and we use the same set of features for each tuple
    $(\Gamma_1, \Gamma_2, z)$. The $i$-th such feature vector is given by
    \begin{align}
    \psi (i) & \mydefn \begin{cases}
    \sqrt{\frac{3 - \rho}{2} \usedim} e_i & i \in \{1,2, \cdots, d\},\\
    -\sqrt{\frac{3 - \rho}{2} \usedim} e_{i - d} & i \in \{d + 1, \cdots, 2d\},\\
    0 & \mbox{otherwise}.
    \end{cases}\label{eq:mrp-approx-factor-lb-feature}
\end{align}
\end{subequations}
It is easy to see that for any tuple $(z, \Gamma_1, \Gamma_2)$, the
Markov chain is irreducible and aperiodic, and furthermore, that the stationary distribution
of the transition kernel $\transition^{(\Gamma_1, \Gamma_2, z)}$ is independent of the tuple
$(\Gamma_1, \Gamma_2, z)$, and given by
\begin{align*}
    \stationary = \left[ \begin{matrix} \undermat{2d}{\frac{1}{(3 -
            \rho)d}& \cdots &\frac{1}{(3 - \rho)d}}& \undermat{D -
          2d}{\frac{1 - \rho}{ (3 - \rho) (D - 2d)} & \cdots & \frac{1
            - \rho}{ (3 - \rho) (D - 2d)}}
    \end{matrix} \right].
\end{align*}
\vspace{0.1cm}

Clearly, we have $\Exs_\stationary [\psi (s) \psi (s)^\top] = I_d$
under the stationary distribution. For the projected transition
kernel, we have
\begin{align}
    \Exs [\psi (s) \psi (s^+)^\top] = \frac{3 - \rho}{2} \cdot
    \left(\rho - \frac{1 - \rho}{2} \right) I_d \preceq \rho I_d
    \preceq \kappamrp I_d. \label{eq:cov-ineq-in-mrp-construction}
\end{align}
Given the discount factor $\discount \in (0, 1)$, let $c_0 \mydefn
\frac{(1 - \rho) / 2}{1 - \discount (\rho - (1 - \rho) (1 -
  \discount^2)/2)}$ for convenience. Straightforward calculation then yields 
  that the value function for
the problem instance $\left( \transition^{(\Gamma_1, \Gamma_2, z)},
\reward^{(\Gamma_1, \Gamma_2, z)} \right)$ at state $i$ is given by
\begin{align*}
    \valuestar_{\Gamma_1, \Gamma_2, z} (i) = \begin{cases} c_0 z \tau&
      i \in \{1,2,\cdots, d\},\\ - c_0 z \tau& i \in \{d + 1,\cdots,
      2d\},\\ (1 + \discount^2 c_0) z \tau & i \in \Gamma_1,\\ - (1 +
      \discount^2 c_0) z \tau& i \in \bar{\Gamma}_1,\\ \discount c_0 z
      \tau& i \in \Gamma_2,\\ - \discount c_0 z\tau & i \in
      \bar{\Gamma}_2.
    \end{cases} 
\end{align*}
For $\rho > 1/2$, we have the bounds
\begin{align*}
    c_0 \geq \frac{1}{4} \cdot \frac{1 - \rho}{1 - \discount \rho}
    \geq \frac{1 - \rho}{4 (1 - \rho^2)} \geq \frac{1}{8},\quad
    \mbox{and}\quad c_0 \leq \frac{1 - \rho}{1 - \discount \rho} \leq
    1.
\end{align*}
Consequently, we have $|\valuestar_{\Gamma_1, \Gamma_2, z} (i)| \asymp |\valuestar_{\Gamma_1, \Gamma_2, z} (j)|$
for each pair $(i, j)$.

Note that by our construction, the subspace $\lspace$ spanned by the
basis functions $\psi (1), \psi(2), \cdots, \psi (2d)$ is given by
\begin{align*}
    \lspace = \left\{\valuefunc \in \Ltwospace (\statespace,
    \stationary): \valuefunc (s) = 0 ~ \mbox{for}~ s
    \notin \statespace_0, ~\mbox{and}~ \valuefunc (i + d) = -
    \valuefunc (i) \text{ for all } i \in [d]\right\}.
\end{align*}
Consequently, we have
\begin{align}
    \inf_{\valuefunc \in \lspace} \statnorm{\valuefunc -
      \valuestar_{\Gamma_1, \Gamma_2, z}}^2 = \frac{1 - \rho}{3 -
      \rho} \cdot \left( \frac{1}{2} (1 + \discount^2 c_0)^2 \tau^2 +
    \frac{1}{2} \discount^2 c_0^2 \tau^2 \right) \leq 2 (1 - \rho)
    \tau^2 = \oracleErr^2. \label{eq:oracle-error-in-mrp-construction}
\end{align}
Putting the equations~\eqref{eq:cov-ineq-in-mrp-construction}
and~\eqref{eq:oracle-error-in-mrp-construction} together, for any
tuple $(\Gamma_1, \Gamma_2, z)$, we conclude that the problem instance
$(\transition^{(\Gamma_1, \Gamma_2, z)}, \reward^{(\Gamma_1, \Gamma_2,
  z)}, \discount, \psi^{((\Gamma_1, \Gamma_2, z))})$ belongs to the
class $\classMRP (\kappamrp, \discount, \bigdim, \oracleErr)$.

In order to apply Le Cam's lemma, we define the following mixture
distributions for each $z \in \{-1, 1\}$:
\begin{align*}
    \ProbInst_z^{(n)} \mydefn
    \binom{|\statespace_1|}{|\statespace_1|/2}^{-2}
    \sum_{\stackrel{\Gamma_1 \subseteq \statespace_1, \Gamma_2
        \subseteq \statespace_2}{|\Gamma_1| = |\Gamma_2| = \frac{1}{2}
        |\statespace_1|}} \ProbInst_{\Gamma_1, \Gamma_2, z}^{\otimes
      n},
\end{align*}
where $\ProbInst_{\Gamma_1, \Gamma_2, z}$ is the law of an observed
tuple $(s_i, s_i^+, \reward (s_i))$ under the MRP $\big(
\transition^{(\Gamma_1, \Gamma_2, z)}, \discount, \reward^{(\Gamma_1,
  \Gamma_2, z)} \big)$, and $\ProbInst_{\Gamma_1, \Gamma_2,
  z}^{\otimes n}$ denotes its $n$-fold product.
Our next result gives a bound on the total variation distance.
\begin{lemma}\label{lemma:tv-bound-in-mrp-lb}
    Under the set-up above, we have
$
        \totalvariation \left(\ProbInst_1^{(n)}, \ProbInst_{-1}^{(n)}
        \right) \leq \frac{C n^2}{D - 2 d}.
$
\end{lemma}
Taking this lemma as given, we now turn to the proof of the
proposition. Consider any estimator $\valuehat$ for the value
function. For any pair $\Gamma_1, \Gamma_2$ and $\Gamma_1',
\Gamma_2'$, we have
\begin{align*}
    \statnorm{\valuehat - \valuestar_{\Gamma_1, \Gamma_2, 1}}^2 +
    \statnorm{\valuehat - \valuestar_{\Gamma_1', \Gamma_2', -1}}^2
    \geq \frac{1}{2} \statnorm{\valuestar_{\Gamma_1, \Gamma_2, 1} -
      \valuestar_{\Gamma_1', \Gamma_2', -1}}^2 \geq \frac{1}{2} c_0^2
    \tau^2 \geq \frac{\oracleErr^2}{64(1 - \rho)}.
\end{align*}
Invoking Le Cam's lemma, for $D > 2 C (n^2 + d)$, we have
\begin{align*}
    \inf_{\valuehat_n} \sup_{(\transition, \discount, \reward, \psi)
      \in \classMRP} \geq \frac{c}{1 - \rho} \oracleErr^2 \left(1 -
    \totalvariation(\ProbInst_1^{(n)}, \ProbInst_{-1}^{(n)}) \right)
    \geq \frac{c'}{1 - \kappamrp \discount} \oracleErr^2,
\end{align*}
which completes the proof.
\qed

\subsubsection{Proof of Lemma~\ref{lemma:tv-bound-in-mrp-lb}}

In contrast to the proof of Theorem~\ref{thm:linear-lower-bound}, the underlying mixing components 
here are indexed by
random subsets of a given size, instead of random bits. This sampling procedure
introduces additional dependency, so that the
arguments in the proof of Theorem~\ref{thm:linear-lower-bound} do not directly
apply. Instead, we use an induction-type argument by constructing the
coupling directly.

Similarly to before, we construct a probability distribution $\measureQ^{(n)}$ and bound
the total variation distance between $\measureQ^{(n)}$ and
$\ProbInst^{(n)}_z $ for each $z \in \{-1, 1\}$. In particular, for $k \in [\numobs]$, we let
$\measureQ^{(k)}$ be the law of $k$ independent samples drawn from the following observation model:
{\small{ \bcar
    \item (Initial state:) Generate the state $s_i \sim \stationary$.
    \item (Next state:) If $s_i \in \statespace_1$, then generate $s_i^+ \sim \mathcal{U}
    (\statespace_2)$. If $s_i \in \statespace_2$, then generate $s_i^+ \sim
    \mathcal{U} (\statespace_0)$. On the other hand, if $s_i \in \statespace_0$, 
    then generate $S \sim \mathcal{U} (\statespace_1)$ and let\footnote{The expression $a \mod b$ denotes the remainder of $a$ divided by $b$, when $a$ and $b$ are integers.}
    \begin{align}
        s_i^+ =
        \begin{cases} s_i & \mathrm{w.p.} ~\rho,\\ 
        (s_i +
          d)\mod 2d & \mathrm{w.p.}~ \frac{1 - \rho}{2},\\ 
          S & \mathrm{w.p.}~ \frac{1 -
            \rho}{2}.
        \end{cases} \label{eq:mrp-construct-next-step-conditional-law-s0}
    \end{align}
    
    \item (Reward:) If $s_i \in \statespace_1$, randomly draw $R_i = \zeta^{(i)}
      \sim \mathcal{U} (\{-1, 1\})$, and output $\zeta^{(i)} \tau$ as
      the reward. Otherwise, output the reward $R_i = 0$.
      \ecar
}}

 To bound the total variation distance $\totalvariation
 (\measureQ^{(\numobs)}, \ProbInst_z^{(\numobs)})$, we use the
 following recursive relation, which holds for each $k = 0,1,\cdots, n - 1$:
\begin{multline} \label{eq:recursion-over-k}
    \totalvariation \left(\measureQ^{(k + 1)}, \ProbInst_z^{(k + 1)} \right) \leq
    \totalvariation \left(\measureQ^{(k)}, \ProbInst_z^{(k)} \right)\\ \quad \quad +
    \sup_{(s_i, s_i^+, R_i)_{i = 1}^k} \totalvariation \left(\measureQ^{(k +
      1)} | (s_i, s_i^+, R_i)_{i = 1}^k, \ProbInst_z^{(k + 1)} |(s_i, s_i^+,
    R_i)_{i = 1}^k \right).
\end{multline}
Owing to the i.i.d. nature of the sampling model for $\measureQ^{(k+1)}$, note that we have the equivalence $(s_{k + 1}, s_{k + 1}^+, R_{k + 1}) | (s_i, s_i^+, R_i)_{i = 1}^k \overset{d}{=} (s_{k + 1}, s_{k + 1}^+, R_{k + 1})$.

At this juncture, it is helpful to view the probability distributions $\ProbInst^{(k)}_1$ and $\ProbInst^{(k)}_{-1}$ via the following two-step sampling procedure: First, for $j \in \{1, 2\}$, sample the subsets $\Gamma_j \subseteq \statespace_j$ uniformly at random from the collection of all subsets of size $|\statespace_j| / 2$. Then, generate $k$ $\mathrm{i.i.d.}$ samples $(s_i, s_i^+, R_i)_{i = 1}^k$ according to the observation model~\eqref{eq:mrp-approx-factor-lb-transition}-\eqref{eq:mrp-approx-factor-lb-reward}. Consequently, for the rest of this proof, we view $\Gamma_1$ and $\Gamma_2$ as \emph{random} sets.
%
%
With this equivalence at hand, the following technical lemma 
shows that the posterior distribution of the subsets $(\Gamma_1, \Gamma_2)$ conditioned on sampling the tuple
$(s_i, s_i^+, R_i)_{i = 1}^k$ is very close to the distribution of subsets chosen uniformly at random.

\begin{lemma}\label{lemma:technicality-in-mrp-lower-bound-proof}
There is a universal positive constant $c$ such that for each bit $z \in \{\pm 1\}$ and indices $j \in \{1, 2\}$ and $k \in [n]$, the following statement is true almost surely. For each tuple $(s_i, s_i^+, R_i)_{i = 1}^k$ in the support of $\ProbInst_{z}^{(k)}$, the posterior distribution of $\Gamma_j$ conditioned on $(s_i, s_i^+, R_i)_{i = 1}^k \sim \ProbInst_{z}^{(k)}$ satisfies
    \begin{align*}
    \max_{s \in \statespace_j \setminus \cup_{i = 1}^k \{s_i, s_i^+\} } \abss{ \Prob \left( \Gamma_j \ni s  \mid  (s_i, s_i^+, R_i)_{i = 1}^k  \right) - \frac{1}{2}} &\leq \frac{c k}{\bigdim - \usedim}. 
    \end{align*}
\end{lemma}
In words, for any ``observable'' tuple $(s_i, s_i^+, R_i)_{i = 1}^k$ and each state $s \in \statespace_j \setminus \cup_{i = 1}^k \{s_i, s_i^+\}$, the posterior probability of the event $\{\Gamma_j \ni s\}$ conditioned on observing the tuple $(s_i, s_i^+, R_i)_{i = 1}^k$ is close to $1/2$ provided $\bigdim - \usedim$ is large relative to $k$.
%
In addition to the sets $\Gamma_j, j = 1, 2$ being close to uniformly random, we also require the following analog of a ``birthday-paradox'' argument in this setting. For convenience, we let $\mathcal{T}_k \defn \bigcup_{i = 1}^k \{s_i, s_i^+\}$ denote the subset of states seen up until sample $k$.

\begin{lemma} \label{lem:one-step-birthday-argument-in-mrp-lb-proof}
There is a universal positive constant $c$ such that for each $k \in [n]$ and each distribution $\mathbb{M}^{(k+1)} \in \left\{ \mathbb{P}_{-1}^{(k+1)}, \mathbb{P}_{-1}^{(k+1)}, \measureQ^{(k + 1)} \right\}$, the following statement holds almost surely. For each tuple $(s_i, s_i^+, R_i)_{i = 1}^{k+1}$ in the support of $\mathbb{M}^{(k+1)}$, the probability the tuple of states $\left\{s_{k + 1}, s_{k + 1}^+ \right\}$ conditioned on $(s_i, s_i^+, R_i)_{i = 1}^k \sim \mathbb{M}^{(k)}$ satisfies
\begin{align}
    \Prob \Big( \underbrace{\left\{s_{k + 1}, s_{k + 1}^+ \right\} \cap \mathcal{T}_k \cap (\statespace_1 \cup \statespace_2)\neq \varnothing}_{\mydefn \Event_{k + 1}^{(1)}} \mid  (s_i, s_i^+, R_i)_{i = 1}^k \Big) \leq \frac{c k}{\bigdim - \usedim}. 
\end{align}
\end{lemma}
In words, Lemma~\ref{lem:one-step-birthday-argument-in-mrp-lb-proof} ensures that if $D - d$ is large relative to $k$, then the states seen in sample $k + 1$ are different from those seen up until that point (provided we only count states in the set $\statespace_1 \cup \statespace_2$).
 Lemmas~\ref{lemma:technicality-in-mrp-lower-bound-proof} and~\ref{lem:one-step-birthday-argument-in-mrp-lb-proof} are both proved at the end of this section; we take them as given for the rest of this proof.

Now consider tuples $(s_{k + 1}, s_{k + 1}^+, R_{k + 1}) \sim
\ProbInst_z^{(k + 1)} |(s_i, s_i^+, R_i)_{i = 1}^k $ and $(\widetilde{s}_{k + 1},
\widetilde{s}_{k + 1}^+, \widetilde{R}_{k + 1}) \sim \measureQ^{(k + 1)} | (s_i,
s_i^+, R_i)_{i = 1}^k$; we will now construct  a coupling between these two
tuples in order to show that the total variation between between the respective laws is small.
%
First, note that under both $\ProbInst_z^{(k + 1)}$ and $\measureQ^{(k + 1)}$, the initial
state is drawn from the stationary distribution, i.e., $s_{k + 1}, \widetilde{s}_{k + 1}\sim \stationary$, regardless of
the sequence $(s_i, s_i^+, R_i)_{i = 1}^k$. We can therefore couple
the two conditional laws together so that $s_{k + 1} = \widetilde{s}_{k +
  1}$ almost surely. To construct the coupling for the rest, we consider the following three
cases:

\paragraph{Coupling on the event $s_{k + 1} \in \statespace_0$:}

We begin by coupling the reward random variables; we have $R_{k + 1} = \widetilde{R}_{k + 1} = 0$ under both conditional distributions, so this component of the distribution can be coupled trivially. Next, we couple the next state: By construction of the observation models~\eqref{eq:mrp-approx-factor-lb-transition} and~\eqref{eq:mrp-construct-next-step-conditional-law-s0}, we have
\begin{align*}
     \Prob \big(s_{k + 1}^+ = s_{k + 1} | s_{k + 1} \big) &= \Prob \big(\widetilde{s}_{k + 1}^+ = \widetilde{s}_{k + 1} | \widetilde{s}_{k + 1} \big) = \rho, \quad \mbox{and}\\
     \Prob \big(s_{k + 1}^+ = s_{k + 1} + d \mod 2d \mid s_{k + 1} \big)& = \Prob \big(\widetilde{s}_{k + 1}^+ = \widetilde{s}_{k + 1} + d \mod 2d \mid \widetilde{s}_{k + 1} \big) = \frac{1 - \rho}{2},
\end{align*}
and so these two components of the distribution can be coupled trivially.
It remains to handle the case where $s_{k + 1} \in \statespace_0$ and $s_{k + 1}^+ \in \statespace_1$. By the symmetry of elements within set $\statespace_1$, we note that on the event $\big(\Event_{k + 1}^{(1)} \big)^C$, both random variables $\widetilde{s}_{k + 1}^+$ and $s_{k + 1}^+$ are uniformly distributed on the set $\statespace_1 \setminus \mathcal{T}_k$. Consequently, on the event $\big(\Event_{k + 1}^{(1)} \big)^C$, we can couple the conditional laws so that $s_{k + 1}^+ = \widetilde{s}_{k + 1}^+$ almost surely. 
    
\paragraph{Coupling on the event $s_{k + 1} \in \statespace_1$:} As before, we begin by coupling the rewards, but first, note that on the event $\big(\Event_{k + 1}^{(1)}\big)^C$, we have $s_{k + 1} \in \statespace_1 \setminus \mathcal{T}_k$.
    Invoking Lemma~\ref{lemma:technicality-in-mrp-lower-bound-proof}, under $\ProbInst_z^{(k)}$ and conditionally on the value of $s_{k + 1}$, we have the bound
    \begin{align*}
        \abss{\Prob \left( s_{k + 1} \in \Gamma_1 \mid (s_i, s_i^+, R_i)_{i = 1}^k \right) - \frac{1}{2}} \leq \frac{c k}{\bigdim - \usedim}.
    \end{align*}
    Now the reward function~\eqref{eq:mrp-approx-factor-lb-reward} satisfies $r (s) = z \tau$ for $s \in \Gamma_1$ and $r (s) = - z\tau$ for $s \in \bar{\Gamma}_1$. On the other hand, under $\measureQ^{(k + 1)}$, the reward $\widetilde{R}_{k + 1}$ takes value of $\tau$ and $- \tau$, each with probability half. Consequently, there exists a coupling between $R_{k + 1}$ and $\widetilde{R}_{k + 1}$, such that
    \begin{align*}
        \Prob \Big( \underbrace{ R_{k + 1} \neq \widetilde{R}_{k + 1},~ s_{k + 1} \in \statespace_1 }_{\mydefn \Event^{(2)}_{k + 1}} \mid  (s_i, s_i^+, R_i)_{i = 1}^k  \Big) \leq \frac{c k}{\bigdim - \usedim}.
    \end{align*}
    Next, we construct the coupling for next-step transition conditionally on the current step. 
    By the symmetry of elements within set $\statespace_2$, we note that under $\big(\Event_{k + 1}^{(1)} \big)^C$, both random variables $\widetilde{s}_{k + 1}^+$ and $s_{k + 1}^+$ are uniformly distributed on the set $\statespace_2 \setminus \mathcal{T}_k$. Consequently, on the event $\big(\Event_{k + 1}^{(1)} \big)^C$, we can couple the conditional laws so
    that $s_{k + 1}^+ = \widetilde{s}_{k + 1}^+$ almost surely.

    \paragraph{Coupling on the event $s_{k + 1} \in \statespace_2$:}

 In this case, we have $R_{k + 1} = \widetilde{R}_{k + 1} = 0$ under both
 conditional distributions, so this coupling is once again trivial. It remains to construct a coupling between next-step transitions $s_{k + 1}^+$ and $\widetilde{s}_{k + 1}^+$. On the event $\big(\Event_{k + 1}^{(1)}\big)^C$, we have $s_{k + 1} \in \statespace_2 \setminus \mathcal{T}_k$. Under $\ProbInst_z^{(k)}$ and conditionally on the value of $s_{k + 1}$, Lemma~\ref{lemma:technicality-in-mrp-lower-bound-proof} leads to the bound
    \begin{align*}
        \abss{\Prob \left( s_{k + 1} \in \Gamma_2 \mid (s_i, s_i^+, R_i)_{i = 1}^k \right) - \frac{1}{2}} \leq \frac{c k}{\bigdim - \usedim}.
    \end{align*}
 
    By definition, under $\ProbInst_z^{(n)}$, we have
    that $s_{k + 1}^+ \sim \mathcal{U} (\{1,2,\cdots, d\})$ when $s_{k
      + 1} \in \Gamma_2$, and $s_{k + 1}^+ \sim \mathcal{U} (\{d +
    1,\cdots, 2d\})$ when $s_{k + 1} \in \bar{\Gamma}_2$. Under
    $\measureQ^{(n)}$, we have $\widetilde{s}_{k + 1}^+ \sim \mathcal{U}
    (\{1,2,\cdots, 2d\})$. Consequently, there exists a coupling such
    that
    \begin{align*}
      \Prob \Big( \underbrace{s_{k + 1}^+ \neq \widetilde{s}_{k +1}^+, ~ s_{k + 1} \in \statespace_2 }_{\mydefn \Event_{k + 1}^{(3)}} \mid  (s_i, s_i^+, R_i)_{i = 1}^k  \Big) \leq \frac{c k}{D - d}.
    \end{align*}
\medskip

    Putting together our bounds from the three cases, note that for any sequence $(s_i, s_i^+,
    R_i)_{i = 1}^k$ on the support of $\measureQ^{(k)}$ and $\ProbInst_z^{(k)}$, we almost surely
    have
    \begin{align*}
        &\totalvariation \left( \law \bigg[(s_{k + 1}, s_{k + 1}^+, R_{k + 1}) \; \big| \;
      (s_i, s_i^+, R_i)_{i = 1}^k\bigg], \law \bigg[ (\widetilde{s}_{k + 1}, \widetilde{s}_{k +
        1}^+, \widetilde{R}_{k + 1}) \; \big| \; (s_i, s_i^+, R_i)_{i = 1}^k\bigg]
      \right)\\ &\leq \sum_{j = 1}^3 \Prob \left( \Event_{k + 1}^{(j)} \mid  (s_i, s_i^+, R_i)_{i = 1}^k 
      \right) \leq \frac{c'k}{D - d},
    \end{align*}
    where the final inequality follows from applying Lemma~\ref{lem:one-step-birthday-argument-in-mrp-lb-proof}.
    Substituting into the recursion~\eqref{eq:recursion-over-k}, we conclude that for any $z \in \{-1, 1\}$, we have
    \begin{align*}
        \totalvariation \left( \measureQ^{(n)} , \ProbInst_z^{(n)} \right) \leq
        \sum_{k = 0}^{n -1} \sum_{j = 1}^3 \sup_{(s_i, s_i^+, R_i)_{i = 1}^k} \Prob \left( \Event_{k +
          1}^{(j)} \mid (s_i, s_i^+, R_i)_{i = 1}^k \right) \leq \frac{c'n^2}{D - d},
    \end{align*}
    which completes the proof of this lemma.
    \qed

\noindent It remains to prove the two helper lemmas.

    
    \paragraph{Proof of Lemma~\ref{lemma:technicality-in-mrp-lower-bound-proof}:}
    
    Given $z \in \{\pm 1\}$, we define the sets
    \begin{align*}
        Z_1 \mydefn \left\{ s_i: i \in [k], s_i \in \statespace_1, R_i = z \tau \right\},& \quad\bar{Z}_1 \mydefn  \big(\{s_i\}_{i \in [k]} \cap \statespace_1\big) \setminus Z_1, \quad
        \mbox{and}\\
         Z_2 \mydefn \left\{ s_i: i \in [k], s_i \in \statespace_2, s_i^+ \in [d] \right\}, & \quad \bar{Z}_2 \mydefn \big(\{s_i\}_{i \in [k]} \cap \statespace_2\big) \setminus Z_2
    \end{align*}
    
    By the reward model~\eqref{eq:mrp-approx-factor-lb-reward} in our construction, for any valid pair of subsets  $(\Gamma_1, \Gamma_2)$, under the law $\ProbInst_{\Gamma_1, \Gamma_2, z}^{\otimes k}$, the observations $(s_i, s_i^+, R_i)_{i = 1}^k$ have positive probability if and only if $Z_1 \subseteq \Gamma_1$ and $\Gamma_1 \cap \bar{Z}_1 = \varnothing$. Furthermore, by the symmetry between the elements in $\Gamma_1$, for any $\Gamma_1$ such that $Z_1 \subseteq \Gamma_1$ and $\Gamma_1 \cap \bar{Z}_1 = \varnothing$, the probability of observing $(s_i, s_i^+, R_i)_{i = 1}^k$ under $\ProbInst_{\Gamma_1, \Gamma_2, z}^{\otimes k}$ is independent of the choice of $\Gamma_1$. Consequently, the probability under the mixture distribution $\ProbInst_{z}^{(k)}$ can be calculated as
    \begin{align*}
        \Prob \left( \Gamma_1 \ni s \mid (s_i, s_i^+, R_i )_{i = 1}^k \right)&= \sum_{\stackrel{s \in \Gamma'}{|\Gamma'| = |\statespace_1|/2}} \frac{\Prob \left( (s_i, s_i^+, R_i )_{i = 1}^k \mid \Gamma_1 = \Gamma' \right) \cdot \Prob (\Gamma_1 = \Gamma') }{\Prob \left( (s_i, s_i^+, R_i )_{i = 1}^k \right) } \\
        &= \frac{\abss{ \left\{ \Gamma' \subseteq \statespace_1: ~|\Gamma'| = \frac{1}{2} |\statespace_1|,~ Z_1 \subseteq \Gamma',~ \bar{Z}_1 \cap \Gamma' = \varnothing, ~ s \in \Gamma' \right\} }}{\abss{\left\{ \Gamma' \subseteq \statespace_1: ~|\Gamma'| = \frac{1}{2} |\statespace'|,~ Z_1 \subseteq \Gamma' ~ \bar{Z}_1 \cap \Gamma' = \varnothing \right\} }}\\
        &= \binom{|\statespace_1| - |Z_1| - |\bar{Z}_1| }{|\statespace_1| / 2 - |Z_1|}^{-1} \binom{|\statespace_1| - |Z_1| - |\bar{Z}_1| - 1}{|\statespace_1| / 2 - |Z_1| - 1}\\
        &= \frac{|\statespace_1| /2 - |Z_1|}{|\statespace_1| - |Z_1| - |\bar{Z}_1|}.
    \end{align*}
    By definition, we have $|Z_1| + |\bar{Z}_1| \leq k$, and $|\statespace_1| = \frac{\bigdim - 2 \usedim}{2}$. For $\bigdim \geq \usedim + 8 k$, this yields
    \begin{align*}
        \abss{ \Prob \left(  \Gamma_1 \ni s \mid (s_i, s_i^+, R_i )_{i = 1}^k \right) - \frac{1}{2}} \leq \frac{4 k}{\bigdim - 2 \usedim} \leq \frac{8 k}{\bigdim - \usedim}.
    \end{align*}
    Similarly, by the transition model~\eqref{eq:mrp-approx-factor-lb-transition} in our construction, for any $\Gamma_2 \subseteq \statespace_2$ with $|\Gamma_2| = \frac{1}{2} |\statespace_2|$, under the law $\ProbInst_{\Gamma_1, \Gamma_2, z}^{\otimes k}$, the observations $(s_i, s_i^+, R_i)_{i = 1}^k$ have positive probability if and only if $Z_2 \subseteq \Gamma_2$ and $\Gamma_2 \cap \bar{Z}_2 = \varnothing$. Following exactly the same calculation as above, we arrive at the bound
    \begin{align*}
        \abss{ \Prob \left( \Gamma_2 \ni s \mid (s_i, s_i^+, R_i )_{i = 1}^k  \right) - \frac{1}{2}} \leq \frac{8 k}{\bigdim - \usedim},
    \end{align*}
    as desired.
\qed

    \paragraph{Proof of Lemma~\ref{lem:one-step-birthday-argument-in-mrp-lb-proof}:} 
    Under the conditional distribution $\mathbb{M}^{(k+1)} | (s_i, s_i^+, R_i)_{i = 1}^k$,
for each
    $s \in \statespace_1 \cup \statespace_2$, we have
    \begin{align*}
        \Prob \left( s_{k + 1} = s \right) \leq \frac{2}{|\statespace_1|},\quad \mbox{and} \quad \Prob \left( s_{k + 1}^+ = s \right) \leq \frac{2}{|\statespace_1|}.
    \end{align*}
    Applying a union bound, we arrive at the inequality
    \begin{align*}
         &\Prob \left( \Event_{k + 1}^{(1)} \mid (s_i, s_i^+, R_i)_{i = 1}^k  \right) \\
         &\leq \sum_{\stackrel{i \in [k]}{s_i \in \statespace_1 \cup \statespace_2}} \left( \Prob \left( s_{k + 1} = s_i \right) + \Prob \left( s_{k + 1}^+ = s_i \right) \right) +\sum_{\stackrel{i \in [k]}{s_i \in \statespace_1 \cup \statespace_2}}  \left( \Prob \left( s_{k + 1} = s_i^+ \right) + \Prob \left( s_{k + 1}^+ = s_i^+ \right) \right)\\
         &\leq \frac{8k}{|\statespace_1|} \leq \frac{32 k}{\bigdim - \usedim},
    \end{align*}
    which completes the proof. \qed

\subsection{Proof for elliptic equations} \label{app:elliptic}

In this section, we prove the results for the elliptic equation
example in Section~\ref{subsubsec:elliptic-example}.
    
\subsubsection{Technical results from Section~\ref{subsubsec:elliptic-example}}
\label{Appendix:subsubsec-proof-of-elliptic-lemma}

The main technical result that was assumed in
Section~\ref{subsubsec:elliptic-example} is collected as a lemma
below.
\begin{lemma}
\label{lemma:elliptic-bounded-linear-operator}
    There exists a bounded, self-adjoint, linear operator
    $\widetilde{A} \in \AClass$ and a function $g \in \sobolevone$,
    such that, for all $u, v \in \sobolevone$,
  \begin{subequations}
      \begin{align}
\label{eq:elliptic-bounded-A}        
          \inprod{u}{\widetilde{A} v}_{\sobolevone} & =
          \inprod{u}{\Amat v}_{\Ltwospace}, \\
\label{eq:elliptic-bounded-b}          
          \inprod{u}{g}_{\sobolevone} & =
          \inprod{u}{\bfunc}_{\Ltwospace},
      \end{align}
and such that
\begin{align}
\label{eq:elliptic-bounded-bounds}  
      \strongconvex \vecnorm{u}{\sobolevone}^2 \leq
      \inprod{u}{\widetilde{A} u}_{\sobolevone} \leq \smooth
      \vecnorm{u}{\sobolevone}^2.
    \end{align}
    \end{subequations}
\end{lemma}
\noindent We prove the three claims in turn.

\paragraph{Proof of equation~\eqref{eq:elliptic-bounded-A}:}

For any pair of test functions $u, v \in \sobolevone$, integration by
parts and the uniform ellipticity condition yield
\begin{align*}
  \inprod{u}{\Amat v}_{\Ltwospace} = - \int_\domain u (x) \nabla \cdot
  \big( \afunc (x) \nabla v (x) \big) dx = \int_\domain \nabla u^\top
  \afunc \nabla v dx \leq \smooth \vecnorm{u}{\sobolevone} \cdot
  \vecnorm{v}{\sobolevone}.
\end{align*}
Now given a fixed function $\vvec \in \sobolevone$, the above equation
ensures that $\inprod{\cdot}{\Amat \vvec}_{\Ltwospace}$ is a bounded
linear functional. By the Riesz representation theorem, there exists a
unique function $\vvec'\in \sobolevone$ with
$\vecnorm{\vvec'}{\sobolevone} \leq \smooth
\vecnorm{\vvec}{\sobolevone}$, such that
\begin{align*}
    \forall u \in \sobolevone, \quad \inprod{u}{\Amat v}_{\Ltwospace} = \inprod{u}{\vvec'}_{\sobolevone}.
\end{align*}
Clearly, the mapping from $\vvec$ to $\vvec'$ is linear, and we have
$\vecnorm{\vvec'}{\sobolevone} \leq \smooth
\vecnorm{\vvec}{\sobolevone}$ for any $\vvec \in \sobolevone$. Thus,
the mapping $\vvec \mapsto \vvec'$ is a bounded linear operator. Using
$\widetilde{\Amat}$ to denote this operator,
equation~\eqref{eq:elliptic-bounded-A} then directly follows. It
remains to verify that $\widetilde{\Amat}$ is self-adjoint. Indeed,
for $u, v \in \sobolevone$, we have the identity
\begin{align*}
    \inprod{u}{\widetilde{\Amat} v}_{\sobolevone} = \inprod{u}{\Amat
      v}_{\Ltwospace} = \int_\domain \nabla u^\top \afunc \nabla v dx
    = - \int_\domain v \nabla \cdot (\afunc \nabla u) dx =
    \inprod{v}{\Amat u}_{\Ltwospace} = \inprod{v}{\widetilde{\Amat}
      u}_{\sobolevone},
\end{align*}
which proves the self-adjoint property.

\paragraph{Proof of equation~\eqref{eq:elliptic-bounded-b}:}
Since the domain $\domain$ is bounded and connected, there exists a
constant $\poincare$ depending only on $\domain$, such that the
following Poincar\'{e} equation holds:
\begin{align}
    \forall \vvec \in \sobolevone, \quad \vecnorm{\vvec}{\Ltwospace}^2
    \leq \frac{1}{\poincare}
    \vecnorm{\vvec}{\sobolevone}^2. \label{eq:poincare-ineq-in-domain}
\end{align}
For any test function $u \in \sobolevone$,
equation~\eqref{eq:poincare-ineq-in-domain} leads to the bound
\begin{align*}
    \inprod{u}{\bfunc}_{\Ltwospace} \leq \vecnorm{\bfunc}{\Ltwospace}
    \cdot \vecnorm{u}{\Ltwospace} \leq \frac{1}{\poincare}
    \vecnorm{\bfunc}{\Ltwospace} \cdot \vecnorm{u}{\sobolevone}.
\end{align*}
So $\inprod{\cdot}{\bfunc}_{\Ltwospace}$ is a bounded linear
functional on $\sobolevone$. Again, by the Riesz representation
theorem, there exists a unique $g \in \sobolevone$, such that
$\inprod{u}{\bfunc}_{\Ltwospace} = \inprod{u}{g}_{\sobolevone}$
\mbox{for all $u \in \sobolevone$,} which completes the proof.

\paragraph{Proof of equation~\eqref{eq:elliptic-bounded-bounds}:}

For any test function $u \in \sobolevone$, we note that
\begin{align*}
    \inprod{u}{\widetilde{\Amat} u}_{\sobolevone} = \inprod{u}{\Amat
      u}_{\Ltwospace} = \int_\domain \big( \nabla u (x) \big)^\top
    \afunc (x) \big( \nabla u (x) \big) dx.
\end{align*}
From our uniform ellipticity condition, we know that $\strongconvex
I_\domaindim \preceq \afunc (x) \preceq \smooth I_\domaindim$ for any
$x \in \domain$. Substituting this relation yields
$\strongconvex \vecnorm{u}{\sobolevone}^2 \leq
    \inprod{u}{\widetilde{\Amat} u}_{\sobolevone} \leq \smooth
    \vecnorm{u}{\sobolevone}^2$, as claimed.
    

\subsubsection{Proof of Corollary~\ref{corr:elliptic}}
\label{Appendix:subsubsec-proof-of-corr-elliptic}

The matrices $\SpecMat, \Sigma_\Lmat, \Sigma_\bvec$ for the projected
problem instances can be obtained by straightforward calculation. In
order to apply Theorem~\ref{thm:linear-oracle-ineq}, it remains to
verify the assumptions.

By Lemma~\ref{lemma:elliptic-bounded-linear-operator}, the operator
$\Lmat$ is self-adjoint in $\Xspace$, and is sandwiched as $0 \leq
\inprod{u}{\Lmat u}_{\sobolevone} \leq \big ( 1 -
\frac{\strongconvex}{\smooth} \big) \vecnorm{u}{\sobolevone}^2$ for
all $u \in \Xspace$.  This yields the operator norm bound
$\hilopnorm{\Lmat} \leq 1 - \frac{\strongconvex}{\smooth}$.

Now we verify the conditions in
Assumption~\ref{assume-second-moment}. For any basis function $\phi_j$
with $j \in [\usedim]$ and vector $\vvec \in \sobolevone$, we have
\begin{align*}
\Exs \inprod{\phi_j}{(\Lmat_i - \Lmat) u}_{\sobolevone}^2 & \leq
\frac{1}{\smooth^2} \Exs \left( \int_\domain \delta_{x_i} \nabla
\phi_j (x)^\top \afunc (x_i) \nabla u (x) dx \right)^2 +
\frac{1}{\smooth^2} \Exs \left( \int_\domain \delta_{x_i} \nabla
\phi_j (x)^\top W_i \nabla u (x) dx \right)^2 \\
&= \frac{1}{\smooth^2} \int_\domain \left( \nabla \phi_j (x)^\top
\afunc (x) \nabla u (x) \right)^2 dx + \frac{1}{\smooth^2}
\int_\domain \Exs \left( \nabla \phi_j (x)^\top W_i \nabla u (x)
\right)^2 dx \\ &\leq \frac{1}{\smooth^2} \int_\domain \vecnorm{\nabla
  \phi_j (x)}{2}^2 \opnorm{\afunc (x)}^2 \vecnorm{\nabla u (x) }{2}^2
dx + \frac{2}{\smooth^2} \int_\domain \vecnorm{\nabla \phi_j (x)}{2}^2
\cdot \vecnorm{\nabla u (x)}{2}^2 dx \\
& \leq \left(1 + \frac{2}{\smooth^2} \right) \max_{j \in [\usedim]}
\sup_{x \in \domain} \vecnorm{\nabla \phi_j}{2}^2 \cdot \int_\domain
\vecnorm{\nabla u (x) }{2}^2 dx \\
& \leq \sigmaA^2 \vecnorm{u}{\sobolevone}^2,
\end{align*}
and
\begin{align*}
    \Exs \inprod{\phi_j}{\bvec_i - \bvec}_{\sobolevone}^2 & \leq
    \frac{1}{\smooth^2} \Exs \left( \int_\domain \delta_{y_i} \phi_j
    (y) \bfunc (y_i) dy \right)^2 + \frac{1}{\smooth^2} \Exs \left(
    \int_\domain \delta_{y_i} \phi_j (y) g_i dy \right)^2\\ &=
    \frac{1}{\smooth^2} \int_\domain \phi_j (y)^2 \bfunc (y)^2 dy +
    \frac{1}{\smooth^2} \int_\domain \phi_j (y)^2 \Exs [g_i^2] dy
    \\ &\leq \frac{1}{\smooth^2} \sup_{x \in \domain} |\phi_j|^2
    \int_\domain \big( \bfunc (y)^2 + 1 \big) dy\\ &= \frac{
      \vecnorm{\bfunc}{\Ltwospace}^2 + 1}{\smooth^2} \max_{j \in
      [\usedim]} \sup_{x \in \domain} |\phi_j|^2.
\end{align*}
Therefore, Assumption~\ref{assume-second-moment} is satisfied with
constants $(\sigmaA, \sigmab)$. Invoking
Theorem~\ref{thm:linear-oracle-ineq} completes the proof.


\section{Models underlying simulations in Figure~\ref{figure:simulation}}
\label{Appendix:simulation-details}

The simulation results shown in Figure~\ref{figure:simulation} are
generated by constructing random transition matrices based on the
following random graph models: {\small{
\begin{description}
\item[Erd\"{o}s-R\'{e}nyi random graph:] Given $\usedim, N \in
  \mathbb{N}_+$ and $a > 1$, we consider the following sampling
  procedure. Let $G$ be an Erd\"{o}s-R\'{e}nyi random graph with
  $N$ vertices and edge probability $p = \frac{a}{N}$, and
  take $\widetilde{G}$ to be its largest connected
      component. (When $c > 1$, the number of vertices in
      $\widetilde{G}$ is of order $\Theta (N)$. See the
      monograph~\cite{durrett2007random} for details.) For each vertex
      $v \in V (\widetilde{G})$, we associate it with an independent
      standard Gaussian random vector $\phi_v \sim \mathcal{N} (0,
      I_d)$. Let $V (\widetilde{G})$ be the state space and let the Markov transition kernel
      $\transition$ be the simple random walk on $\widetilde{G}$.
      
 In Figure~\ref{figure:simulation} (a), we take the number of vertices to be
  $N = 3000$ and the feature dimension to be $\usedim = 1000$. The edge density
 parameter is chosen as $a = 3$. The resulting giant connected
 component contains $2813$ vertices.
\item[Random geometric graph:] Given a pair of positive integeres $(\usedim, N)$
  and scalar $r > 0$, we consider the following sampling procedure. For each
  vertex $i \in [N]$, we associate it with an independent standard
  Gaussian random vector $\phi_i \sim \mathcal{N} (0, I_\usedim)$. The
  graph $G$ is then constructed such that $(i, j) \in E (G)$ if and
  only if $\vecnorm{\phi_i - \phi_j}{2} \leq r$. (See the
  monograph~\cite{penrose2003random} for more details of this random
  graph model.) Take $\widetilde{G}$ to be the largest connected
  component of $G$. We take $V (\widetilde{G})$ as the state space,
  and let $\transition$ be the simple random walk on $\widetilde{G}$.
      
In Figure~\ref{figure:simulation} (b), we take the number of vertices to be
$N = 3000$ and the feature dimension to be $\usedim = 2$. The distance threshold is
chosen as $r = 0.1$. The resulting giant connected component contains
$2338$ vertices.
\end{description}
}}

Despite their simplicity, the two random graph models capture
distinct types of the behavior of the resulting random walk in feature
space: in the former model, the transition kernel makes ``big jumps''
in the feature space, and the correlation between two consecutive
states is small; in the latter model, the transition kernel makes
``local moves'' in the feature space, leading to large correlation.

\end{document}